%% file: paper.tex
\documentclass[preprint,authoryear,12pt]{elsarticle}

\usepackage{natbib}
\usepackage{url}

\usepackage{amsmath}
\usepackage{amssymb}

\input{config/packages}

\input{config/definitions}

\graphicspath{{images/}{images/segmentation_fig/}{images/overview_fig/}}

\acrodef{convnet}[ConvNet]{Convolutional Neural Network} 
\acrodef{slam}[SLAM]{Simultaneous Localization And Mapping}
\acrodef{iou}[IoU]{Intersection over Union}
\acrodef{miou}[mIoU]{Mean Intersection over Union}
\acrodef{eo}[EO]{Earth Observation}
\acrodef{eot}[EOT]{Earth Observation Tiles}
\acrodef{crs}[CRS]{Coordinate Reference System}

\makeatletter
\def\ps@pprintTitle{%
	\let\@oddhead\@empty
	\let\@evenhead\@empty
	\def\@oddfoot{}%
	\let\@evenfoot\@oddfoot}
\makeatother

\begin{document}
\hypersetup{
	allcolors   = darkmidnightblue,
}

\title{Geo-Tiles for Semantic Segmentation of Earth Observation Imagery}

\author[1]{Sebastian Bullinger\corref{cor1}}		
\ead{sebastian.bullinger@iosb.fraunhofer.de}
\author[1]{Florian Fervers}	
\author[1]{Christoph Bodensteiner}	
\author[1]{Michael Arens}
\cortext[cor1]{Corresponding author}
\affiliation[1]{organization={Fraunhofer IOSB},
	addressline={Gutleuthausstr.1},
	postcode={76275},
	city={Ettlingen},
	country={Germany}
}

\begin{abstract}
\input{abstract}
\end{abstract}
\maketitle

\section{Introduction}
\label{section_introduction}
\input{content/introduction}

\section{Image Tiling}
\label{section_image_tiling}
\input{content/image_tiling}

\section{Earth Observation Tiles with Consistent Spatial Extent}
\label{section_earth_observation_tiles}
\input{content/gsd_based_image_tiling}

\section{Data Augmentation and Tile Fusion}
\label{section_augmentation_and_fusion}
\input{content/improvement}

\section{Experiments}
\label{section_experiments}
\input{content/experiments}

\section{Conclusion}
\input{content/conclusion}

\FloatBarrier
\newpage

{\small
\bibliographystyle{elsarticle-harv}
\bibliography{egbib}
}

\FloatBarrier
\newpage

\section{Supplementary Material}
\input{content/supplementary}
\FloatBarrier
\newpage

\end{document}

%% file: config/packages.tex
\usepackage{geometry}
\usepackage{pdflscape}

\usepackage{xspace}
\usepackage{amssymb}	%
\usepackage{pifont}		%

\usepackage{color}
\usepackage[table, dvipsnames]{xcolor}

\usepackage[breaklinks=true,colorlinks,bookmarks=false]{hyperref}
\definecolor{darkmidnightblue}{rgb}{0.0, 0.2, 0.4}

\usepackage{placeins}	%

\usepackage{caption} 	%
\usepackage{subcaption} %
\usepackage{graphicx}

\usepackage{pgfplots}
\usepgfplotslibrary{groupplots}
\usetikzlibrary{matrix}

\usetikzlibrary{decorations.pathreplacing,calligraphy}
\usetikzlibrary{calc}
\usetikzlibrary{fit}
\usetikzlibrary{patterns}
\usetikzlibrary{shapes,arrows,positioning}
\pgfplotsset{compat=newest}

\usepackage{makecell}
\usepackage{tabularx}
\usepackage{arydshln}	%

\usepackage{mathtools}
\usepackage{amsmath}
\usepackage{amssymb}
\usepackage{siunitx}
\usepackage{nicefrac}

\usepackage{fancyvrb}
\usepackage{tgcursor}
\usepackage{listings}
\usepackage{xifthen}	%
\usepackage{ifthen}

\usepackage[printonlyused,withpage]{acronym}

%% file: config/definitions.tex
\definecolor{cadmiumgreen}{rgb}{0.0, 0.42, 0.24}
\definecolor{cornellred}{rgb}{0.7, 0.11, 0.11}
\definecolor{cornflowerblue}{rgb}{0.39, 0.58, 0.93} 
\definecolor{gray}{rgb}{0.80, 0.80, 0.80}
\definecolor{hlcolor}{rgb}{1.00, 1.00, 0.5}
\definecolor{dkgreen}{rgb}{0,0.6,0}
\definecolor{gray}{rgb}{0.5,0.5,0.5}
\definecolor{mauve}{rgb}{0.58,0,0.82}

\pgfplotscreateplotcyclelist{customcolormarkers}{
	teal,every mark/.append style={fill=teal!80!black},mark=*\\
	teal,every mark/.append style={fill=teal!80!black},mark=square*\\
	orange,every mark/.append style={fill=orange!80!black},mark=*\\
	orange,every mark/.append style={fill=orange!80!black},mark=square*\\
	cyan!60!black,every mark/.append style={fill=cyan!80!black},mark=*\\
	cyan!60!black,every mark/.append style={fill=cyan!80!black},mark=square*\\
	lime!80!black,every mark/.append style={fill=lime},mark=*\\
	lime!80!black,every mark/.append style={fill=lime},mark=square*\\
	red,every mark/.append style={solid,fill=red!80!black},mark=*\\
	red,every mark/.append style={solid,fill=red!80!black},mark=square*\\
	yellow!60!black,every mark/.append style={solid,fill=yellow!80!black},mark=*\\
	yellow!60!black,every mark/.append style={solid,fill=yellow!80!black},mark=square*\\
	blue,every mark/.append style=solid,mark=*\\
	blue,every mark/.append style=solid,mark=square*\\
	black,every mark/.append style={solid,fill=gray},mark=*\\
	black,every mark/.append style={solid,fill=gray},mark=square*\\
}

\makeatletter
\DeclareRobustCommand\onedot{\futurelet\@let@token\@onedot}
\def\@onedot{\ifx\@let@token.\else.\null\fi\xspace} %
        
\newcommand{\etal}{\emph{et al}\@onedot} 	%
\newcommand{\eg}{\emph{e.g}\onedot} 			%
\newcommand{\ie}{\emph{i.e}\onedot} 			%
\newcommand{\wrt}{w.r.t\onedot} 				 %

\makeatother

\acrodef{ConvNet}{Convolutional Neural Network}
\acrodef{gsd}[GSD]{Ground Sampling Distance}
\acrodef{crs}[CRS]{Coordinate Reference System}

\newcommand{\cmark}{{\large\ding{51}}}                             %
\newcommand{\xmark}{{\large\ding{55}}}                             %
\newcommand{\figref}[1]{Fig.~\ref{#1}}                         %
\newcommand{\tabref}[1]{Tab.~\ref{#1}}                        %

\newcommand{\secref}[1]{Section~\ref{#1}}

\makeatletter
\newcommand\appendtographicspath[1]{%
		\g@addto@macro\Ginput@path{#1}%
	}
\makeatother

\newcommand{\subtableoffset}{\medskip}	%

\newcommand{\tablefontsizestiny}{\scriptsize}

\renewcommand{\vec}[1]{\ensuremath{\mathbf{#1}}}          %

\DeclarePairedDelimiter\squaredbracket{[}{]}
\DeclarePairedDelimiter\roundbracket{(}{)}
\newcommand{\sqbr}[1]{\squaredbracket*{#1}}
\newcommand{\rbr}[1]{\roundbracket*{#1}}

\DeclareSIUnit{\px}{px}

%% file: abstract.tex
To cope with the high requirements during the computation of semantic segmentations of earth observation imagery, current state-of-the-art pipelines divide the corresponding data into smaller images. Existing methods and benchmark datasets oftentimes rely on pixel-based tiling schemes or on geo-tiling schemes employed by web mapping applications. The selection of subimages (comprising size, location and orientation) is crucial. It affects the available context information of each pixel, defines the number of tiles during training, and influences the degree of information degradation while down- and up-sampling the tile contents to the size required by the segmentation model. We propose a new segmentation pipeline for earth observation imagery relying on a tiling scheme that creates geo-tiles based on the geo-information of the raster data. This approach exhibits several beneficial properties compared to pixel-based or common web mapping approaches. The proposed tiling scheme shows flexible customization properties regarding tile granularity, tile stride and image boundary alignment. This allows us to perform a tile specific data augmentation during training and a substitution of pixel predictions with limited context information using data of overlapping tiles during inference. The generated tiles show a consistent spatial tile extent \wrt heterogeneous sensors, varying recording distances and different latitudes. We demonstrate how the proposed tiling system allows to improve the results of current state-of-the-art semantic segmentation models. To foster future research we make the source code publicly available.

%% file: content/introduction.tex
The resolution of earth observation images often exceeds the possible input size of many state-of-the-art deep-learning-based methods for semantic segmentation. Publicly available (benchmark) datasets \citep{Demir_2018_CVPR_Workshops,Etten_2018_Arxiv_SpaceNet,MaggioriIGARSS2017,MohantyFAI2020} and processing pipelines such as RasterVision \citep{AzaveaGithub2022} or TorchSat \citep{TorchSatGithub2022} tackle this problem by dividing the earth observation images into smaller tiles based on a fixed number of pixels. \\
For specific datasets (with homogeneous sensors and a constant recording distance) this approach allows to create tiles with a constant spatial extent. However, images from multi-source datasets with varying geo-properties cover different spatial extents (such as earth observation imagery captured from distinct distances or by varying sensors) resulting in image tiles with distinct spatial properties - see \tabref{table_tile_comparison}. \\
To generate tiles that are independent of specific sensor properties or recording distances, well known earth observation imagery pipelines such as RoboSat \citep{MapboxGithub2018} or Label Maker \citep{BollingerGithub2022} apply a tiling scheme on a spherical mercator projection (\eg EPSG:3857 \citep{EPSG3857}) of the data (as it is used by mapping services such as OpenStreetMap or Google Maps). However, common web map tiling schemes exhibit several properties that are unsuitable for the segmentation of raster image datasets (as shown in \tabref{table_tile_comparison}) such as a)~variation of the spatial tile extents in meters over different latitudes because of the used spherical mercator projection, b)~non-optimal alignment to and coverage of the corresponding raster images (see \figref{figure_tiling_type_c}), c)~restricted customization of tile size granularity and tile overlap, d)~limitation to specific coordinate systems such as EPSG:3857, and e)~restriction to normalized raster images. \\
We exploit the geo-information of the raster images (including \ac{gsd}, \ac{crs} and transformation matrix) to generate geo-tiles for multi-source and multi-site datasets covering a fixed spatial extent (which overcomes the limitations of pixel-based and web map tiling schemes). A comparison of the different tiling schemes is shown in \tabref{table_tile_comparison}. \\
For images with varying degrees of detail information (\ie different \ac{gsd} values) a specific tile size in meters corresponds to different tile sizes in pixels. Thus, when processing multi-source datasets it is necessary to down- or up-sample the corresponding tile information to match the required input size (measured in pixels) of the segmentation model. \\
The selection of the tile size always represents a trade-off between available detail and context information. The determination of an optimal tile size is (in general) a challenging task, because a)~the complexity of modern segmentation models makes it difficult to derive specific requirements regarding detail or context information, b)~different semantic categories usually require a different degree of detail or context information for reliable predictions of class labels, and c)~real-world datasets contain heterogeneous data \wrt geo-properties and semantic categories. Using an empirical evaluation allows us to adapt the configuration of the proposed tiling scheme to the requirements for individual databases. \\
In contrast to common web maps, our tiling scheme is able to create overlapping tiles. This creates two beneficial capabilities in the context of semantic segmentation of earth observation imagery. During training it enables us to perform an earth observation specific data augmentation and during inference it allows to substitute pixel predictions with a limited amount of context information (\ie pixels close to image boundaries) using suitable pixel predictions of adjacent tiles. We propose a tile fusion approach that is optimal \wrt to the spatial properties of the corresponding tiling scheme.

\input{content/tables/table_tile_comparison}
\input{content/figures/figure_tiling_type}

\subsection{Related Work}
Many challenges and datasets in the context of earth observation such as the \emph{DeepGlobe Challenge} \citep{Demir_2018_CVPR_Workshops}, the \emph{SpaceNet Building Detection Challenge} \citep{Etten_2018_Arxiv_SpaceNet}, the \emph{Mapping Challenge} \citep{MohantyFAI2020}, the \emph{Inria Aerial Image Labeling Dataset} \citep{MaggioriIGARSS2017}, or the \emph{SEN12MS} dataset \citep{SchmittISPRS2019} provide the corresponding imagery in the form of fixed (pixel-based) tiles and focus on the design of novel methods for detection or segmentation. Although the derived results are subject to the pre-defined tiling of the image data, the impact of different tiling methods is not analyzed. \\
Other datasets such as the \emph{Open Cities AI Challenge Dataset} \citep{OpenCitiesAI}, the \emph{ISPRS 2D Semantic Labeling Dataset} \citep{RottensteinerPRS2014}, or the \emph{Christchurch Aerial Semantic Dataset} \citep{RandrianarivoIGARSS2013} provide full raster images. Existing pipelines like \emph{RasterVision} \citep{AzaveaGithub2022} and \emph{GeoTile} \citep{GeoTile2022}, or previous works such as \cite{Weber2021IGARSS} propose to perform a pixel based tiling of the plain raster images to address the processing limitations of current models. Other libraries such as \emph{TorchGeo} \citep{TorchGeo2022} allow to define tiles in units of the raster image \ac{crs}. However, datasets such as \emph{Open Cities AI} \citep{OpenCitiesAI} consist of data captured by different sensors - resulting in raster images with heterogeneous geo-properties and varying pixel extents caused by image specific \ac{gsd} values. Such settings are of particular interest since they reflect the challenges of real world applications. \\
To process such inhomogeneous image collections, pipelines like RoboSat \citep{MapboxGithub2018} use a tiling scheme based on a spherical mercator projection \citep{EPSG3857}. This tiling scheme (which is the de facto standard for web mapping applications) shows some properties that limit its applicability to learning based contexts. For instance, the created tiles show different spatial extents in dependence of the latitude. \\
We propose a new method that defines tiles in real world units (\ie meters) using the geo-properties of the corresponding raster images. Based on this approach we are able to generate additional overlapping tiles allowing to perform a tile specific data augmentation during training and to fuse predictions of overlapping tiles during inference. \\
We conduct experiments on the \emph{Open Cities AI Challenge Dataset} \citep{OpenCitiesAI} and the \emph{ISPRS 2D Semantic Labeling Dataset} \citep{RottensteinerPRS2014} considering several modern segmentation models including PSPNet \citep{Zhao2017CVPR}, K-Net \citep{Zhang2021NeurIPS}, UPerNet \citep{Xiao2018ECCV}, Segmenter \citep{Strudel2021ICCV} and BEiT \citep{Bao2022ICLR}.

\subsection{Contribution and Paper Overview}
This paper proposes a semantic segmentation pipeline which specifically addresses the requirements of processing earth observation data in the context of semantic segmentation. The individual contributions with the corresponding sections are listed below.\\
1.)~The tiling system allows to determine an optimal coverage for each raster image \wrt the number of required tiles and overlapping pixels (\secref{section_image_tiling}). This reduces the generation of border tiles with marginal image data, which could potentially hamper the result of the training and inference phase.\\
2.)~The tiling scheme exhibits several beneficial properties including a consistent spatial extent \wrt heterogeneous sensors, different recording distances and varying latitudes. It also supports arbitrary configurations (regarding granularity, tile overlap as well as tiling scheme offset), arbitrary coordinate systems and non-normalized raster images (\secref{section_earth_observation_tiles}).
\\
3.)~We introduce a fusion algorithm (\secref{section_augmentation_and_fusion}) for tile predictions that substitutes pixels with limited context information using suitable data of adjacent (overlapping) tiles. Since limited context information potentially hampers the inference result, this approach usually leads to better semantic segmentations. For the presented fusion step we derive a criterion allowing us to determine segmentation results with maximum context information regarding the parameter configuration of the tiling scheme. Our approach is applicable to arbitrary semantic segmentation approaches, which demonstrates its versatility. \\
4.)~Due to the flexibility of the presented tiling scheme we are able to assess different aspects relevant for the task of semantic segmentation including the quantitative evaluation of varying (real world) tile sizes, tiling specific training data augmentations as well as fusion of tile predictions (\secref{section_experiments}). We perform the experiments using the Open Cities AI Challenge Dataset \citep{OpenCitiesAI} and the ISPRS 2D Semantic Labeling Dataset \citep{RottensteinerPRS2014} which show that the proposed tiling system is able to reduce the training error while also improving the segmentation results during inference.\\
5.)~We make the full tile processing source code (comprising tiling, rasterization, tile fusion and tile aggregation) publicly available\footnote{\label{source_code}Project page: \url{https://sbcv.github.io/projects/earth_observation_tiles}}.

%% file: content/tables/table_tile_comparison.tex
\begin{table*}
	\resizebox{\textwidth}{!}{%
	\begin{tabular}{l | c c c} 
		Tiling Scheme Property & \makecell[c]{Pixel-based \\ Tiling} & \makecell[c]{Web Map \\ Tiling} & \makecell[c]{Earth Observation \\ Tiling (ours)} \\
		\hline
	 	Consistent spatial extent for different sensors & \xmark & \cmark & \cmark \\
	 	Consistent spatial extent for different recording distances  & \xmark & \cmark & \cmark \\
		Consistent spatial extent for different latitudes  & \cmark & \xmark & \cmark \\
		Optimal raster image coverage / alignment to raster image & \cmark  & \xmark & \cmark \\
		Customization (\eg arbitrary granularity or overlap) & \cmark  & \xmark & \cmark \\
		Arbitrary coordinate system support & \cmark & \xmark &  \cmark \\
		Non-normalized raster image support & \cmark  & \xmark & \cmark \\
	\end{tabular}
	} \\
	\caption{Property comparison of different tiling schemes.}
	\label{table_tile_comparison}
\end{table*}

%% file: content/figures/figure_tiling_type.tex
\begin{figure*}
	\centering
	\newcommand{\tilingtypefigurewidth}{0.4\textwidth}
	\begin{subfigure}[t]{\tilingtypefigurewidth}
		\centering
		\includegraphics[width=\textwidth]{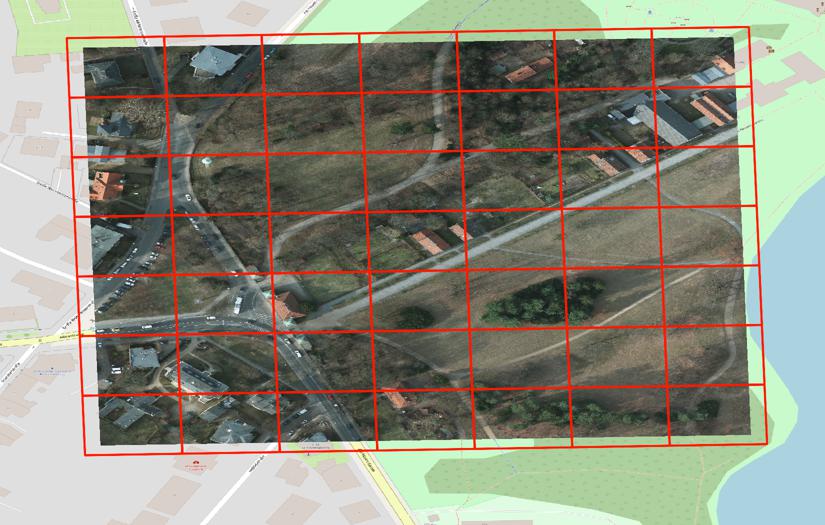} \\
		\vspace{0.05cm}
		\includegraphics[width=\textwidth]{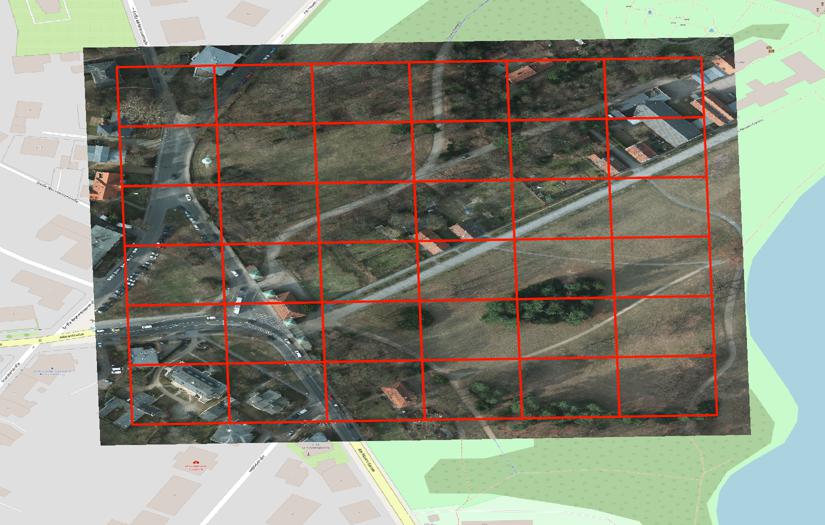}
		\caption{Proposed tiling approach centered \wrt the raster image using a tile size and a tile stride of 45m.}
		\label{figure_tiling_type_a}
	\end{subfigure}
	\hfil
	\begin{subfigure}[t]{\tilingtypefigurewidth}
		\centering
		\includegraphics[width=\textwidth]{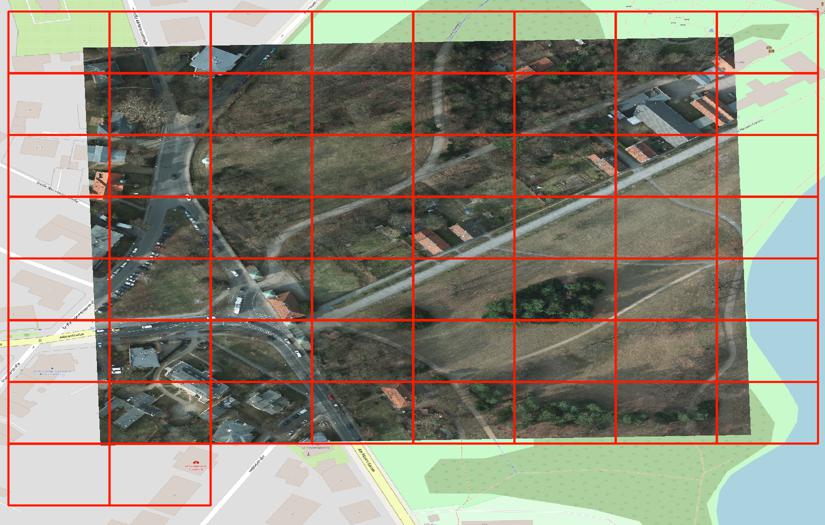} \\
		\vspace{0.05cm}
		\includegraphics[width=\textwidth]{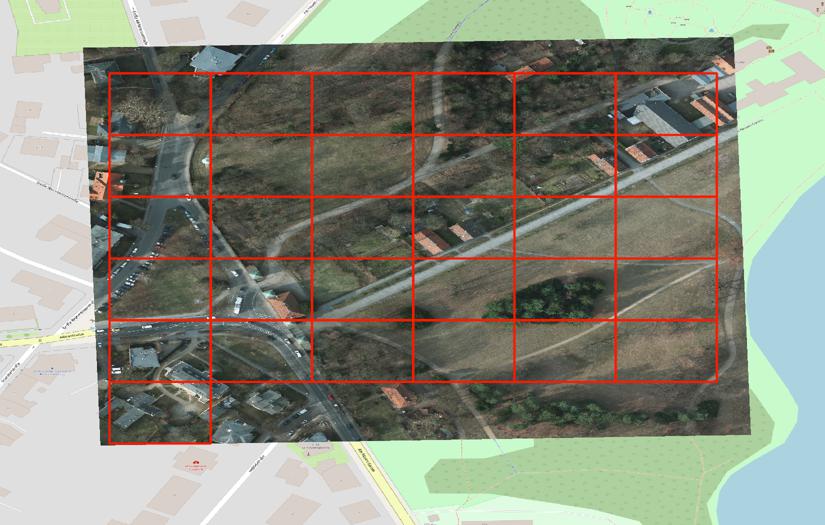}
		\caption{Web map tiling scheme on zoom level 19 ($\sim$46m on the ISPRS potsdam dataset).}
		\label{figure_tiling_type_c}
	\end{subfigure}
	\caption{Visual comparison of the proposed tiling scheme with commonly used web map tiles on a raster image of the ISPRS Potsdam dataset. Apparent differences include the tile alignment and the raster image coverage. The first row shows tiling configurations fully covering the raster image, while the second row depicts only tiles without missing tile data. Note that the proposed tiling scheme (\ie \figref{figure_tiling_type_a}) is more suitable to represent the raster information than a common web map tiling scheme (see \figref{figure_tiling_type_c}), due to its raster image alignment.}
	\label{figure_tiling_type}
\end{figure*}

%% file: content/image_tiling.tex
While tiling of raster images represents a common approach in the context of earth observation, previous works \cite{Demir_2018_CVPR_Workshops,Etten_2018_Arxiv_SpaceNet,MaggioriIGARSS2017,MohantyFAI2020} and available processing chains \cite{AzaveaGithub2022,TorchSatGithub2022} utilize only basic forms of image tiling (\ie adjacent tiles without overlap and no offset values) and provide no formalization of the process. \\
We address this issue by providing a mathematical description of a tiling scheme for arbitrary tiling parameters (\ie arbitrary tile stride and tile size values). The proposed tiling scheme is optimal in the sense that it allows to cover images with a minimum number of tiles. We show how to obtain tiles that are symmetrical \wrt raster image boundaries including cases where tile stride and tile extent are not aligned.  \\
Tiling schemes starting at one of the image corners usually result in an asymmetrical tile coverage (when image and tiling extent differ). Defining a tile at the center of the raster image to obtain a symmetrical tiling scheme is also not suitable, since this does not ensure a tiling with a minimum number of tiles. For example, raster images covered by an even number of tiles do not contain a center tile (see \figref{figure_tiling_type_a}). \\
\input{content/figures/figure_gsd_based_tiling}

\subsection{Tiling Scheme Definition}

The proposed tiling scheme is oriented along the raster image and is defined by several parameters comprising an offset in x- and y-direction ($o_x$ and $o_y$) relative to the raster image (representing the origin of the tiling scheme), the stride size in x- and y-direction ($s_x$ and $s_y$), the tile extent (identical for all tiles) in x- and y-direction ($t_x$ and $t_y$), as well as number of tiles in x- and y-direction ($n_x$ and $n_y$). \\
The tile stride defines the distance of the starting point of subsequent tiles. For instance, $s_x < t_x$ results in overlapping tiles while $s_x > t_x$ creates tiles with a gap between subsequent columns. An example illustration is shown in \figref{figure_gsd_based_tiling}. \\
While tile stride and tile size must be provided by the user, the remaining parameters (\ie tiling scheme offset and number of tiles) are derived from the raster image. \\
The tiling scheme does not consider parameters like individual tile orientation and size, because these adjustments can be efficiently performed during data augmentation.

\input{content/figures/figure_training_sample_enhancement}

\subsection{Tile Number Computation for Optimal Raster Image Coverage}
\label{section_tile_number_optimal_coverage}

In order to determine the optimal number of tiles required to cover the raster image we analyze both directions individually. In the following we derive the solution only for the x-direction. \\
We start by considering a solution for cases where the first tile starts at the raster image origin (\ie the offset $o_x$ is 0) and the boundary of last tile aligns with the boundary of the raster image. Afterwards we extend the result to arbitrary cases. \\
In the aligned case (defined above) the number of tiles along the x-direction $\nu (r_x, s_x, t_x)$ depends on the raster extent (or raster size) $r_x$, the tiling stride $s_x$, as well as the tile extent (or tile size) $t_x$ and is given by \eqref{equation_tiles_in_raster_x_aligned}. As a result of the alignment \eqref{equation_tiles_in_raster_x_aligned} yields a true integer value.
\begin{equation}
	\nu (r_x, s_x, t_x) =  (r_x + \underbrace{s_x - t_x}_{\coloneqq c_x}) / s_x	
	\label{equation_tiles_in_raster_x_aligned}
\end{equation}
To prove the correctness of \eqref{equation_tiles_in_raster_x_aligned} we discuss all (\ie three) possible cases separately. \\
Case 1 ($t_x = s_x$): In this case \eqref{equation_tiles_in_raster_x_aligned} simplifies to the trivial case of $r_x  / s_x = r_x / t_x $. \\
Case 2 ($t_x > s_x$): In this case $r_x / s_x$ represents a value greater than the true number of tiles covering the raster image, since the pixels of the raster image overlapping with the last stride are already covered by the tile corresponding to the second last stride (see tile $2$ and the magenta area in \figref{figure_training_sample_enhancement_a}). To obtain the actual number of tiles (\ie an integer result) the area $r_x$ must be trimmed by some value $c_x$ when dividing with stride $s_x$. Concretely, $c_x$ is given by the negative number of pixels the tile size $t_x$ is exceeding the stride $s_x$ (\ie $c_x = s_x - t_x$). The resulting area $r_x + c_x$ is a multiple of stride $s_x$ with $c_x$ representing a (negative) trimming value. \\
Case 3 ($t_x < s_x$): In this case $r_x / s_x$ represents a value lesser than the true number of tiles covering the raster image, since the last stride does not fit into the raster image -- but the actual tile does (see the cyan area in \figref{figure_training_sample_enhancement_b}). To obtain the actual number of tiles (\ie an integer result) the area $r_x$ must be padded by some value $c_x$ when dividing with stride $s_x$. Concretely, $c_x$ is given by the number of pixels inside a stride $s_x$ not covered by the tile size $t_x$ (\ie $c_x = s_x - t_x$). In this case the resulting area $r_x + c_x$ is a multiple of stride $s_x$ with $c_x$ representing a padding value (\ie a positive number). \\
In general the boundary of the last tile does not align with the boundary of the raster image resulting in remaining pixels that are not covered by the tiling result. In this case there are two options: Either the tiling result covers only a part of the raster image (by applying a floor operation) or the tiling result covers the full raster image by introducing additional tiles exhibiting an overhang \wrt the image boundaries (by applying a ceil operation) - see \figref{figure_tiling_type_a} for an example. \\
We model both cases by introducing an additional function $\rho(\cdot)$ in \eqref{equation_tiles_in_raster_x_aligned} representing a floor or ceil operation as shown in \eqref{equation_tiles_in_raster_x}. 
\begin{equation}
	\mu (r_x, s_x, t_x, \rho) = \rho \left(\frac{r_x + s_x - t_x}{s_x}\right)
	\label{equation_tiles_in_raster_x}
\end{equation}
Extending \eqref{equation_tiles_in_raster_x} for two dimensions (denoted in the following as $\mu_{x,y}$) yields \eqref{equation_tiles_in_raster}.
\begin{equation} 
	\begin{split}
		\mu_{x,y}(r_x, r_y, s_x, s_y, t_x, t_y, \rho) 
		\coloneqq \mu(r_x, s_x, t_x, \rho) \mu(r_y, s_y, t_y, \rho)  
	\end{split}
	\label{equation_tiles_in_raster}
\end{equation}

\subsection{Tiling Scheme Alignment}

Using \eqref{equation_tiles_in_raster_x} to determine the minimal number of tiles required to cover a given raster image along the x-direction allows to compute the corresponding covered distance (defined by the outermost boundary pixels of the outermost tiles) according to \eqref{equation_tile_coverage_x}.
\begin{equation}
	\begin{split}
	\kappa (r_x, s_x, t_x, \rho) = \mu(r_x, s_x, t_x, \rho) s_x + \underbrace{t_x - s_x}_{\coloneqq d_x}
	= \rho \left( \frac{r_x + s_x - t_x}{s_x} \right) s_x + t_x - s_x
	\end{split}
	\label{equation_tile_coverage_x}
\end{equation}
Here, $d_x = t_x - s_x  = - c_x$ serves as a correction of $\mu(r_x, s_x, t_x) s_x$ regarding a potential difference of the stride $s_x$ and the tile size $t_x$ (similar to $c_x$ in \eqref{equation_tiles_in_raster_x}). \\
In order to shift the tiling scheme \wrt the raster image we compute a tiling scheme offset $\delta(r_x, s_x, t_x, \rho)$ according to \eqref{equation_tiling_offset_x}.
\begin{equation}
\begin{split}
\delta(r_x, s_x, t_x, \rho) = \frac{r_x - \kappa (r_x, s_x, t_x, \rho)}{2}
= \frac{r_x - \rho \left( \frac{r_x + s_x - t_x}{s_x} \right) s_x - t_x + s_x}{2}
\end{split}
\label{equation_tiling_offset_x}
\end{equation}
Extending \eqref{equation_tile_coverage_x} to two dimensions (denoted in the following as $\kappa_{x,y}$) yields the corresponding covered area (defined by the convex hull of the boundary pixels corresponding to the outermost tiles) and is defined according to \eqref{equation_tile_coverage}.
\begin{equation}
	\begin{split}
		\kappa_{x,y} (r_x, r_y, s_x, s_y, t_x, t_y, \rho) 
		\coloneqq \kappa (r_x, s_x, t_x, \rho) \kappa (r_y, s_y, t_y, \rho) 
	\end{split}
	\label{equation_tile_coverage}
\end{equation}
The two-dimensional offset $\delta_{x,y}$ is defined according to \eqref{equation_tile_offset}.
\begin{equation}
	\begin{split}
		\delta_{x,y}(r_x, r_y, s_x, s_y, t_x, t_y, \rho) 
		\coloneqq \left(\delta (r_x, s_x, t_x, \rho), \delta (r_y, s_y, t_y, \rho)\right) 
	\end{split}
	\label{equation_tile_offset}
\end{equation}
An example is shown in \figref{figure_tiling_type}. Note that in the top case the raster image center coincides with the center of the center tile, while in the bottom case the raster image center lies at the corner of four different tiles.

%% file: content/figures/figure_gsd_based_tiling.tex
\begin{figure*}

	\newcommand{\tileschemelabelfontsize}{\tiny}
	\newcommand{\tileschemelabelheight}{0.5cm}
   	\tikzset{custom_color_options/.style=cyan}
  	\tikzset{custom_color_fill_options/.style={black, fill=white, draw=cyan}}
 	\tikzset{custom_stroke_options/.style={custom_color_options, thick,}}
	\tikzset{custom_brace_options/.style={decorate, decoration = {brace}}}
	\tikzset{custom_arrow_options/.style={<->, custom_stroke_options}}
	\tikzset{custom_label_options/.style={custom_color_options, fill opacity = 0.5, text opacity=1.0, inner sep=1pt}}
		\begin{subfigure}{0.31\textwidth}
		\centering
		\begin{tikzpicture}
			\node[anchor=south west,inner sep=0] (image2) at (0,0) 	{\includegraphics[width=\textwidth]{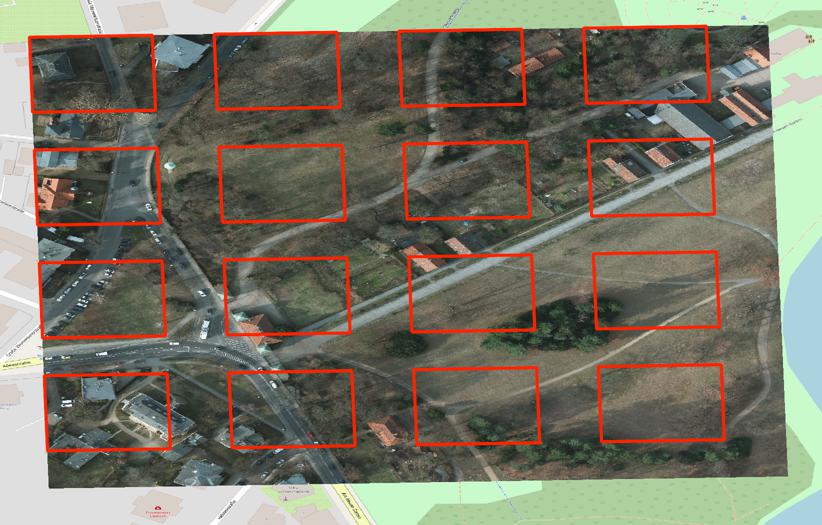}};
			\begin{scope}[x={(image2.south east)},y={(image2.north west)}]
				
				\coordinate (widthlabelpos) at (0.5,1.05);	%
				\coordinate (hbracepos) at (0.50,0.52);	%
				\coordinate (hbracedir) at (0.14,0.005);	%
				\coordinate (hbraceend) at ($(hbracepos) + (hbracedir)$);
				\coordinate (hbracecenter) at ($(hbracepos)!0.5!(hbraceend)$);
				\coordinate (hbraceanchor) at ($(hbracecenter)+(0.0, 0.05)$);
				\draw [custom_stroke_options, custom_brace_options] (hbracepos) --  ($(hbracepos) + (hbracedir)$) node {};
				\draw[custom_stroke_options] (hbraceanchor) --  (widthlabelpos) node[custom_color_fill_options, minimum height=\tileschemelabelheight] {\tileschemelabelfontsize{Tile Width}};
				
				\coordinate (heightlabelpos) at (0.85,1.05);	%
				\coordinate (vbracepos) at (0.65, 0.52);  	%
				\coordinate (vbracedir) at (0.0 , -0.15); 	%
				\coordinate (vbraceend) at ($(vbracepos) + (vbracedir)$);
				\coordinate (vbracecenter) at ($(vbracepos)!0.5!(vbraceend)$);
				\coordinate (vbraceanchor) at ($(vbracecenter)+(0.03, 0.0)$);
				\draw [custom_stroke_options, custom_brace_options] (vbracepos) -- (vbraceend) node {};
				\draw[custom_stroke_options] (vbraceanchor) --  (heightlabelpos) node[custom_color_fill_options, minimum height=\tileschemelabelheight] {\tileschemelabelfontsize{Tile Height}};
				
				\coordinate (stridelabelpos) at (0.15,1.05);
				\coordinate (arrowpos) at (0.26, 0.72);		%
				\coordinate (arrowvec) at (0.23, -0.21);		%
				\coordinate (arrowend) at ($(arrowpos) + (arrowvec)$);
				\coordinate (arrowcenter) at ($(arrowpos)!0.5!(arrowend)$);
				\coordinate (arrowanchor) at ($(arrowcenter)+(0.01, 0.02)$);
				\draw [custom_stroke_options, custom_arrow_options] (arrowpos) -- (arrowend)  node {};
				\draw[custom_stroke_options] (arrowanchor) --  (stridelabelpos) node[custom_color_fill_options, minimum height=\tileschemelabelheight] {\tileschemelabelfontsize{Tile Stride}};
			\end{scope}
		\end{tikzpicture}
		\caption{Image tiling where the stride is equal to 1.5 of the tile size.}
		\label{figure_gsd_based_tiling_b}
	\end{subfigure}
	\hfill
	\begin{subfigure}{0.31\textwidth}
		\centering
		\begin{tikzpicture}
	    \node[anchor=south west,inner sep=0] (image) at (0,0) {\includegraphics[width=\textwidth]{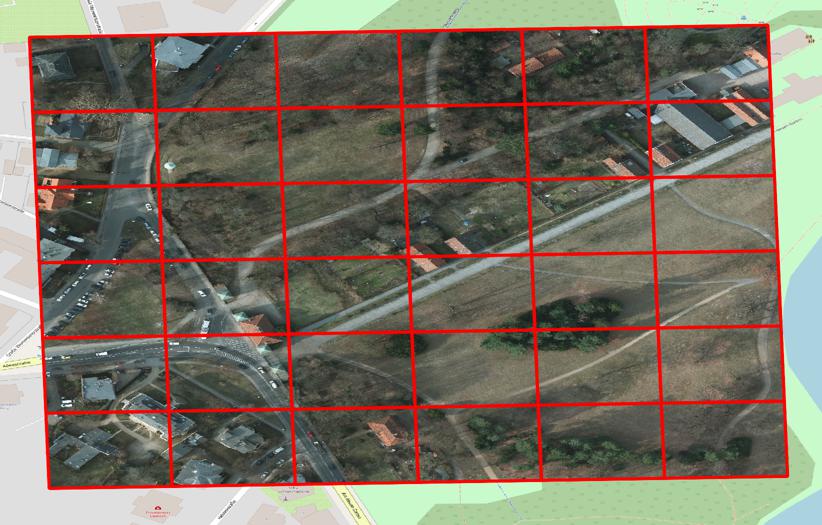}};
		\begin{scope}[x={(image.south east)},y={(image.north west)}]
			\coordinate (widthlabelpos) at (0.5,1.05);	%
			\coordinate (hbracepos) at (0.35,0.63);	%
			\coordinate (hbracedir) at (0.14,0.005);	%
			\coordinate (hbraceend) at ($(hbracepos) + (hbracedir)$);
			\coordinate (hbracecenter) at ($(hbracepos)!0.5!(hbraceend)$);
			\coordinate (hbraceanchor) at ($(hbracecenter)+(0.0, 0.05)$);
			\draw [custom_stroke_options, custom_brace_options] (hbracepos) --  ($(hbracepos) + (hbracedir)$) node {};
			\draw[custom_stroke_options] (hbraceanchor) --  (widthlabelpos) node[custom_color_fill_options, minimum height=\tileschemelabelheight] {\tileschemelabelfontsize{Tile Width}};

			\coordinate (heightlabelpos) at (0.85,1.05);	%
			\coordinate (vbracepos) at (0.50, 0.63);  	%
			\coordinate (vbracedir) at (0.0 , -0.15); 	%
			\coordinate (vbraceend) at ($(vbracepos) + (vbracedir)$);
			\coordinate (vbracecenter) at ($(vbracepos)!0.5!(vbraceend)$);
			\coordinate (vbraceanchor) at ($(vbracecenter)+(0.03, 0.0)$);
			\draw [custom_stroke_options, custom_brace_options] (vbracepos) -- (vbraceend) node {};
			\draw[custom_stroke_options] (vbraceanchor) --  (heightlabelpos) node[custom_color_fill_options, minimum height=\tileschemelabelheight] {\tileschemelabelfontsize{Tile Height}};

			\coordinate (stridelabelpos) at (0.15,1.05);
			\coordinate (arrowpos) at (0.19, 0.78);		%
			\coordinate (arrowvec) at (0.15, -0.15);		%
			\coordinate (arrowend) at ($(arrowpos) + (arrowvec)$);
			\coordinate (arrowcenter) at ($(arrowpos)!0.5!(arrowend)$);
			\coordinate (arrowanchor) at ($(arrowcenter)+(0.01, 0.02)$);
			\draw [custom_stroke_options, custom_arrow_options] (arrowpos) -- (arrowend)  node {};
			\draw[custom_stroke_options] (arrowanchor) --  (stridelabelpos) node[custom_color_fill_options, minimum height=\tileschemelabelheight] {\tileschemelabelfontsize{Tile Stride}};

		\end{scope}
		\end{tikzpicture}
		\caption{Image tiling where the stride is equal to the tile size.}
		\label{figure_gsd_based_tiling_a}
	\end{subfigure}
	\hfill
	\begin{subfigure}{0.31\textwidth}
		\centering
		\begin{tikzpicture}
			 \node[anchor=south west,inner sep=0] (image3) at (0,0) 	{\includegraphics[width=\textwidth]{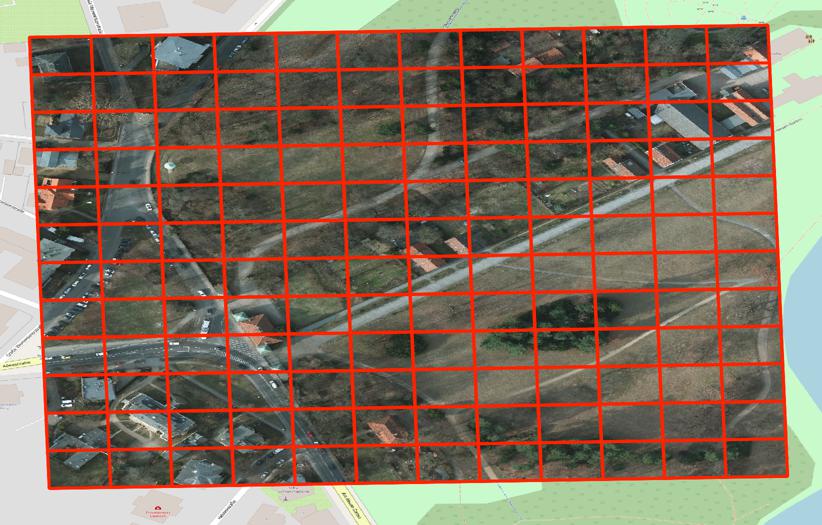}};
 	 		\begin{scope}[x={(image3.south east)},y={(image3.north west)}]
			 	\coordinate (widthlabelpos) at (0.5,1.05);	%
			 	\coordinate (hbracepos) at (0.19,0.79);	%
			 	\coordinate (hbracedir) at (0.14,0.005);	%
				\coordinate (hbraceend) at ($(hbracepos) + (hbracedir)$);
				\coordinate (hbracecenter) at ($(hbracepos)!0.5!(hbraceend)$);
				\coordinate (hbraceanchor) at ($(hbracecenter)+(0.0, 0.05)$);
				\draw [custom_stroke_options, custom_brace_options] (hbracepos) --  ($(hbracepos) + (hbracedir)$) node {};
				\draw[custom_stroke_options] (hbraceanchor) --  (widthlabelpos) node[custom_color_fill_options, minimum height=\tileschemelabelheight] {\tileschemelabelfontsize{Tile Width}};

			 	\coordinate (heightlabelpos) at (0.85,1.05);	%
			 	\coordinate (vbracepos) at (0.35, 0.78);  	%
			 	\coordinate (vbracedir) at (0.0 , -0.15); 	%
				\coordinate (vbraceend) at ($(vbracepos) + (vbracedir)$);
				\coordinate (vbracecenter) at ($(vbracepos)!0.5!(vbraceend)$);
				\coordinate (vbraceanchor) at ($(vbracecenter)+(0.03, 0.0)$);
				\draw [custom_stroke_options, custom_brace_options] (vbracepos) -- (vbraceend) node {};
				\draw[custom_stroke_options] (vbraceanchor) --  (heightlabelpos) node[custom_color_fill_options, minimum height=\tileschemelabelheight] {\tileschemelabelfontsize{Tile Height}};

			 	\coordinate (stridelabelpos) at (0.15,1.05);
			 	\coordinate (arrowpos) at (0.11, 0.86);		%
			 	\coordinate (arrowvec) at (0.075, -0.075);		%
				\coordinate (arrowend) at ($(arrowpos) + (arrowvec)$);
				\coordinate (arrowcenter) at ($(arrowpos)!0.5!(arrowend)$);
				\coordinate (arrowanchor) at ($(arrowcenter)+(0.01, 0.02)$);
				\draw [custom_stroke_options, custom_arrow_options] (arrowpos) -- (arrowend)  node {};
				\draw[custom_stroke_options] (arrowanchor) --  (stridelabelpos) node[custom_color_fill_options, minimum height=\tileschemelabelheight] {\tileschemelabelfontsize{Tile Stride}};

			 \end{scope}
		\end{tikzpicture}
		\caption{Image tiling where the stride is equal to 0.5 of the tile size.}
		\label{figure_gsd_based_tiling_c}
	\end{subfigure}
	\caption{Earth observation tiles based tiling scheme with different stride values. Depending on the stride value the tiling represents a watertight cover, exhibits non-covered areas or contains overlapping raster pixels (redundant information). A tile stride $<$ tile size is useful (\ref{figure_gsd_based_tiling_c}), since it allows to increase the number of training data samples and to fuse predictions during the inference step.}
	\label{figure_gsd_based_tiling}
\end{figure*}

%% file: content/figures/figure_training_sample_enhancement.tex
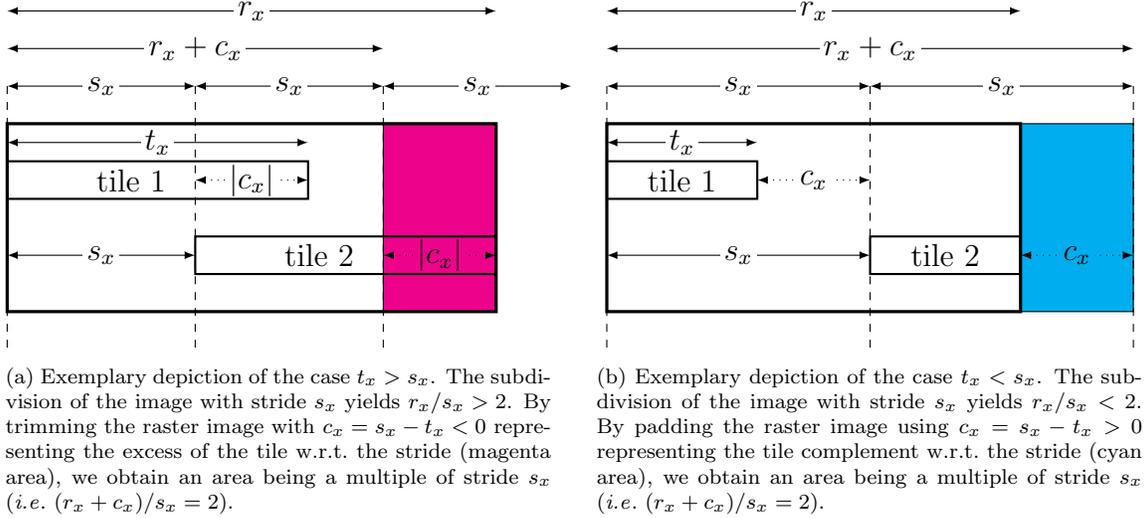
\begin{figure*}

	\tikzset{custom_number_options/.style={pos=.5}}	%

	\newcommand\drawtile{} %
	\def\drawtile[#1](#2,#3)(#4,#5)(#6){%
		\coordinate (start_coord) at (#2, #3);
		\coordinate (end_coord) at ($(start_coord) + (#4,#5)$);
		\node[draw=black,thick,rectangle,inner sep=0mm,fit=(start_coord) (end_coord)] (center_tile) {};
		\node[#1] at ($(start_coord)!0.5!(end_coord)$) {#6};
	}

	\newcommand\drawstride{} %
	\def\drawstride(#1,#2)(#3,#4){%
		\draw [draw=black, dashed] (#1,#2) -- (#3,#4);
	}

	\newcommand\drawextend{} %
	\def\drawextend[#1][#2](#3,#4)(#5,#6)(#7){%
		\draw [latex-latex, #1] (#3,#4) -- node[midway, fill=white, inner sep=0.5mm, #2]{#7} (#5,#6);
	}

	\newcommand\drawraster{} %
	\def\drawraster[#1](#2,#3)(#4,#5){%
		\coordinate (start_coord) at (#2, #3);
		\coordinate (end_coord) at ($(start_coord) + (#4,#5)$);
		\node[#1,draw=black, rectangle,inner sep=0mm,fit=(start_coord) (end_coord)] (center_tile) {};
	}

	\pgfmathsetmacro{\descry}{2.5}
	\pgfmathsetmacro{\tileh}{0.5}
	\pgfmathsetmacro{\yupperbound}{1.5}
	\pgfmathsetmacro{\ylowerbound}{-2}
	
	\newcommand\trimmingcolor{magenta}
	\newcommand\paddingcolor{cyan}

	\begin{subfigure}{0.48\textwidth}
		\centering
		\begin{tikzpicture}
			\pgfmathsetmacro{\stridex}{2.5}
			\pgfmathsetmacro{\tilew}{4}
			\pgfmathsetmacro{\rasterw}{\stridex + \tilew}
			
			\drawraster[fill=\trimmingcolor](\rasterw - \tilew + \stridex, 1)(\tilew - \stridex,-2.5)
			\drawraster[very thick](0, 1)(\rasterw,-2.5)
			\drawextend[draw=black, dotted][text=black, fill=\trimmingcolor](\rasterw - \tilew + \stridex, -0.75)(\rasterw, -0.75)($|c_x|$)
			
			\drawstride(0, \yupperbound)(0, \ylowerbound)
			
			\drawextend[draw=black,dotted][text=black](\stridex, 0.25)(\tilew, 0.25)($|c_x|$)
			\drawtile[xshift=-10](0, 0)(\tilew,\tileh)(tile $1$)
			\drawstride(\stridex, \yupperbound)(\stridex, \ylowerbound)
			\drawextend[draw=black][text=black](0, 0.75)(\tilew, 0.75)($t_x$)
			\drawextend[draw=black][text=black](0, -0.75)(\stridex, -0.75)($s_x$)
			
			\drawtile[xshift=-10](\stridex, -2.0 * \tileh)(\tilew,\tileh)(tile $2$)
			\drawstride(2.0 * \stridex, \yupperbound)(2.0 * \stridex, \ylowerbound)

			\drawextend[draw=black][text=black](0, \descry)(\rasterw, \descry)($r_x$)
			
			\drawextend[draw=black][text=black](0, \descry - 0.5)(\rasterw - \tilew + \stridex, \descry - 0.5)($r_x + c_x$)
			
			\drawextend[draw=black][text=black](0, \descry - 1)(\stridex, \descry - 1)($s_x$)
			\drawextend[draw=black][text=black](\stridex, \descry - 1)(\stridex + \stridex, \descry - 1)($s_x$)
			\drawextend[draw=black][text=black](\stridex + \stridex, \descry - 1)(\stridex + \stridex + \stridex, \descry - 1)($s_x$)
			
		\end{tikzpicture}
		\caption{Exemplary depiction of the case $t_x > s_x$. The subdivision of the image with stride $s_x$ yields $r_x / s_x > 2$. By trimming the raster image with $c_x = s_x - t_x < 0$ representing the excess of the tile \wrt the stride (magenta area), we obtain an area being a multiple of stride $s_x$ (\ie $(r_x + c_x) / s_x = 2$).}
		\label{figure_training_sample_enhancement_a}
	\end{subfigure}
	\hfill
	\begin{subfigure}{0.48\textwidth}
		\centering
		\begin{tikzpicture}
			\pgfmathsetmacro{\stridex}{3.5}
			\pgfmathsetmacro{\tilew}{2}
			\pgfmathsetmacro{\correctionx}{\stridex - \tilew}
			\pgfmathsetmacro{\rasterw}{\stridex + \tilew}
			
			\drawraster[fill=\paddingcolor](\rasterw + \correctionx, 1)(- \correctionx,-2.5)
			\drawraster[very thick](0, 1)(\rasterw,-2.5)

			\drawextend[draw=black, dotted][text=black, fill=\paddingcolor](\rasterw, -0.75)(\rasterw + \correctionx, -0.75)($c_x$)

			\drawstride(0, \yupperbound)(0, \ylowerbound)
			
			\drawtile[xshift=0](0, 0)(\tilew,\tileh)(tile $1$)
			\drawstride(\stridex, \yupperbound)(\stridex, \ylowerbound)
			\drawextend[draw=black][text=black](0, 0.75)(\tilew, 0.75)($t_x$)
			\drawextend[draw=black,dotted][text=black](\tilew, 0.25)(\stridex, 0.25)($c_x$)
			\drawextend[draw=black][text=black](0, -0.75)(\stridex, -0.75)($s_x$)
			
			\drawtile[xshift=0](\stridex, -2.0 * \tileh)(\tilew,\tileh)(tile $2$)
			\drawstride(2 * \stridex, \yupperbound)(2 * \stridex, \ylowerbound)

			\drawextend[draw=black][text=black](0, \descry)(\rasterw, \descry)($r_x$)
			\drawextend[draw=black][text=black](0, \descry - 0.5)(\rasterw - \tilew + \stridex, \descry - 0.5)($r_x + c_x$)
			\drawextend[draw=black][text=black](0, \descry - 1)(\stridex, \descry - 1)($s_x$)
			\drawextend[draw=black][text=black](\stridex, \descry - 1)(\stridex + \stridex, \descry - 1)($s_x$)

		\end{tikzpicture}
		\caption{Exemplary depiction of the case $t_x < s_x$. The subdivision of the image with stride $s_x$ yields $r_x / s_x < 2$. By padding the raster image using $c_x = s_x - t_x > 0$ representing the tile complement \wrt the stride (cyan area), we obtain an area being a multiple of stride $s_x$ (\ie $(r_x + c_x) / s_x = 2$).}
		\label{figure_training_sample_enhancement_b}
	\end{subfigure}
	\caption{Illustration of the motivation of \eqref{equation_tiles_in_raster} using a raster image (with width $r_x$) containing exactly two tiles defined by stride $s_x$ and tile width $t_x$.}
	\label{figure_training_sample_enhancement}
\end{figure*}

%% file: content/gsd_based_image_tiling.tex
\ac{eot} are an extension of plain pixel-based tiles which possess not only a pixel location and a pixel extent, but also a real world counterpart with corresponding geo-information. \figref{figure_tiling_type} shows an example comparing \ac{eot} with tiles obtained by a common web map tiling scheme using a spherical mercator projection.

\subsection{Raster Image Tiling with Earth Observation Tiles}
The \ac{gsd} $r_{g}$ of a raster image $r$ containing earth observation data represents the approximated distance between two adjacent pixel centers (in meter), \ie $\sqbr{r_{g}} = \nicefrac{\si{\meter}}{\si{\px}}$. The \ac{gsd} is typically part of the raster image meta data and allows to convert real world distances to pixel units of the corresponding raster image (and vice versa). We assume that the raster images provide sufficiently accurate \ac{gsd} values (with negligible deviations). The accuracy of the \ac{gsd} value decreases with increasing image sizes when the curvature of the earth increasingly affects the projection during the imaging process. Since the impact of the curvature is a general problem of earth observation imagery it also influences pixel-based and web map tiling schemes, and is therefore not specific to the proposed tiling scheme.\\
The \ac{gsd} allows to rewrite the tile extent in pixel (along the x-axis) $t_{e,x}$ as $t_{e,x} = \nicefrac{t'_{e,x}}{r_{g,x}}$, where $t'_{e,x}$ denotes the tile extent in meter (along the x-axis) and $r_{g,x}$ the \ac{gsd} of the raster image (also along the x-axis). \\
In order to compute tiles for multi-site or multi-sensor datasets with a consistent spatial extent we define the parameters of the tiling scheme presented in \secref{section_image_tiling} not with pixel distances, but with real world values. For each tile $t$ let the tuple of real world values $(t'_{e,x}, t'_{e,y}, t'_{o,x}, t'_{o,y})$ denote the tile extent $(t'_{e,x}, t'_{e,y})$ and the tile offset relative to the origin of the area covered by the raster image $(t'_{o,x}, t'_{o,y})$. \\ 
The \ac{gsd} of each raster image allows to express the tiling scheme parameters (\ie the tile extent and the stride) as well as the resulting tiles with their pixel-based counterparts. This allows us to rewrite $(t'_{e,x}, t'_{e,y}, t'_{o,x}, t'_{o,y})$ with the corresponding raster pixel values $(t_{e,x}, t_{e,y}, t_{o,x}, t_{o,y})$ according to \eqref{eq_raster_pixel_coordinates}.
\begin{equation}
	\begin{split}
		\rbr{t_{e,x}, t_{e,y}, t_{o,x}, t_{o,y}}
		= 
		\rbr{\frac{t'_{e,x}}{r_{g,x}}, \frac{t'_{e,y}}{r_{g,y}}, \frac{t'_{o,x}}{r_{g,x}}, \frac{t'_{o,y}}{r_{g,y}}}
		\label{eq_raster_pixel_coordinates}
	\end{split}
\end{equation}
This tiling scheme shows different beneficial properties as highlighted in \tabref{table_tile_comparison}.

\subsection{Georeferencing Earth Observation Tile Data}

In the following we will use the relation in \eqref{eq_raster_pixel_coordinates} to derive a (local) transformation matrix that maps data defined relative to the tile (such as real world objects or (resized) output information of the segmentation model) onto a tile in the raster image. By combining this local transformation matrix with the georeferencing matrix of the raster image we obtain a georeferencing matrix for the data processed by the segmentation model. \\
Pixels in the raster image are georeferenced with the corresponding georeferencing matrix $\vec{R}_{g}$ of the raster image (see \eqref{georeferencing_raster}), where $\rbr{r_{m,x},r_{m,y}}$ describes the pixel scaling, $\rbr{r_{s,x},r_{s,y}}$ the corresponding pixel skew, and $\rbr{r_{o,x},r_{o,y}}$ the offset \wrt a specific \ac{crs} - denoted in the following as $r_{crs}$.
\begin{equation}
	\vec{R}_{g}
	= 
	\begin{bmatrix} 
		r_{m,x} & r_{s,x} & r_{o,x} \\ 
		r_{s,y} & r_{m,y} & r_{o,y} \\
		0 & 0  & 1
	\end{bmatrix}
	\label{georeferencing_raster}
\end{equation}
The georeferencing matrix $\vec{T}_{g}$ of tile $t$ is obtained by applying a local transformation matrix $\vec{T}_{l}$ to the georeferencing matrix $\vec{R}_{g}$ of the raster image according to \eqref{georeferencing_tile}. \\
\begin{equation}
	\begin{split}
		\vec{T}_{g} = \vec{R}_{g} \vec{T}_{l}
		= 
		\begin{bmatrix} 
			r_{m,x} & r_{s,x}  & r_{o,x} \\ 
			r_{s,y} & r_{m,y}  & r_{o,y} \\
			0 & 0  & 1
		\end{bmatrix}
		\begin{bmatrix} 
			t_{f,x} & 0  & t_{o,x}\\ 
			0 & t_{f,y}  & t_{o,y} \\
			0 & 0  & 1
		\end{bmatrix}
		\label{georeferencing_tile}
	\end{split}
\end{equation}
The local matrix $\vec{T}_{l}$ contains an offset $(t_{o,x}, t_{o,y})$ and potentially two scaling factors $(t_{f,x}, t_{f,y})$ that depend on the data associated with the tile - or on the data that should be georeferenced. \\
In the simplest case (when the tile data is identical to the corresponding part in the raster image) the tuple $(t_{f,x}, t_{f,y})$ is equal to $(1,1)$. However, in general the raster data of the tile is not directly processed by the segmentation model, but resized to match the required input size. To georeference the adjusted tile information we define $\rbr{t_{f,x}, t_{f,y}}$ according to $\rbr{t_{f,x}, t_{f,y}} \coloneqq \rbr{\nicefrac{t_{e,x}}{n_{e,x}}, \nicefrac{t_{e,y}}{n_{e,y}}}$, where $\rbr{n_{e,x}, n_{e,y}}$ denotes the required input size of the segmentation model.

%% file: content/improvement.tex
In contrast to web map tiling schemes, the introduced \ac{eot} tiling scheme allows to create overlapping tiles. This enables us to perform an augmentation of training data and a fusion of tile predictions during inference time.

\subsection{Training Data Augmentation}
\label{section_training_enhancement}

By selecting a stride $\rbr{s_x, s_y} \coloneqq \rbr{\nicefrac{t_x}{m}, \nicefrac{t_y}{m}}$ with $m \geq 2$ we obtain an augmented set of tiles, which contains in addition to the standard tiles defined by stride $\rbr{s_x, s_y} \coloneqq \rbr{t_x, t_y}$ also overlapping tiles at stride $\rbr{s_x, s_y} \coloneqq \rbr{\nicefrac{t_x}{m}, \nicefrac{t_y}{m}}$ with $m \neq 1$. The corresponding number of tiles is given in \eqref{equation_tiles_in_raster_s_eq_t_2}.
\begin{equation}
	\begin{split}
		\nu_{x,y}(r_x, r_y, \frac{t_x}{m}, \frac{t_y}{m}, t_x, t_y, \rho)
		= \rho \rbr{m \frac{r_x}{t_x} - 1} \rho \rbr{ m \frac{r_y}{t_y} - 1}
	\end{split}
	\label{equation_tiles_in_raster_s_eq_t_2}
\end{equation}
Comparing \eqref{equation_tiles_in_raster_s_eq_t_2} with the number of tiles of a standard non-overlapping tiling scheme shown in \eqref{equation_tiles_in_raster_s_eq_t} demonstrates that we are able to increase the tile number by almost a factor of $m^2$. 
\begin{equation} 
	\begin{split}
		\nu_{x,y}(r_x, r_y, t_x, t_y, t_x, t_y, \rho)
		= \rho \rbr{ \frac{r_x}{t_x}} \rho \rbr{ \frac{r_y}{t_y}}
	\end{split}
	\label{equation_tiles_in_raster_s_eq_t}
\end{equation}
For instance, a stride of $\rbr{s_x, s_y} \coloneqq \rbr{\nicefrac{t_x}{2}, \nicefrac{t_y}{2}}$ increases the number of tiles by roughly a factor of 4 for each earth observation image. Since high values for $m$ result in small relative offsets (\ie similar tiles), it is feasible to chose a low value for $m$.

\subsection{Tile Fusion during Inference}
\label{section_improve_inference_reliability}

During inference current methods for semantic segmentation use context information to derive the label of a pixel. By tiling the images the context information is artificially limited, since relevant data from adjacent tiles is missing. Thus, the outer regions of the tiles oftentimes contain pixels with a lower prediction reliability. An empirical analysis of the prediction error in dependence of the available context information is shown in \figref{figure_reliable_area_statistic_examples}.
\input{content/figures/figure_prediction_error_statistic_examples}\\
In general, it is difficult to identify reasonable thresholds to determine reliable pixel predictions. Thus, we propose an approach defining an optimal decision criterion \wrt the chosen tiling parameters -- under the assumption that the reliability of a pixel prediction depends on the available context (\figref{figure_reliable_area_statistic_examples}). \\
To circumvent pixel predictions with lacking context information in the final segmentation result, we propose an aggregation algorithm that uses an overlapping tiling scheme to substitute pixel labels close to boundaries with pixel predictions from adjacent (overlapping) tiles containing more context information. \\
Since this problem is symmetrical for each image dimension, we derive the parameters for the aggregation scheme by analyzing each dimension individually. In the following we consider only one direction (referred to as x-direction), since the argument for the second direction is identical. \\
We define a pixel prediction $p^{(t)}$ in the current tile $t$ as reliable, if the spatial distance between the corresponding coordinate in the \emph{raster} image $p_{x}$ and the tile center $t_{c,x}$ of $t$ (also in \emph{raster} pixel coordinates) is smaller than the distance of $p_{x}$ and the tile center $t'_{c,x}$ of any other tile $t'$. The raster coordinate $p_{x}$ of the prediction $p^{(t)}$ is given by $p_{x} = t_{f,x} p_{x}^{(t)} + t_{o,x}$, where $p_{x}^{(t)}$ denotes the coordinate of the prediction relative to the tile data, $t_{f,x}$ a potential scaling factor and $t_{o,x}$ a tile specific offset (see \eqref{georeferencing_tile} for further information). More formally, the reliability assignment function $\theta(\cdot)$ is defined according to \eqref{equation_reliability}.\\
\begin{equation} 
 	\theta(p^{(t)}) = \begin{cases}
			1 & \text{if } |p_{x} - t_{c,x}| \leq \min\limits_{t'_{c,x}} |p_{x} - t'_{c,x}| \text{ with } p_{x} = t_{f,x} p_{x}^{(t)} + t_{o,x} \\
			0 & \text{otherwise} 
		\end{cases}
	\label{equation_reliability}
\end{equation}
By definition the tile centers $t_{c,x}$ and $t'_{c,x}$ of two consecutive overlapping tiles (given in raster image coordinates) differ by stride $s_x$. The interval that defines reliable pixel predictions in tile $t$ in x-direction is given by \eqref{equation_reliable_interval} as shown in \figref{figure_reliable_area_1d}. Note that the reliable area $t_{r,x}$ depends on the stride $s_x$, but not on the extent of the tile $t_{e,x}$.
\begin{equation} 
	t_{r,x} = \sqbr{t_{c,x} - \frac{s_x}{2}, t_{c,x} + \frac{s_x}{2}}
	\label{equation_reliable_interval}
\end{equation}
Other pixel labels (\ie the white tile areas in \figref{figure_reliable_area_1d}) are substituted with predictions from adjacent tiles. This aggregation scheme yields segmentation results that are exclusively derived from pixel predictions with maximum context information.
\input{content/figures/figure_reliable_area_1d} \\
The number of overlapping tiles required to fill a tile with reliable pixel predictions in x-direction $\sigma(s_x, t_{e,x})$ is given by \eqref{equation_number_of_overlapping_tiles1}. Here, $\nicefrac{t_{e,x}}{2}-\nicefrac{s_x}{2}$ defines the area of unreliable pixel predictions left and right of the tile center $t_{c,x}$.
\begin{equation}
	\sigma(s_x, t_{e,x}) = 2 \left\lceil (\frac{t_{e,x}}{2}-\frac{s_x}{2})\frac{1}{s_x} \right\rceil  = 2 \left\lceil \frac{t_{e,x} - s_x}{2 s_x} \right\rceil
	\label{equation_number_of_overlapping_tiles1}
\end{equation}
The expression for $\sigma(s_x, t_{e,x})$ shows that depending on $t_{e,x}$ and $s_x$ multiple neighbors might be required to substitute all unreliable pixel predictions. Some examples for $s_x = \nicefrac{t_{e,x}}{2}$, $s_x = \nicefrac{t_{e,x}}{3}$, $s_x = \nicefrac{t_{e,x}}{4}$ and $s_x = \nicefrac{t_{e,x}}{5}$ are shown in \figref{figure_reliable_area_1d_example}.\\
The number of neighbor tiles required for the substitution process extends to the two-dimensional case according to \eqref{equation_number_of_overlapping_tiles2}. The terms $(2 \left\lceil \frac{t_{e,x} - s_x}{2s_x} \right\rceil + 1)$ and $(2 \left\lceil \frac{t_{e,y} - s_y}{2s_y} \right\rceil + 1)$ define the total tile number in x- and y-direction (including the center tile) required to produce the final segmentation result.
\begin{equation} 
	\begin{split}
		\sigma_{x,y}(s_x, s_y, t_{e,x}, t_{e,y})
		= (2 \left\lceil \frac{t_{e,x} - s_x}{2s_x} \right\rceil + 1)
		(2 \left\lceil \frac{t_{e,y} - s_y}{2s_y} \right\rceil + 1)
		-1 
	\end{split}
	\label{equation_number_of_overlapping_tiles2}
\end{equation}
The tile substitution approach is depicted in \figref{figure_reliable_area_2d_tile_example} for $s_x=\nicefrac{t_{e,x}}{2}$ and $s_y=\nicefrac{t_{e,y}}{2}$. It illustrates the position of the reliable areas in adjacent tiles and the corresponding aggregation result.
\input{content/figures/figure_reliable_area_1d_tile_examples}
\input{content/figures/figure_reliable_area_2d_tile_example}
\input{content/figures/figure_reliable_area_2d_raster_examples} \\
\figref{figure_reliable_area_2d_raster_example} shows how different tile strides ($s_x = \nicefrac{t_x}{2}$, $s_x = \nicefrac{t_x}{3}$ and $s_x = \nicefrac{t_x}{4}$) affect the size of the corresponding reliable areas and the resulting aggregated segmentation.
By substituting $\rbr{s_x, s_y}$ with $\rbr{\nicefrac{t_{e,x}}{m}, \nicefrac{t_{e,y}}{m}}$ in \eqref{equation_tiles_in_raster} we observe that the number of tiles shows (roughly) a quadratic growth \wrt the number of strides per tile division $m$ -- see \eqref{equation_tiles_in_raster_prediction}.
\begin{equation} 
	\begin{split}
		\nu_{x,y}(r_{e,x}, r_{e,y}, \frac{t_{e,x}}{m}, \frac{t_{e,y}}{m}, t_{e,x}, t_{e,y}) = 
		\rho \rbr{\frac{m(r_{e,x} - t_{e,x})+ t_{e,x}}{t_{e,x}}} \rho \rbr{ \frac{m(r_{e,y} - t_{e,y})+ t_{e,y}}{t_{e,y}} }
	\end{split}
	\label{equation_tiles_in_raster_prediction}
\end{equation}
Because of the quadratic growth in dependence of $m$ it is feasible to chose a low value for $m$ in practice. Since a value of $m=2$ already substitutes predictions in the border areas of the tile, it is sufficient to replace the predictions with the most limited context information - see \figref{figure_reliable_area_statistic_examples}. \\

\subsection{Implementation Details}

The tiling results are subject to several pixel discretization effects, since the tiling scheme supports arbitrary parameters (\eg floating-point values for tile strides) and the tile data is resized to match the required input size of the segmentation model. To reduce resulting errors during tiling, substitution and aggregation, the pipeline performs the computations with the actual floating-point values and applies the discretization only when necessary. \\
In order to substitute the content of each tile the corresponding overlapping neighbors must be determined. Due to the discretization of the tile locations it is not suitable to use the tile stride to determine adjacent tiles with their exact locations. Instead we identify overlapping tiles based on the actual discretized tile data contained in the tiling result. A naive approach must consider $\mathcal{O}((\nu_{x,y})^2)$ different tiles. The number of tile neighbors remains constants (per tile) with increasing raster image sizes and does not affect the computation of the $\mathcal{O}$-expression. To accelerate the identification of overlapping neighbors we leverage the spatial properties of the tiles (\ie the layout of the tiles in the tiling scheme). Concretely, we represent the tiles as nodes in an R-tree \citep{Guttman1984SIGMOD} by mapping each tile on a tuple of the form $(t_{o,x}, t_{o,y}, t_{e,x}, t_{e,y})$ where $t_{o,x}$ and $t_{o,y}$ represents the tile offset. This allows to retrieve the tile neighbors in average with $\mathcal{O}(\log \nu_{x,y})$ - resulting in an overall effort of $\mathcal{O}(\nu_{x,y} \log \nu_{x,y})$.\\
The segmentations are represented with numpy arrays, which allow to efficiently perform the substitution steps using slicing operations. \\

%% file: content/figures/figure_prediction_error_statistic_examples.tex
\begin{figure*}
	\begin{subfigure}[t]{0.4\textwidth}
		\centering
		\includegraphics[width=\textwidth]{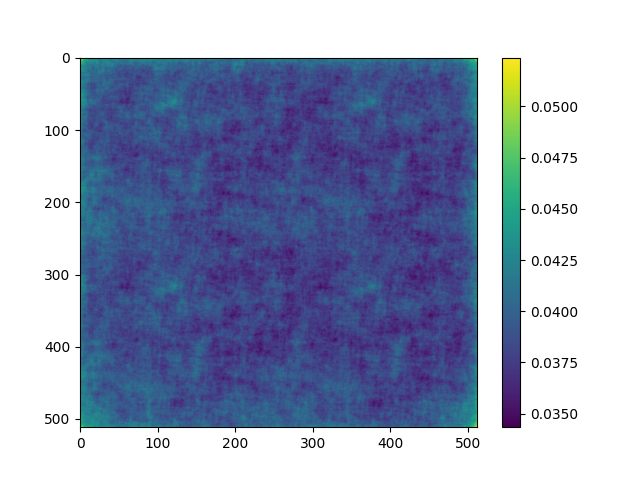}
		\caption{Average prediction error per pixel of the Open Cities AI dataset.}
	\end{subfigure}
	\hfill
	\input{content/figures/figure_prediction_error_statistic_examples_open_cities_ai_subfigure}

	\begin{subfigure}[t]{0.4\textwidth}
		\centering
		\includegraphics[width=\textwidth]{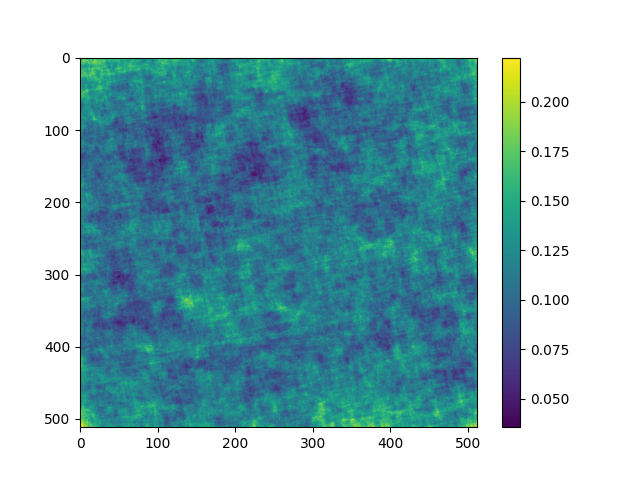}
		\caption{Average prediction error per pixel of the Potsdam dataset.}
	\end{subfigure}
	\hfill
	\input{content/figures/figure_prediction_error_statistic_examples_potsdam_subfigure}

	\caption{Pixel prediction errors relative to the tile position using UPerNet \citep{Xiao2018ECCV} with ConvNext \citep{Liu2022CVPR}. From a statistical point of view the most reliable pixels reside in the tile center.}
	\label{figure_reliable_area_statistic_examples}
\end{figure*}

%% file: content/figures/figure_prediction_error_statistic_examples_open_cities_ai_subfigure.tex
\begin{subfigure}[t]{0.56\textwidth}
	\centering
	\begin{tikzpicture}%
		\begin{axis}[
			width=7.5cm,
			height=5cm,
			label style={font=\scriptsize},
			tick label style={font=\scriptsize}, 
			axis y line*=left,
			scaled ticks=false,
			yticklabel style={/pgf/number format/fixed},
			xlabel=Distance to Tile Center {in px},
			ylabel=Average Prediction Error {in \%},
		]%
			\addplot[color=orange,mark=*,only marks,fill opacity=0.5,draw opacity=0,] coordinates {%
		(362,0.05144940114563444)%
		(360,0.04929796138787633)%
		(358,0.04712337750004822)%
		(356,0.0458331886825204)%
		(354,0.04489729792281435)%
		(352,0.04421975351501478)%
		(350,0.043784950189067126)%
		(348,0.04327474975409359)%
		(346,0.042785697933883246)%
		(344,0.042388722196047286)%
		(342,0.04212940461725396)%
		(340,0.04193449321031329)%
		(338,0.04178230222830319)%
		(336,0.04167491567248556)%
		(334,0.04157393473642235)%
		(332,0.04155087369091018)%
		(330,0.04132923873648717)%
		(328,0.04116401499929872)%
		(326,0.04095019133615392)%
		(324,0.040768246530719414)%
		(322,0.040558186822724406)%
		(320,0.04038024960554223)%
		(318,0.04029824701583092)%
		(316,0.040232735319795286)%
		(314,0.04011931235732803)%
		(312,0.03998139428918606)%
		(310,0.03990378124643029)%
		(308,0.039845365203293064)%
		(306,0.03971336741493148)%
		(304,0.03962982863185097)%
		(302,0.039617908022208506)%
		(300,0.039481718657293906)%
		(298,0.039440461433171266)%
		(296,0.03933870216245371)%
		(294,0.03932238534643839)%
		(292,0.039308035551878555)%
		(290,0.03926676794724762)%
		(288,0.03920486524043377)%
		(286,0.03912161331801253)%
		(284,0.03906117229865763)%
		(282,0.03902908607628842)%
		(280,0.03906957731394919)%
		(278,0.039064481003191084)%
		(276,0.03910228813371409)%
		(274,0.039081314027228746)%
		(272,0.03905041563000256)%
		(270,0.039030085615747896)%
		(268,0.03904249268944231)%
		(266,0.039085154869967345)%
		(264,0.03905203470401401)%
		(262,0.03916271247259537)%
		(260,0.039224046641972286)%
		(258,0.03931637051583528)%
		(256,0.03953082688600025)%
		(254,0.03930826789373888)%
		(252,0.039216696221953236)%
		(250,0.03909928207086989)%
		(248,0.03900621844251588)%
		(246,0.03888052381086927)%
		(244,0.038846993448841766)%
		(242,0.03880182376341434)%
		(240,0.03877720616694247)%
		(238,0.03872300312866841)%
		(236,0.03864104592497309)%
		(234,0.038644124657773234)%
		(232,0.03866231710040387)%
		(230,0.03866659476257318)%
		(228,0.03864415176763278)%
		(226,0.03858413717856516)%
		(224,0.03854171602594326)%
		(222,0.03850972502379244)%
		(220,0.038483496559704185)%
		(218,0.03847428685452433)%
		(216,0.03844044149988293)%
		(214,0.03841171079184733)%
		(212,0.038412669138128956)%
		(210,0.03837576320471417)%
		(208,0.03831007313087842)%
		(206,0.03824687847178982)%
		(204,0.03821601492901913)%
		(202,0.03818186005778173)%
		(200,0.038190104348765616)%
		(198,0.03817332144668784)%
		(196,0.038207899623132695)%
		(194,0.038259145547647894)%
		(192,0.03834074692698445)%
		(190,0.03834285867532082)%
		(188,0.038322266320289296)%
		(186,0.038260819855996275)%
		(184,0.03819491294407057)%
		(182,0.038160811453325595)%
		(180,0.03814982069454545)%
		(178,0.038158137868089685)%
		(176,0.038162283351813896)%
		(174,0.03813339760860772)%
		(172,0.0381148075521573)%
		(170,0.03815277577919475)%
		(168,0.03816563110088535)%
		(166,0.038146866036815774)%
		(164,0.03810448218625248)%
		(162,0.03813090748067126)%
		(160,0.03814996912195012)%
		(158,0.03816657564249675)%
		(156,0.03815283109858239)%
		(154,0.038131687709176035)%
		(152,0.03809290912617464)%
		(150,0.03811784670953284)%
		(148,0.0381276534479883)%
		(146,0.03813646997740392)%
		(144,0.038075851129140595)%
		(142,0.03807217268669907)%
		(140,0.038077107653978697)%
		(138,0.038032558313743574)%
		(136,0.03805874362408438)%
		(134,0.03802613028734021)%
		(132,0.03798488229842779)%
		(130,0.03795636618916242)%
		(128,0.03796919127736612)%
		(126,0.03805862335821347)%
		(124,0.03806265017892708)%
		(122,0.038064253583339334)%
		(120,0.03806178472469066)%
		(118,0.038066425534937734)%
		(116,0.03809342396625828)%
		(114,0.038155791089285705)%
		(112,0.038154577252746666)%
		(110,0.03815401737851897)%
		(108,0.038052478113087373)%
		(106,0.03802191930366859)%
		(104,0.03808318828439896)%
		(102,0.038101559871507366)%
		(100,0.03811066446357996)%
		(98,0.038070746403848964)%
		(96,0.038024223283383174)%
		(94,0.03798898549809862)%
		(92,0.03803551227337214)%
		(90,0.03813287775940459)%
		(88,0.03811094759896669)%
		(86,0.038033728624554806)%
		(84,0.0380675339692807)%
		(82,0.038085011619160544)%
		(80,0.03799299549710515)%
		(78,0.03800168129672153)%
		(76,0.038070068853786845)%
		(74,0.03803125063955856)%
		(72,0.03799755477234457)%
		(70,0.03792530366578778)%
		(68,0.03788175979548988)%
		(66,0.03788936873222187)%
		(64,0.03793355219141751)%
		(62,0.03802797546722191)%
		(60,0.03809189538733273)%
		(58,0.0380507338617908)%
		(56,0.03800167059701286)%
		(54,0.038115736173851764)%
		(52,0.03813874048786953)%
		(50,0.038214834295670456)%
		(48,0.03829870088343565)%
		(46,0.038264677270507745)%
		(44,0.038179256109085895)%
		(42,0.038251549309584965)%
		(40,0.03832010011702712)%
		(38,0.038469001085616565)%
		(36,0.03840962959663158)%
		(34,0.03826150616855361)%
		(32,0.03823879349425247)%
		(30,0.03817677712957448)%
		(28,0.03810223515781745)%
		(26,0.03804980702510077)%
		(24,0.03797139474539242)%
		(22,0.03771600139871052)%
		(20,0.037699408144867365)%
		(18,0.03790853522453943)%
		(16,0.03795126799200849)%
		(14,0.03797641778461084)%
		(12,0.03799554030416729)%
		(10,0.038007869004223815)%
		(8,0.03820035101930604)%
		(6,0.03841292744315222)%
		(4,0.03855543432439804)%
		(2,0.039027946537059544)%
		(0,0.03916681131747961)%
			};%
			\label{plot_prediction_error_open_cities_ai}%
		\end{axis}%
		\begin{axis}[
			width=7.5cm,
			height=5cm,
			label style={font=\scriptsize},
			tick label style={font=\scriptsize},
			axis y line*=right,
			ylabel=Number Averaged Error Values,
			legend style={
				nodes={scale=0.75, transform shape},
				at={(0.0,1.0)},anchor=north west,
			}
		]%
			\addlegendimage{/pgfplots/refstyle=plot_prediction_error_open_cities_ai}%
			\addlegendentry{Average Prediction Error}%
			\addplot[color=cyan,mark=*,only marks,fill opacity=0.5,draw opacity=0,] coordinates {%
				(362,5)%
				(360,36)%
				(358,72)%
				(356,80)%
				(354,120)%
				(352,168)%
				(350,204)%
				(348,240)%
				(346,276)%
				(344,280)%
				(342,320)%
				(340,364)%
				(338,408)%
				(336,444)%
				(334,484)%
				(332,480)%
				(330,532)%
				(328,584)%
				(326,612)%
				(324,668)%
				(322,700)%
				(320,712)%
				(318,756)%
				(316,812)%
				(314,848)%
				(312,896)%
				(310,936)%
				(308,940)%
				(306,992)%
				(304,1048)%
				(302,1104)%
				(300,1148)%
				(298,1188)%
				(296,1216)%
				(294,1252)%
				(292,1336)%
				(290,1388)%
				(288,1452)%
				(286,1452)%
				(284,1528)%
				(282,1572)%
				(280,1696)%
				(278,1684)%
				(276,1816)%
				(274,1804)%
				(272,1944)%
				(270,1980)%
				(268,2084)%
				(266,2176)%
				(264,2280)%
				(262,2380)%
				(260,2544)%
				(258,2732)%
				(256,3138)%
				(254,3204)%
				(252,3184)%
				(250,3112)%
				(248,3076)%
				(246,3096)%
				(244,3076)%
				(242,3104)%
				(240,3012)%
				(238,2968)%
				(236,2928)%
				(234,2972)%
				(232,2904)%
				(230,2884)%
				(228,2880)%
				(226,2800)%
				(224,2828)%
				(222,2784)%
				(220,2804)%
				(218,2728)%
				(216,2712)%
				(214,2684)%
				(212,2688)%
				(210,2612)%
				(208,2632)%
				(206,2548)%
				(204,2632)%
				(202,2472)%
				(200,2532)%
				(198,2472)%
				(196,2508)%
				(194,2432)%
				(192,2416)%
				(190,2356)%
				(188,2360)%
				(186,2332)%
				(184,2328)%
				(182,2300)%
				(180,2256)%
				(178,2232)%
				(176,2196)%
				(174,2208)%
				(172,2156)%
				(170,2152)%
				(168,2072)%
				(166,2100)%
				(164,2048)%
				(162,2076)%
				(160,2008)%
				(158,1968)%
				(156,1948)%
				(154,1936)%
				(152,1908)%
				(150,1880)%
				(148,1876)%
				(146,1808)%
				(144,1848)%
				(142,1764)%
				(140,1768)%
				(138,1724)%
				(136,1744)%
				(134,1640)%
				(132,1676)%
				(130,1624)%
				(128,1644)%
				(126,1536)%
				(124,1604)%
				(122,1504)%
				(120,1512)%
				(118,1476)%
				(116,1448)%
				(114,1444)%
				(112,1432)%
				(110,1360)%
				(108,1348)%
				(106,1328)%
				(104,1308)%
				(102,1296)%
				(100,1264)%
				(98,1228)%
				(96,1192)%
				(94,1188)%
				(92,1160)%
				(90,1140)%
				(88,1088)%
				(86,1056)%
				(84,1076)%
				(82,1032)%
				(80,1020)%
				(78,968)%
				(76,960)%
				(74,916)%
				(72,920)%
				(70,860)%
				(68,880)%
				(66,796)%
				(64,824)%
				(62,768)%
				(60,772)%
				(58,720)%
				(56,724)%
				(54,656)%
				(52,648)%
				(50,636)%
				(48,608)%
				(46,556)%
				(44,576)%
				(42,524)%
				(40,496)%
				(38,472)%
				(36,452)%
				(34,432)%
				(32,412)%
				(30,368)%
				(28,344)%
				(26,324)%
				(24,312)%
				(22,276)%
				(20,248)%
				(18,224)%
				(16,204)%
				(14,168)%
				(12,156)%
				(10,120)%
				(8,108)%
				(6,64)%
				(4,56)%
				(2,20)%
				(0,5)%
			};%
			\addlegendentry{Number Averaged Error Values}%
		\end{axis}%
	\end{tikzpicture}
	\caption{Average prediction error over the pixel distance to the tile center using the Open Cities AI dataset.}
\end{subfigure}

%% file: content/figures/figure_prediction_error_statistic_examples_potsdam_subfigure.tex
\begin{subfigure}[t]{0.56\textwidth}
	\centering
	\begin{tikzpicture}%
		\begin{axis}[
			width=7.5cm,
			height=5cm,
			label style={font=\scriptsize},
			tick label style={font=\scriptsize},
			axis y line*=left,
			scaled ticks=false,
			yticklabel style={/pgf/number format/fixed},
			xlabel=Distance to Tile Center {in px},
			ylabel=Average Prediction Error {in \%},
		]%
			\addplot[color=orange,mark=*,only marks,fill opacity=0.5,draw opacity=0,] coordinates {%
(362,0.18492063492063493)%
(360,0.19036596119929455)%
(358,0.18463403880070542)%
(356,0.17931547619047622)%
(354,0.17169312169312162)%
(352,0.16326530612244894)%
(350,0.15806878306878308)%
(348,0.15696097883597876)%
(346,0.1549919484702093)%
(344,0.15052437641723354)%
(342,0.14407242063492057)%
(340,0.14211582068724923)%
(338,0.1406493152816681)%
(336,0.13781638781638797)%
(334,0.13570772661681746)%
(332,0.13468915343915341)%
(330,0.13327962764052997)%
(328,0.1332422809306372)%
(326,0.13222325967424034)%
(324,0.13182563444539516)%
(322,0.13014172335600907)%
(320,0.12881219903691796)%
(318,0.12929369278575584)%
(316,0.12953514739229013)%
(314,0.12908992213237502)%
(312,0.12747572987528352)%
(310,0.12598358431691786)%
(308,0.12606382978723404)%
(306,0.12441196236559142)%
(304,0.12222828062522736)%
(302,0.12125819530710859)%
(300,0.12058929262762016)%
(298,0.11979250173694606)%
(296,0.11804250208855481)%
(294,0.11784002738475598)%
(292,0.11790110255679143)%
(290,0.1186445039110745)%
(288,0.11955868205868202)%
(286,0.1196297389479207)%
(284,0.11941379955123428)%
(282,0.11854275213053853)%
(280,0.11901954177897597)%
(278,0.11995956339780593)%
(276,0.11933606041535608)%
(274,0.11893763418153722)%
(272,0.1186985596707826)%
(270,0.1183521725188398)%
(268,0.11818122962556818)%
(266,0.11720756010737732)%
(264,0.11632553606237979)%
(262,0.11637655062024979)%
(260,0.11575633922332271)%
(258,0.11627187478212565)%
(256,0.1180473357815608)%
(254,0.1172096387452185)%
(252,0.11679054997208291)%
(250,0.1151137940180359)%
(248,0.11388734080541618)%
(246,0.11353230999548809)%
(244,0.11384605857947849)%
(242,0.11382905825560481)%
(240,0.11381458293808891)%
(238,0.11308053095451963)%
(236,0.11266290441495375)%
(234,0.11199902796470773)%
(232,0.11168366566093897)%
(230,0.11138355018382819)%
(228,0.11098434744268133)%
(226,0.11010345804988719)%
(224,0.11028743180440552)%
(222,0.10957597837985834)%
(220,0.10905763874736857)%
(218,0.10832111436950219)%
(216,0.10739306784660847)%
(214,0.10674514465498162)%
(212,0.10623641817838327)%
(210,0.1069590291450939)%
(208,0.10574371592608713)%
(206,0.10440183399367238)%
(204,0.10428426689824959)%
(202,0.10358073919967371)%
(200,0.10299437047067594)%
(198,0.1022708301227747)%
(196,0.1024056353003748)%
(194,0.10316154970760474)%
(192,0.10414695679596586)%
(190,0.10471912307651095)%
(188,0.10506456820016379)%
(186,0.10472821203953507)%
(184,0.10347460862925013)%
(182,0.1023636991028319)%
(180,0.10387819430372855)%
(178,0.10532585196563923)%
(176,0.10576228062567111)%
(174,0.10627300437083274)%
(172,0.10583237624054172)%
(170,0.10645873900985632)%
(168,0.10616419991420227)%
(166,0.10607331821617784)%
(164,0.10576326884920886)%
(162,0.10573600024467301)%
(160,0.10628122430911499)%
(158,0.10707026713124478)%
(156,0.107668426713603)%
(154,0.10762659058113759)%
(152,0.10806212771621737)%
(150,0.10856973995272014)%
(148,0.10926659220902417)%
(146,0.1095571709509771)%
(144,0.10962731052016854)%
(142,0.10962414066155651)%
(140,0.10989908784026518)%
(138,0.11077044893750329)%
(136,0.11026694699286547)%
(134,0.11009727061556433)%
(132,0.11000776603401984)%
(130,0.10951550160294067)%
(128,0.10864664194956236)%
(126,0.10699301421957713)%
(124,0.10545808098800688)%
(122,0.10445689800743096)%
(120,0.10411417653481268)%
(118,0.10412633888243747)%
(116,0.10449552749276603)%
(114,0.10547750956338339)%
(112,0.10659694510951577)%
(110,0.10660014005602284)%
(108,0.10729887899769251)%
(106,0.10708608242493851)%
(104,0.10703970681034958)%
(102,0.1060742455418388)%
(100,0.10639880952381009)%
(98,0.10769867121658708)%
(96,0.10755566208586387)%
(94,0.1083152958152964)%
(92,0.10826491516146755)%
(90,0.10853174603174644)%
(88,0.10946982959850624)%
(86,0.10978835978836021)%
(84,0.10927081489349219)%
(82,0.10976913375169271)%
(80,0.10977279800809286)%
(78,0.10891381345926864)%
(76,0.10978835978836038)%
(74,0.10953853885076645)%
(72,0.11010610766045606)%
(70,0.11160022148394315)%
(68,0.11228805916305974)%
(66,0.11120084549732817)%
(64,0.11113037447988937)%
(62,0.11108527612433906)%
(60,0.11020643144995493)%
(58,0.11078593474426869)%
(56,0.11154411119880792)%
(54,0.11231489547038342)%
(52,0.11363413678228505)%
(50,0.11468628331835895)%
(48,0.11500104427736028)%
(46,0.1144227475162728)%
(44,0.11447310405643746)%
(42,0.11604113655640381)%
(40,0.11527137736815166)%
(38,0.1147767016411084)%
(36,0.11381514257620451)%
(34,0.111221340388007)%
(32,0.10940630297426417)%
(30,0.10774672187715666)%
(28,0.10739664082687345)%
(26,0.10782872819909864)%
(24,0.1065450752950753)%
(22,0.10561881757533942)%
(20,0.10504672299027143)%
(18,0.10859552154195015)%
(16,0.1122782446311858)%
(14,0.11056783824640962)%
(12,0.10673585673585666)%
(10,0.10922619047619035)%
(8,0.10593033509700167)%
(6,0.10280257936507932)%
(4,0.0991354875283446)%
(2,0.09623015873015872)%
(0,0.09999999999999999)%
			};%
			\label{plot_one}%
		\end{axis}%
		\begin{axis}[
			width=7.5cm,
			height=5cm,
			label style={font=\scriptsize},
			tick label style={font=\scriptsize},
			axis y line*=right,
			ylabel=Number Averaged Error Values,
			legend style={
				nodes={scale=0.75, transform shape},
				at={(0.0,1.0)},anchor=north west,
			}
		]%
			\addlegendimage{/pgfplots/refstyle=plot_one}%
			\addlegendentry{Average Prediction Error}%
			\addplot[color=cyan,mark=*,only marks,fill opacity=0.5,draw opacity=0,] coordinates {%
				(362,5)%
				(360,36)%
				(358,72)%
				(356,80)%
				(354,120)%
				(352,168)%
				(350,204)%
				(348,240)%
				(346,276)%
				(344,280)%
				(342,320)%
				(340,364)%
				(338,408)%
				(336,444)%
				(334,484)%
				(332,480)%
				(330,532)%
				(328,584)%
				(326,612)%
				(324,668)%
				(322,700)%
				(320,712)%
				(318,756)%
				(316,812)%
				(314,848)%
				(312,896)%
				(310,936)%
				(308,940)%
				(306,992)%
				(304,1048)%
				(302,1104)%
				(300,1148)%
				(298,1188)%
				(296,1216)%
				(294,1252)%
				(292,1336)%
				(290,1388)%
				(288,1452)%
				(286,1452)%
				(284,1528)%
				(282,1572)%
				(280,1696)%
				(278,1684)%
				(276,1816)%
				(274,1804)%
				(272,1944)%
				(270,1980)%
				(268,2084)%
				(266,2176)%
				(264,2280)%
				(262,2380)%
				(260,2544)%
				(258,2732)%
				(256,3138)%
				(254,3204)%
				(252,3184)%
				(250,3112)%
				(248,3076)%
				(246,3096)%
				(244,3076)%
				(242,3104)%
				(240,3012)%
				(238,2968)%
				(236,2928)%
				(234,2972)%
				(232,2904)%
				(230,2884)%
				(228,2880)%
				(226,2800)%
				(224,2828)%
				(222,2784)%
				(220,2804)%
				(218,2728)%
				(216,2712)%
				(214,2684)%
				(212,2688)%
				(210,2612)%
				(208,2632)%
				(206,2548)%
				(204,2632)%
				(202,2472)%
				(200,2532)%
				(198,2472)%
				(196,2508)%
				(194,2432)%
				(192,2416)%
				(190,2356)%
				(188,2360)%
				(186,2332)%
				(184,2328)%
				(182,2300)%
				(180,2256)%
				(178,2232)%
				(176,2196)%
				(174,2208)%
				(172,2156)%
				(170,2152)%
				(168,2072)%
				(166,2100)%
				(164,2048)%
				(162,2076)%
				(160,2008)%
				(158,1968)%
				(156,1948)%
				(154,1936)%
				(152,1908)%
				(150,1880)%
				(148,1876)%
				(146,1808)%
				(144,1848)%
				(142,1764)%
				(140,1768)%
				(138,1724)%
				(136,1744)%
				(134,1640)%
				(132,1676)%
				(130,1624)%
				(128,1644)%
				(126,1536)%
				(124,1604)%
				(122,1504)%
				(120,1512)%
				(118,1476)%
				(116,1448)%
				(114,1444)%
				(112,1432)%
				(110,1360)%
				(108,1348)%
				(106,1328)%
				(104,1308)%
				(102,1296)%
				(100,1264)%
				(98,1228)%
				(96,1192)%
				(94,1188)%
				(92,1160)%
				(90,1140)%
				(88,1088)%
				(86,1056)%
				(84,1076)%
				(82,1032)%
				(80,1020)%
				(78,968)%
				(76,960)%
				(74,916)%
				(72,920)%
				(70,860)%
				(68,880)%
				(66,796)%
				(64,824)%
				(62,768)%
				(60,772)%
				(58,720)%
				(56,724)%
				(54,656)%
				(52,648)%
				(50,636)%
				(48,608)%
				(46,556)%
				(44,576)%
				(42,524)%
				(40,496)%
				(38,472)%
				(36,452)%
				(34,432)%
				(32,412)%
				(30,368)%
				(28,344)%
				(26,324)%
				(24,312)%
				(22,276)%
				(20,248)%
				(18,224)%
				(16,204)%
				(14,168)%
				(12,156)%
				(10,120)%
				(8,108)%
				(6,64)%
				(4,56)%
				(2,20)%
				(0,5)%
			};%
			\addlegendentry{Number Averaged Error Values}%
		\end{axis}%
	\end{tikzpicture}
	\caption{Average prediction error over the pixel distance to the tile center using the Potsdam dataset.}
\end{subfigure}

%% file: content/figures/figure_reliable_area_1d.tex
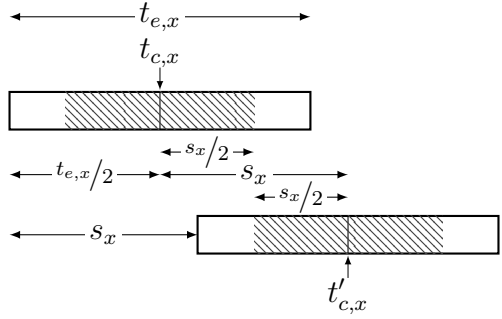
\begin{figure}

	\tikzset{custom_number_options/.style={pos=.5}}	%
	
	\newcommand\drawcenterline{} %
	\def\drawcenterline(#1,#2)(#3,#4){%
		\draw [draw=black] (#1,#2) -- (#3,#4);
	}
	
	\newcommand\drawtile{} %
	\def\drawtile[#1](#2,#3)(#4,#5)(#6)(#7){%
		\coordinate (tile_start_coord) at (#2, #3);
		\coordinate (tile_end_coord) at ($(tile_start_coord) + (#4,#5)$);
		\node[#1,draw=black,thick,rectangle,inner sep=0mm,fit=(tile_start_coord) (tile_end_coord)] (center_tile) {};
		\node at ($(tile_start_coord)!0.5!(tile_end_coord)$) {#7};
		\def\currenttilewhalf{#2+#4/2}
		\def\currentstridexhalf{#6/2}
		\drawcenterline(\currenttilewhalf, #3)(\currenttilewhalf, #3+#5)

		\coordinate (reliable_start_coord) at (\currenttilewhalf - \currentstridexhalf, #3);
		\coordinate (reliable_end_coord) at (\currenttilewhalf + \currentstridexhalf, #3+#5);

		\node[pattern=north west lines, pattern color=darkgray, rectangle,inner sep=0mm,fit=(reliable_start_coord) (reliable_end_coord)] (reliable_area) {};

	}
	
	\newcommand\drawstride{} %
	\def\drawstride(#1,#2)(#3,#4){%
		\draw [draw=black, dashed] (#1,#2) -- (#3,#4);
	}

	\newcommand\drawextend{} %
	\def\drawextend[#1][#2](#3,#4)(#5,#6)(#7){%
		\draw [latex-latex, #1] (#3,#4) -- node[midway, fill=white, inner sep=0.5mm, #2]{#7} (#5,#6);
	}
	
	\newcommand\drawraster{} %
	\def\drawraster[#1](#2,#3)(#4,#5){%
		\coordinate (start_coord) at (#2, #3);
		\coordinate (end_coord) at ($(start_coord) + (#4,#5)$);
		\node[#1,draw=black, rectangle,inner sep=0mm,fit=(start_coord) (end_coord)] (center_tile) {};
	}
	
	\newcommand\drawboundingbox{}
	\def\drawboundingbox(#1,#2)(#3,#4){
		\path[step=1.0,black,thin,xshift=0.5cm,yshift=0.5cm] (#1,#2) grid (#3,#4);
	}

	\newcommand\graywhite{gray!50!white}
	
	\centering
	\begin{tikzpicture}
		\pgfmathsetmacro{\tilew}{4}
		\pgfmathsetmacro{\tilewhalf}{2}
		\pgfmathsetmacro{\stridex}{2.5}
		\pgfmathsetmacro{\stridexhalf}{1.25}
		\pgfmathsetmacro{\rasterw}{\stridex + \tilew}
		
		\pgfmathsetmacro{\tileh}{0.5}
		\pgfmathsetmacro{\tilehhalf}{0.25}
		
		\pgfmathsetmacro{\descriptionshifth}{-0.3}
		\pgfmathsetmacro{\descriptionshifthdouble}{-0.6}
		\pgfmathsetmacro{\descriptionshifthtripple}{-0.9}
		
		\pgfmathsetmacro{\secondtiley}{-1.65}
		
		\pgfmathsetmacro{\yupperbound}{1.75}
		\pgfmathsetmacro{\ylowerbound}{-2}
		
		\drawextend[draw=black][text=black](0, 1.5)(\tilew, 1.5)($t_{e,x}$)

		\drawtile[](0, 0)(\tilew,\tileh)(\stridex)()
		\node[pin={[inner sep=0mm, pin distance=0.3cm, pin edge={latex-, black}]90:$t_{c,x}$}, inner sep=0mm] at (\tilewhalf, 0.5) {};

		\drawextend[draw=black][text=black](\tilewhalf, \descriptionshifth)(\tilewhalf + \stridexhalf, \descriptionshifth)($\nicefrac{s_{x}}{2}$)
		
		\drawextend[draw=black][text=black](0, \descriptionshifthdouble)(\tilewhalf, \descriptionshifthdouble)($\nicefrac{t_{e,x}}{2}$)
		\drawextend[draw=black][text=black](\tilewhalf, \descriptionshifthdouble)(\tilewhalf + \stridex, \descriptionshifthdouble)($s_x$)

		\drawextend[draw=black][text=black](\stridexhalf + \tilewhalf, \descriptionshifthtripple)(\stridex + \tilewhalf, \descriptionshifthtripple)($\nicefrac{s_{x}}{2}$)
		
		\drawextend[draw=black][text=black](0, \secondtiley + \tilehhalf)(\stridex, \secondtiley + \tilehhalf)($s_x$)
		\drawtile[](\stridex, \secondtiley)(\tilew, \tileh)(\stridex)()
		\node[pin={[inner sep=0mm, pin distance=0.3cm, pin edge={latex-, black}]270:$t'_{c,x}$}, inner sep=0mm] at (\stridex + \tilewhalf, \secondtiley) {};
		
		\drawboundingbox(0, \ylowerbound - 0.75)(1, \yupperbound - 0.5)
	\end{tikzpicture}
	\caption{Relation of reliable pixel predictions of a tile (highlighted with diagonal line patterns) \wrt the stride $s_x$ for $t_{e,x} > s_x > \nicefrac{t_e,x}{2}$.}
	\label{figure_reliable_area_1d}
\end{figure} %

%% file: content/figures/figure_reliable_area_1d_tile_examples.tex
\begin{figure*}

	\tikzset{custom_number_options/.style={pos=.5}}	%
	
	\newcommand\drawcenterline{} %
	\def\drawcenterline(#1,#2)(#3,#4){%
		\draw [draw=black] (#1,#2) -- (#3,#4);
	}
	
	\newcommand\drawtile{} %
	\def\drawtile[#1](#2,#3)(#4,#5)(#6)(#7){%
		\coordinate (tile_start_coord) at (#2, #3);
		\coordinate (tile_end_coord) at ($(tile_start_coord) + (#4,#5)$);
		\node[#1,draw=black,thick,rectangle,inner sep=0mm,fit=(tile_start_coord) (tile_end_coord)] (center_tile) {};
		\node at ($(tile_start_coord)!0.5!(tile_end_coord)$) {#7};
		\def\currenttilewhalf{#2+#4/2}
		\def\currentstridexhalf{#6/2}
		\drawcenterline(\currenttilewhalf, #3)(\currenttilewhalf, #3+#5)
		
		\coordinate (reliable_start_coord) at (\currenttilewhalf - \currentstridexhalf, #3);
		\coordinate (reliable_end_coord) at (\currenttilewhalf + \currentstridexhalf, #3+#5);
		
		\node[pattern=north west lines, pattern color=darkgray, rectangle,inner sep=0mm,fit=(reliable_start_coord) (reliable_end_coord)] (reliable_area) {};

	}
	
	\newcommand\drawstride{} %
	\def\drawstride(#1,#2)(#3,#4){%
		\draw [draw=black, dashed] (#1,#2) -- (#3,#4);
	}

	\newcommand\drawextend{} %
	\def\drawextend[#1][#2](#3,#4)(#5,#6)(#7){%
		\draw [latex-latex, #1] (#3,#4) -- node[midway, fill=white, inner sep=0.5mm, #2]{#7} (#5,#6);
	}
	
	\newcommand\drawraster{} %
	\def\drawraster[#1](#2,#3)(#4,#5){%
		\coordinate (start_coord) at (#2, #3);
		\coordinate (end_coord) at ($(start_coord) + (#4,#5)$);
		\node[#1,draw=black, rectangle,inner sep=0mm,fit=(start_coord) (end_coord)] (center_tile) {};
	}
	
	\newcommand\drawboundingbox{}
	\def\drawboundingbox(#1,#2)(#3,#4){
		\path[step=1.0,black,thin,xshift=0.5cm,yshift=0.5cm] (#1,#2) grid (#3,#4);
	}
	
	\pgfmathsetmacro{\tileh}{0.5}
	\pgfmathsetmacro{\yupperbound}{1.75}
	\pgfmathsetmacro{\ylowerbound}{-2}
	
	\newcommand\graywhite{gray!50!white}
	
		\begin{subfigure}{0.22\textwidth}
		
		\centering
		\begin{tikzpicture}
			\pgfmathsetmacro{\tilew}{1.5}
			\pgfmathsetmacro{\tilewhalf}{0.5 * \tilew}
			\pgfmathsetmacro{\stridex}{0.5 * \tilew}
			
			\drawtile[](\stridex, 0.25)(\tilew,\tileh)(\stridex)()
			\drawtile[fill=gray](2 * \stridex, -0.5)(\tilew, \tileh)(\stridex)()
			\drawtile[](3 * \stridex, -1.25)(\tilew, \tileh)(\stridex)()

			\draw [-latex, dashed] (\stridex + \tilewhalf, 0.25) -- (\stridex +\tilewhalf, 0);
			\draw [-latex, dashed] (3*\stridex + \tilewhalf, -0.75) -- (3 * \stridex +\tilewhalf, -0.5);
			
			\drawboundingbox(0, \ylowerbound - 0.75)(3,\yupperbound - 0.5)	
		\end{tikzpicture}
		\caption{Reliable pixel predictions for stride $s_x=\nicefrac{t_{e,x}}{2}$.}
		\label{figure_reliable_area_1d_example_a}
	\end{subfigure}
	\hfill
	\begin{subfigure}{0.22\textwidth}
		
		\centering
		\begin{tikzpicture}
			\pgfmathsetmacro{\tilew}{1.5}
			\pgfmathsetmacro{\tilewhalf}{0.5 * \tilew}
			\pgfmathsetmacro{\stridex}{0.33 * \tilew}
			
			\drawtile[](\stridex, 0.25)(\tilew,\tileh)(\stridex)()
			\drawtile[fill=gray](2 * \stridex, -0.5)(\tilew, \tileh)(\stridex)()
			\drawtile[](3 * \stridex, -1.25)(\tilew, \tileh)(\stridex)()

			\draw [-latex, dashed] (\stridex + \tilewhalf, 0.25) -- (\stridex +\tilewhalf, 0);
			\draw [-latex, dashed] (3*\stridex + \tilewhalf, -0.75) -- (3 * \stridex +\tilewhalf, -0.5);
			
			\drawboundingbox(0, \ylowerbound - 0.75)(3,\yupperbound - 0.5)	
		\end{tikzpicture}
		\caption{Reliable pixel predictions for stride $s_x=\nicefrac{t_{e,x}}{3}$.}
		\label{figure_reliable_area_1d_example_b}
	\end{subfigure}
	\hfill
	\begin{subfigure}{0.22\textwidth}
	\centering
	\begin{tikzpicture}
		\pgfmathsetmacro{\tilew}{1.5}
		\pgfmathsetmacro{\tilewhalf}{0.5 * \tilew}
		\pgfmathsetmacro{\stridex}{0.25 * \tilew}
		
		\drawtile[](0, 1)(\tilew,\tileh)(\stridex)()
		\drawtile[](\stridex, 0.25)(\tilew,\tileh)(\stridex)()
		\drawtile[fill=gray](2 * \stridex, -0.5)(\tilew, \tileh)(\stridex)()
		\drawtile[](3 * \stridex, -1.25)(\tilew, \tileh)(\stridex)()
		\drawtile[](4 * \stridex, -2)(\tilew, \tileh)(\stridex)()
		
		\draw [-latex, dashed] (\tilewhalf, 1) -- (\tilewhalf, 0);
		\draw [-latex, dashed] (\stridex + \tilewhalf, 0.25) -- (\stridex +\tilewhalf, 0);
		\draw [-latex, dashed] (3*\stridex + \tilewhalf, -0.75) -- (3 * \stridex +\tilewhalf, -0.5);
		\draw [-latex, dashed] (4*\stridex + \tilewhalf, -1.75) -- (4 * \stridex +\tilewhalf, -0.5);
		
		\drawboundingbox(0, \ylowerbound - 0.75)(3,\yupperbound - 0.5)	
	\end{tikzpicture}
	\caption{Reliable pixel predictions for stride $s_x=\nicefrac{t_{e,x}}{4}$.}
	\label{figure_reliable_area_1d_example_c}
	\end{subfigure}
	\hfill
	\begin{subfigure}{0.22\textwidth}
	\centering
	\begin{tikzpicture}
		\pgfmathsetmacro{\tilew}{1.5}
		\pgfmathsetmacro{\tilewhalf}{0.5 * \tilew}
		\pgfmathsetmacro{\stridex}{0.2 * \tilew}
		
		\drawtile[](0, 1)(\tilew,\tileh)(\stridex)()
		\drawtile[](\stridex, 0.25)(\tilew,\tileh)(\stridex)()
		\drawtile[fill=gray](2 * \stridex, -0.5)(\tilew, \tileh)(\stridex)()
		\drawtile[](3 * \stridex, -1.25)(\tilew, \tileh)(\stridex)()
		\drawtile[](4 * \stridex, -2)(\tilew, \tileh)(\stridex)()
		
		\draw [-latex, dashed] (\tilewhalf, 1) -- (\tilewhalf, 0);
		\draw [-latex, dashed] (\stridex + \tilewhalf, 0.25) -- (\stridex +\tilewhalf, 0);
		\draw [-latex, dashed] (3*\stridex + \tilewhalf, -0.75) -- (3 * \stridex +\tilewhalf, -0.5);
		\draw [-latex, dashed] (4*\stridex + \tilewhalf, -1.75) -- (4 * \stridex +\tilewhalf, -0.5);
		
		\drawboundingbox(0, \ylowerbound - 0.75)(3,\yupperbound - 0.5)	
	\end{tikzpicture}
	\caption{Reliable pixel predictions for stride $s_x=\nicefrac{t_{e,x}}{5}$.}
	\label{figure_reliable_area_1d_example_d}
	\end{subfigure}
	\caption{Pixel substitution of non-reliable pixel predictions with reliable predictions of adjacent tiles (highlighted with diagonal line pattern). The area extent depends on stride $s_x$.}
	\label{figure_reliable_area_1d_example}
\end{figure*}
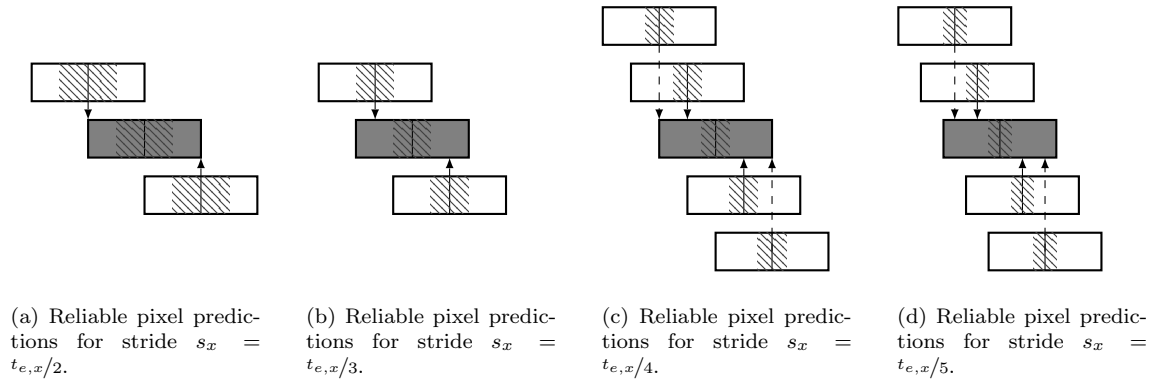

%% file: content/figures/figure_reliable_area_2d_tile_example.tex
\begin{figure*}

	\tikzset{custom_number_options/.style={pos=.5}}	%

	\pgfmathsetmacro{\xwidth}{1}
	\pgfmathsetmacro{\ywidth}{1}
	\pgfmathsetmacro{\xshift}{0.5}
	\pgfmathsetmacro{\yshift}{0.5}
	
	\newcommand\graywhite{gray!50!white}
	
	\newcommand\drawcentertile{} %
	\def\drawcentertile(#1,#2){%
		\coordinate (square_size) at (\xwidth, \ywidth);

		\coordinate (start_coord) at (#1, #2);
		\coordinate (end_coord) at ($(start_coord) + (\xwidth, \ywidth)$);
		\node[draw=black, fill=\graywhite, opacity=0.5, thick,rectangle,inner sep=0mm, fit=(start_coord) (end_coord)] (center_tile) {};
	}
	
	\newcommand\drawreliablearea{} %
	\def\drawreliablearea[#1][#2](#3,#4)(#5){%

		\coordinate (reliable_rel_pos) at (0.25, 0.25);
		\coordinate (reliable_size) at (0.5, 0.5);

		\coordinate (start_coord) at ($(#3, #4) + (reliable_rel_pos)$);
		\coordinate (end_coord) at ($(start_coord) + (reliable_size)$);
		\node[draw=black,dashed,rectangle,inner sep=0mm, fit=(start_coord) (end_coord)] (reliable_area) {};
		\node[#1] at (#2) {\footnotesize #5};
	
	}
	
	\newcommand\drawtile{} %
	\def\drawtile[#1][#2](#3,#4)(#5){%
		\coordinate (square_size) at (\xwidth, \ywidth);
		\draw [draw=black] (#3, #4) rectangle ++(square_size) ;
		\drawreliablearea[#1][#2](#3, #4)(#5)
	}
	\newcommand\drawboundingbox{
		\path[step=1.0,black,thin,xshift=0.5cm,yshift=0.5cm] (-1,-1) grid (1,1);
	}
	\begin{subfigure}[t]{0.23\textwidth}
	\centering
	\begin{tikzpicture}
		\drawtile[anchor=west, inner sep=1pt][reliable_area.center](0, 0)()		%
		\drawcentertile(0, 0)						%
		\drawboundingbox
	\end{tikzpicture}
	\caption{Reference tile (gray) with reliable region (dashed rectangle).}
	\label{figure_reliable_area_2d_tile_example_a}
	\end{subfigure}
	\hfill
	\begin{subfigure}[t]{0.23\textwidth}
		\centering
		\begin{tikzpicture}
			\drawtile[anchor=west, inner sep=1pt][reliable_area.west](-\xshift, \yshift)(1)		%
			\drawtile[anchor=east, inner sep=1pt][reliable_area.east](\xshift, \yshift)(3)		%
			\drawtile[anchor=west, inner sep=1pt][reliable_area.west](-\xshift, -\yshift)(7)	%
			\drawtile[anchor=east, inner sep=1pt][reliable_area.east](\xshift, -\yshift)(5)		%
			\drawcentertile(0, 0)						%
			\drawboundingbox
		\end{tikzpicture}
		\caption{Reference tile (gray) with diagonal overlapping tiles and their reliable regions (dashed rectangles).}
		\label{figure_reliable_area_2d_tile_example_b}
	\end{subfigure}
	\hfill
	\begin{subfigure}[t]{0.23\textwidth}
		\centering
		\begin{tikzpicture}
			\drawtile[anchor=south, inner sep=1pt][reliable_area.center](0, \yshift)(2)		%
			\drawtile[anchor=east, inner sep=1pt][reliable_area.east](\xshift, 0)(4)		%
			\drawtile[anchor=south, inner sep=1pt][reliable_area.south](0, -\yshift)(6)		%
			\drawtile[anchor=west, inner sep=1pt][reliable_area.west](-\xshift, 0)(8)		%
			\drawcentertile(0, 0)			%
			\drawboundingbox
		\end{tikzpicture}
		\caption{Reference tile (gray) with horizontal and vertical overlapping tiles and their reliable regions (dashed rectangles).}
		\label{figure_reliable_area_2d_tile_example_c}
	\end{subfigure}
	\hfill
	\begin{subfigure}[t]{0.23\textwidth}
		\centering
		\begin{tikzpicture}
			\drawreliablearea[anchor=south east, inner sep=1pt][reliable_area.south east](-\xshift, \yshift)(1)		%
			\drawreliablearea[anchor=south west, inner sep=1pt][reliable_area.south west](\xshift, \yshift)(3)		%
			\drawreliablearea[anchor=north east, inner sep=1pt][reliable_area.north east](-\xshift, -\yshift)(7)	%
			\drawreliablearea[anchor=north west, inner sep=1pt][reliable_area.north west](\xshift, -\yshift)(5)		%
			\drawreliablearea[anchor=south, inner sep=1pt][reliable_area.south](0, \yshift)(2)			%
			\drawreliablearea[anchor=west, inner sep=1pt][reliable_area.west](\xshift, 0)(4)			%
			\drawreliablearea[anchor=north, inner sep=1pt][reliable_area.north](0, -\yshift)(6)			%
			\drawreliablearea[anchor=east, inner sep=1pt][reliable_area.east](-\xshift, 0)(8)			%
			\drawcentertile(0, 0)			%
			\drawboundingbox	
		\end{tikzpicture}
		\caption{Reference tile (gray) with reliable areas (dashed rectangles) of horizontal, vertical and diagonal overlapping tiles.}
		\label{figure_reliable_area_2d_tile_example_d}
	\end{subfigure}
	\caption{Tile fusion during inference time using adjacent overlapping tiles with $\rbr{s_x, s_y} \coloneqq \rbr{ \nicefrac{t_{e,y}}{2}, \nicefrac{t_{e,x}}{2}}$. The pixel predictions close to the center of a tile show usually a higher reliability because of additional context information. }
	\label{figure_reliable_area_2d_tile_example}
\end{figure*}
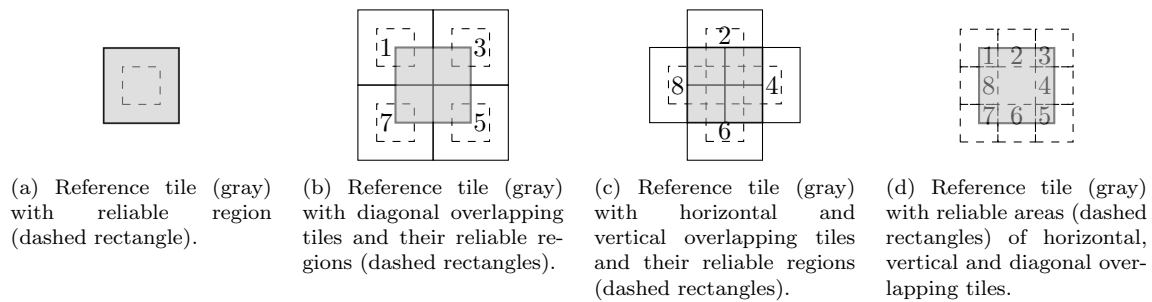

%% file: content/figures/figure_reliable_area_2d_raster_examples.tex
\begin{figure*}
	\centering
	\newcommand{\tilingtypefigurewidth}{0.32\textwidth}	%
	\begin{subfigure}[t]{\tilingtypefigurewidth}
		\centering
		\includegraphics[width=\textwidth]{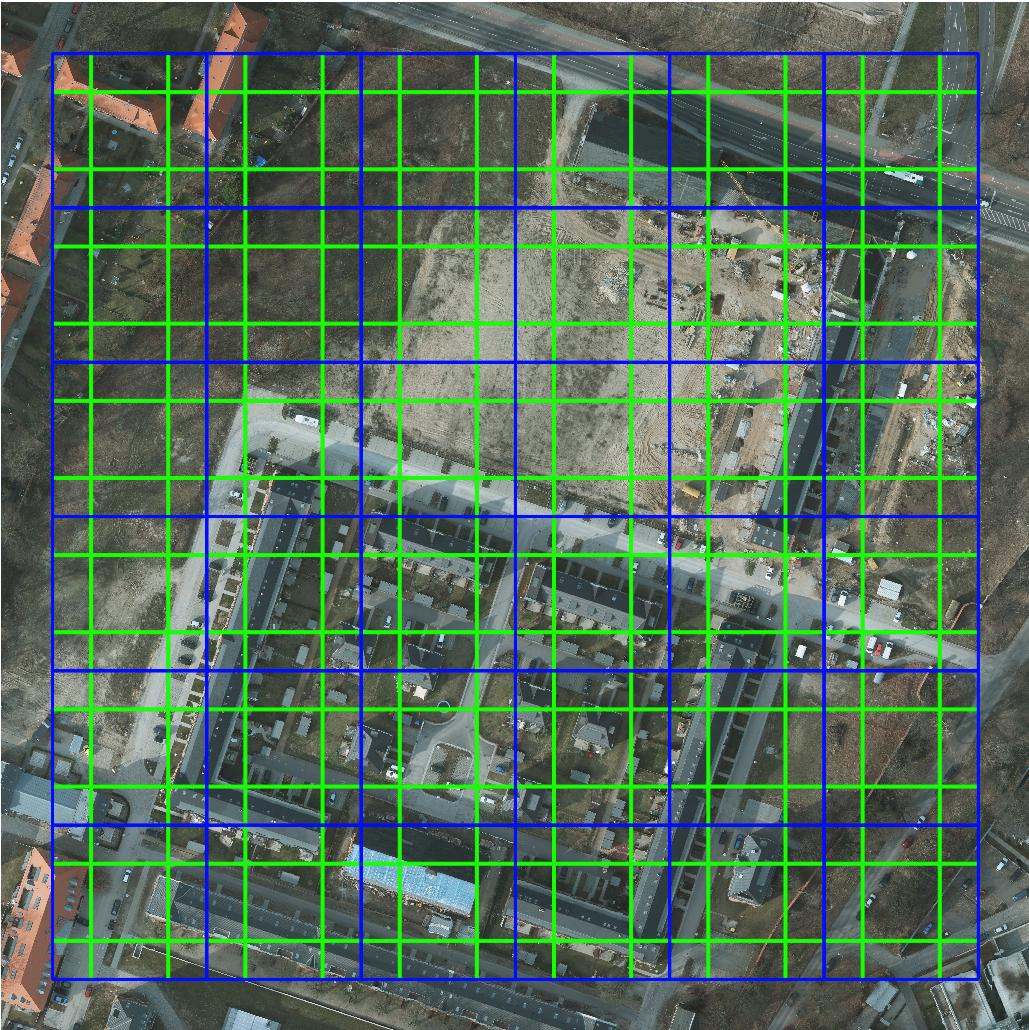} \\
		\caption{Reliable areas for $s_x = \nicefrac{t_x}{2}$.}
		\label{figure_reliable_area_2d_raster_example_a}
	\end{subfigure}
	\hfil
		\begin{subfigure}[t]{\tilingtypefigurewidth}
			\centering
			\includegraphics[width=\textwidth]{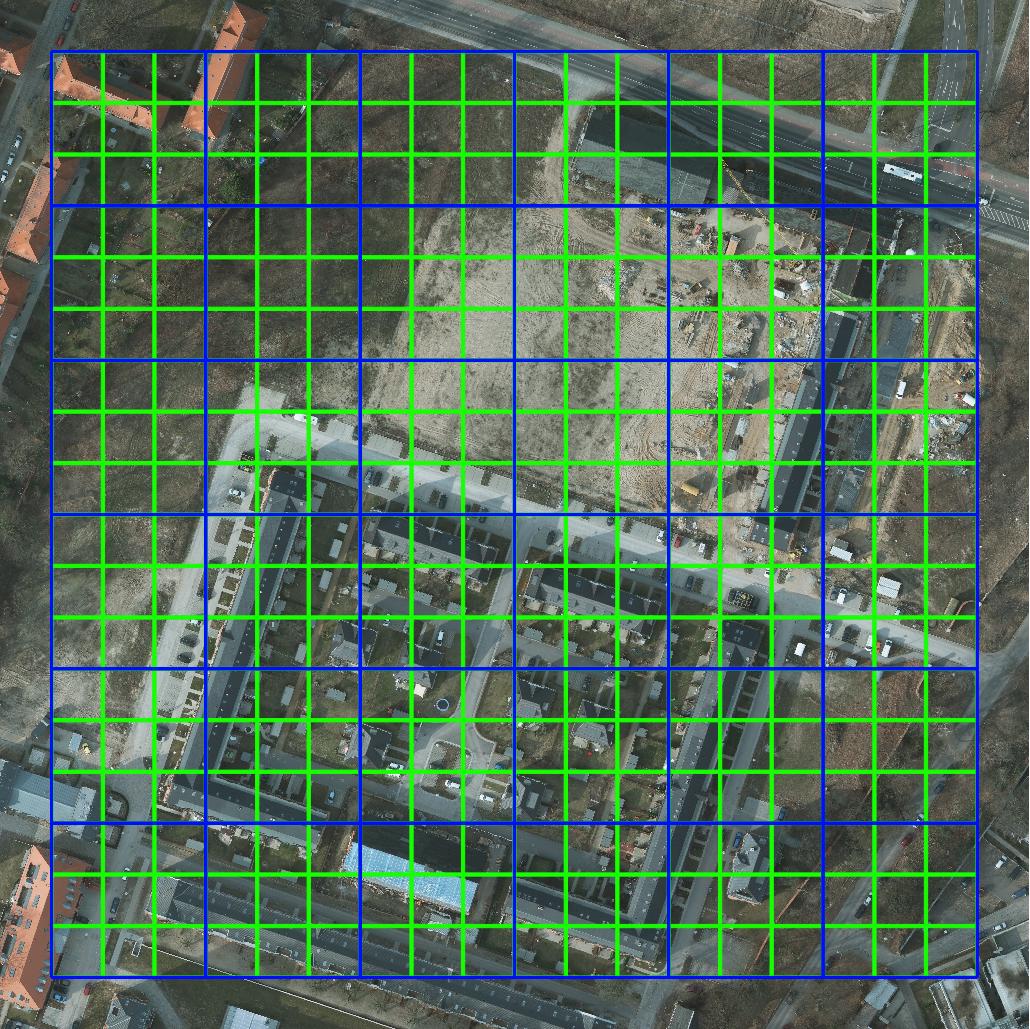} \\
			\caption{Reliable areas for $s_x = \nicefrac{t_x}{3}$}
			\label{figure_tiling_type_b}
		\end{subfigure}
		\hfill
	\begin{subfigure}[t]{\tilingtypefigurewidth}
		\centering
		\includegraphics[width=\textwidth]{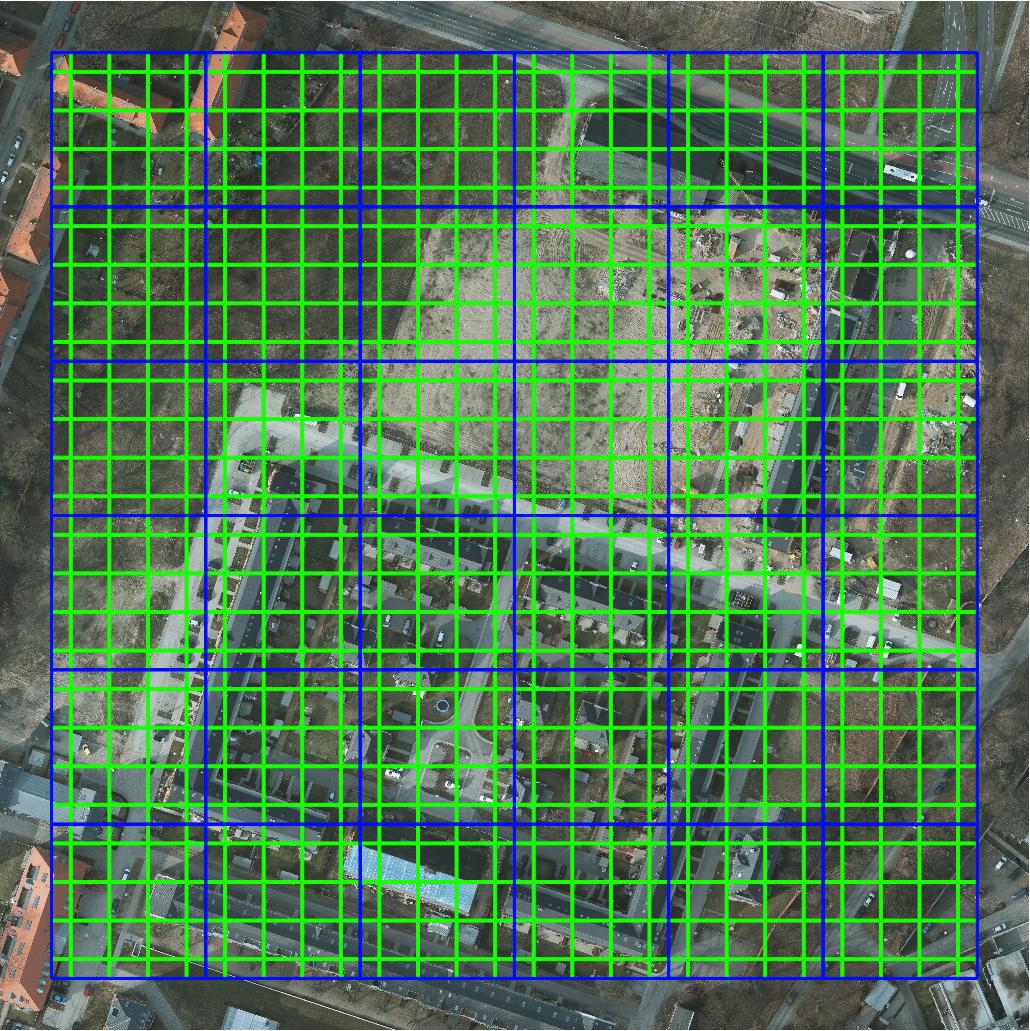} \\
		\caption{Reliable areas for $s_x = \nicefrac{t_x}{4}$.}
		\label{figure_reliable_area_2d_raster_example_c}
	\end{subfigure}
	\caption{Adjacent tiles (blue rectangles) with reliable areas (green rectangles) for different stride values $s_x$ on the Potsdam data set.}
	\label{figure_reliable_area_2d_raster_example}
\end{figure*}

%% file: content/experiments.tex
We utilize the Open Cities AI dataset \citep{OpenCitiesAI} as well as the ISPRS dataset \citep{RottensteinerPRS2014} for evaluation and subdivide each dataset in a training and a test set using 80\% of the imagery as training images. Since the Open Cities AI dataset contains more data (in terms of number of images as well as covered area) and contains several locations of multiple cities, we use this dataset for the majority of our evaluations. To create suitable tiles for our experiments regarding tile size (\secref{section_tile_size_selection}) and tile specific data augmentation (\secref{section_tiling_specific_data_augmentation}) we follow the approach described in \secref{section_tile_number_optimal_coverage} to determine an optimal coverage of the raster data. For the tile fusion evaluation (\secref{section_tile_fusion}) we deviate from this tiling approach for the test set to obtain a consistent tile alignment across different stride values, which is required to perform a fair comparison using different stride values. For the quantitative evaluation we use the standard \ac{iou} metric -- also known as the Jaccard Index \citep{JaccardIndex}.

\subsection{Tile Size Selection}
\label{section_tile_size_selection}
As discussed in \secref{section_introduction}, the tiling process represents a trade-off between detail and context information. In order to determine a suitable tile size we conducted a set of experiments in \tabref{table_experiments_tile_size_evaluation_open_cities} that show the segmentation results in dependence of the tile size. On the Open Cities AI dataset \citep{OpenCitiesAI} we achieve the best results with $75m$ tiles. At the latitudes of the Open Cities AI dataset, standard web map tiles (at zoom level $19$) cover an average area of $76m$. These tile extents are (coincidentally) similar to best tile size determined in the empirical evaluation. Since the tile size does not only affect the quality of the segmentation result but also the processing time of large areas, it is important to note that there is only a flat performance decrease for larger tiles.

\input{content/tables/table_experiments_tile_size_evaluation_open_cities}
\input{content/tables/table_experiments_augmentation_evaluation_open_cities}
\input{content/tables/table_experiments_inference_fusion}

\subsection{Tiling Specific Data Augmentation}
\label{section_tiling_specific_data_augmentation}
In contrast to common mercator tiling schemes the proposed tiling scheme allows to augment the number of training samples by computing overlapping tiles as depicted in \figref{figure_gsd_based_tiling_c}. \tabref{table_experiments_augmentation} shows the corresponding augmentation evaluation. We observe that the augmentation improves the segmentation results reported in \tabref{table_experiments_tile_size_evaluation_open_cities} across different tile sizes. For an evaluation of further earth observation specific data augmentation adjustments we refer to the supplementary material.

\subsection{Tile Fusion During Inference Time}
\label{section_tile_fusion}

We define a set of adjacent non-overlapping \emph{base} tiles (\ie $\rbr{s_x, s_y} = \rbr{t_{e,x}, t_{e,y}}$) that provide a watertight cover of the raster images - see \figref{figure_gsd_based_tiling_a}. Complementary to the base tiles we create a set of overlapping \emph{auxiliary} tiles (\ie $\rbr{s_x, s_y} \coloneqq \rbr{\nicefrac{t_x}{m}, \nicefrac{t_y}{m}}$ with $m > 1$) which are used to substitute base tile pixels with limited context information.\\
We ensure a consistent alignment of the base and the auxiliary tiles (corresponding to different stride values) by defining the center of the raster images as the origin of the tiling schemes. This decision is based on the fact that \secref{section_tile_fusion} compares only results of tiles with a fixed size (in contrast to \secref{section_tile_size_selection} and \secref{section_tiling_specific_data_augmentation}). \\
The evaluation reported in \tabref{table_experiments_inference_fusion} is based on the (fused) segmentation results of the base tiles. An analysis of the obtained \ac{iou}-scores shows an improvement of the segmentation results highlighting the importance of context information for accurate semantic segmentations. By performing the proposed tile fusion method we obtain consistent category labels at tile boundaries as shown in \figref{figure_experiments_inference_fusion_open_cities_potsdam}.
\input{content/figures/figure_experiments_inference_fusion_qualitative_open_cities_potsdam}

%% file: content/tables/table_experiments_tile_size_evaluation_open_cities.tex
\begin{table}[t]
	\tablefontsizestiny
	\newcommand{\grouprowspacing}{\baselineskip}	%
	\newcommand{\tileinfocol}{2cm}
	\newcommand{\tileextent}{3.8cm}
	\newcommand{\numtilecol}{1cm}
	\newcommand{\buildcol}{\numtilecol}
	\newcommand{\backcol}{\numtilecol}
	\newcommand{\mioucol}{2cm}
	\centering
	\begin{tabular}{ccccc}%
		\makecell[c]{Tile Info \\ Size / Train Stride}&\makecell[c]{Train \\ Tiles} &\makecell[c]{IoU \\ Background}&\makecell[c]{IoU\\Building}&\makecell[c]{mIoU} \\%
		\hline%
		25m / 25m&183 k& 95.61&72.75&84.18\\
		Mercator Zoom Level of 20 ($\sim$ 38m*) & 78 k & 95.63&73.20&84.41 \\[\grouprowspacing]%
		40m / 40m&71 k& 95.67&73.25 & 84.46 \\
		50m / 50m&45 k& 95.66&73.19&84.42 \\
		60m / 60m&31 k& 95.68&73.18&84.43 \\
		70m / 70m&22 k& 95.65&73.32&84.49 \\
		75m / 75m&20 k& \textbf{95.73}&73.36& \textbf{84.55} \\
		Mercator Zoom Level of 19 ($\sim$ 76m*) &19 k & 95.62& \textbf{73.45} &84.53 \\[\grouprowspacing]%
		80m / 80m&17 k& 95.67&73.25&84.46 \\
		90m / 90m&14 k & 95.65&73.22&84.44 \\
		100m / 100m&11 k & 95.61&73.11&84.36 \\
		110m / 110m&9 k & 95.67&73.13&84.40 \\
		120m / 120m&8 k & 95.64&73.18&84.41 \\
		150m / 150m&5 k & 95.66&72.99&84.33 \\
		200m / 200m&3 k & 95.40&72.46&83.93 \\
	\end{tabular}
	\caption{Tile size evaluation on the Open Cities AI dataset using UPerNet \citep{Xiao2018ECCV} with ConvNext \citep{Liu2022CVPR} as architecture (batch size: 4, Iterations 320k). The dataset is split into training and test data using a ratio of 80/20. The extracted tiles are resized to 512px$\times$512px to match the required dimension of the model. *The sizes corresponding to $\sim$ 38m for a zoom level of 20 and $\sim$ 76m for a zoom level of 19 are only valid for the Open Cities AI dataset.}
	\label{table_experiments_tile_size_evaluation_open_cities}
\end{table}

%% file: content/tables/table_experiments_augmentation_evaluation_open_cities.tex
\begin{table}[t]
	\tablefontsizestiny
	\newcommand{\grouprowspacing}{\baselineskip}	%
	\newcommand{\tileinfocol}{2cm}
	\newcommand{\tileextent}{3.8cm}
	\newcommand{\numtilecol}{1cm}
	\newcommand{\buildcol}{\numtilecol}
	\newcommand{\backcol}{\numtilecol}
	\newcommand{\mioucol}{2cm}
	\centering
	\begin{tabular}{cccccc}%
		\makecell[c]{Tile Info \\ Size / Train Stride}&\makecell[c]{Train \\ Tiles} &\makecell[c]{IoU \\ Background}&\makecell[c]{IoU\\Building}&\makecell[c]{mIoU} &\makecell[c]{Augmentation \\ Improvement} \\%
		\hline%
		25m / 12.5m&729 k& 95.65&72.94&84.30& +0.12\\[\grouprowspacing]%
		40m / 20m&282 k& \underline{\textbf{95.74}}&73.44 & 84.59& +0.13\\[\grouprowspacing]%
		50m / 25m&179 k& 95.71&73.27&84.49 & +0.07 \\[\grouprowspacing]%
		60m / 30m&124 k& 95.72&73.39&84.56 & +0.13\\[\grouprowspacing]%
		70m / 35m&90 k& 95.68&73.35&84.52 & +0.03\\[\grouprowspacing]%
		75m / 37.5m&78 k& 95.7& \underline{\textbf{73.51}}&\underline{\textbf{84.61}}& +0.06 \\[\grouprowspacing]%
		80m / 40m&69 k& 95.66&73.32&84.49 & +0.03\\[\grouprowspacing]%
		90m / 45m&54 k& 95.68 & 73.35 & 84.51 & +0.07 \\[\grouprowspacing]%
		100m / 50m& 43 k& 95.63 k & 73.14 & 84.39 & +0.03 \\[\grouprowspacing]%
	\end{tabular}
	\caption{Training data augmentation on the Open Cities AI dataset using UPerNet \citep{Xiao2018ECCV} with ConvNext \citep{Liu2022CVPR} as architecture (batch size: 4, Iterations 320k). The dataset is split into training and test data using a ratio of 80/20. The extracted tiles are resized to 512px$\times$512px to match the required dimension of the model.}
	\label{table_experiments_augmentation}
\end{table}

%% file: content/tables/table_experiments_inference_fusion.tex
\begin{table*}[t]
	\newcommand{\tileinfocol}{2cm}
	\newcommand{\tileextent}{3.8cm}
	\newcommand{\numtilecol}{1cm}
	\newcommand{\buildcol}{\numtilecol}
	\newcommand{\backcol}{\numtilecol}
	\newcommand{\mioucol}{2cm}
	\centering
	\begin{subtable}{\textwidth}
		\centering
		\resizebox{\textwidth}{!}{%
		\begin{tabular}{ccccccc}%
			\makecell[c]{Tile Info \\ Size / Train Stride / Test Stride}&  \makecell[c]{Checkpoint} & \makecell[c]{Tile \\ Substitution}  &\makecell[c]{Background \\ IoU}&\makecell[c]{Building \\ IoU}&\makecell[c]{mIoU} & \makecell[c]{Substitution \\ Improvement} \\ %
			\hline%
			75.0m/ 75.0m / 75.0m & 320 k & \xmark & 95.68 & 73.44 & 84.56 & - \\%
			75.0m/ 75.0m / 37.5m & 320 k & \cmark & 95.75 & 73.78 & 84.77 & +0.21 \\%
			75.0m/ 75.0m / 25m 	 & 320 k & \cmark & 95.76 & 73.81 & 84.78 & +0.22 \\%
		\end{tabular}
		}
		\caption{Open Cities AI dataset.}
	\end{subtable} \\
	\subtableoffset	%
	\begin{subtable}{\textwidth}
		\centering
		\resizebox{\textwidth}{!}{%
			\begin{tabular}{ccccccc}%
				\makecell[c]{Tile Info \\ Size / Train Stride / Test Stride}&  \makecell[c]{Checkpoint} & \makecell[c]{Tile \\ Substitution}  &\makecell[c]{Background \\ IoU}&\makecell[c]{Building \\ IoU}&\makecell[c]{mIoU} & \makecell[c]{Substitution \\ Improvement} \\ %
				\hline%
				75.0m/ 75.0m / 75.0m & 96k & \xmark  & 97.87 & 93.42 & 95.65 &  -  \\%
				75.0m/ 75.0m / 37.5m & 96k & \cmark  & 97.92 & 93.56 & 95.74 &  +0.11  \\%
				75.0m/ 75.0m / 25m & 96k & \cmark  & 97.98 & 93.76 & 95.87 & +0.22  \\%
			\end{tabular}
		}
		\caption{Potsdam Building dataset.}
	\end{subtable}
	\caption{Improving the default prediction by tile merging on the Open Cities AI dataset using UPerNet+ConvNext as architecture.}
	\label{table_experiments_inference_fusion}
\end{table*}

%% file: content/figures/figure_experiments_inference_fusion_qualitative_open_cities_potsdam.tex
\begin{figure}
	\centering
	\newcommand{\fusionvspace}{\vspace{0.0cm}}
	\newcommand{\tilingtypefiguretext}{0.27\textwidth}
	\newcommand{\tilingtypefigurewidth}{1\textwidth}
	\begin{subfigure}[t]{\tilingtypefiguretext}
		\centering
		\includegraphics[width=\tilingtypefigurewidth]{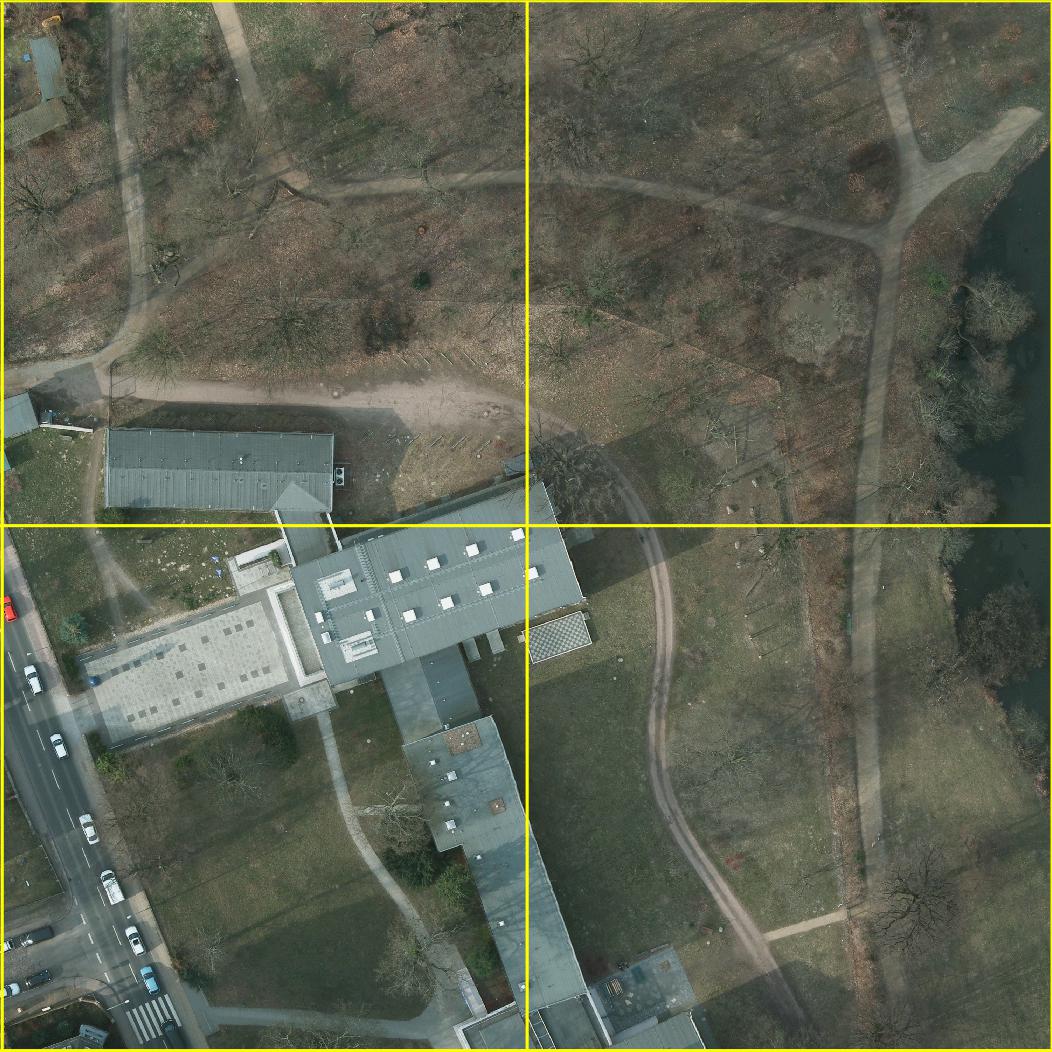} \\
		\fusionvspace
		\includegraphics[width=\tilingtypefigurewidth]{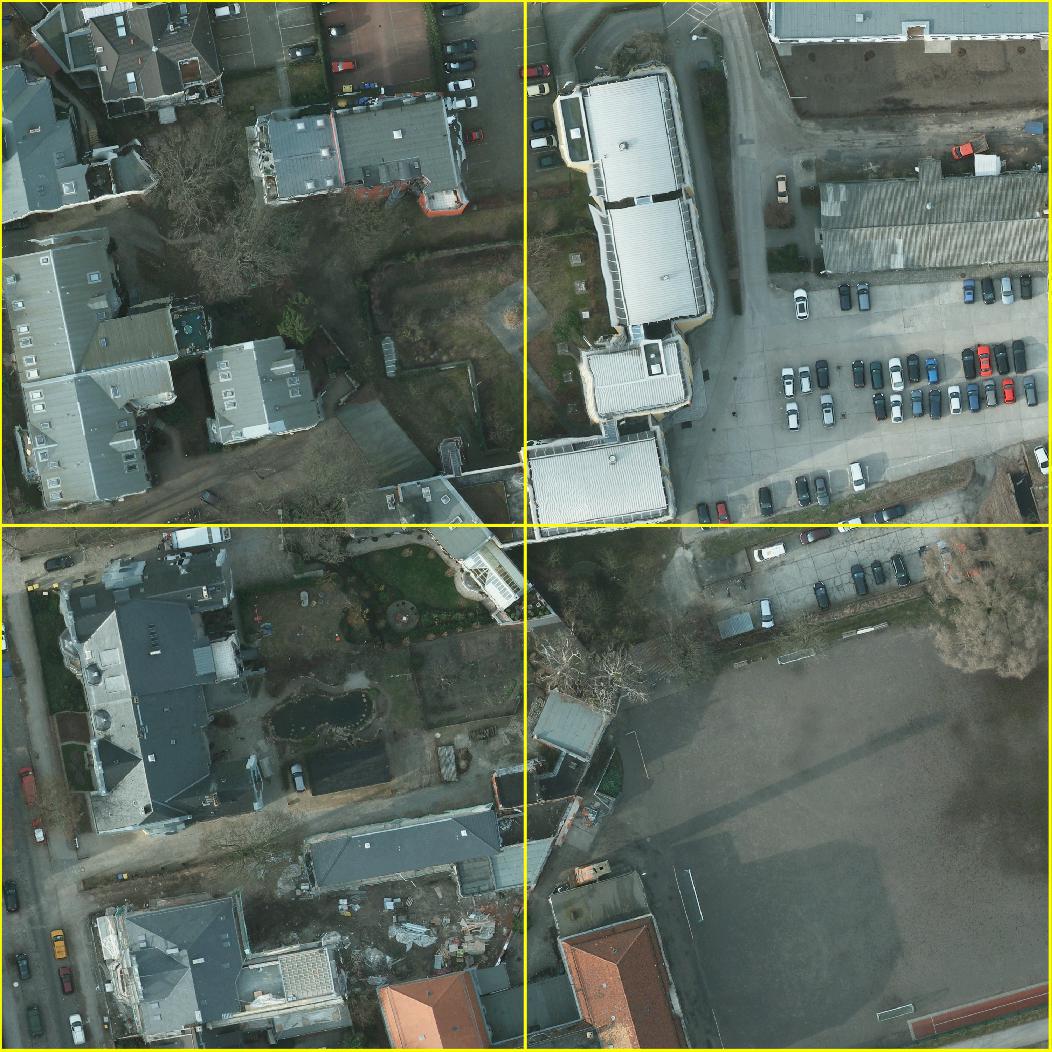} \\
		\fusionvspace
		\includegraphics[width=\tilingtypefigurewidth]{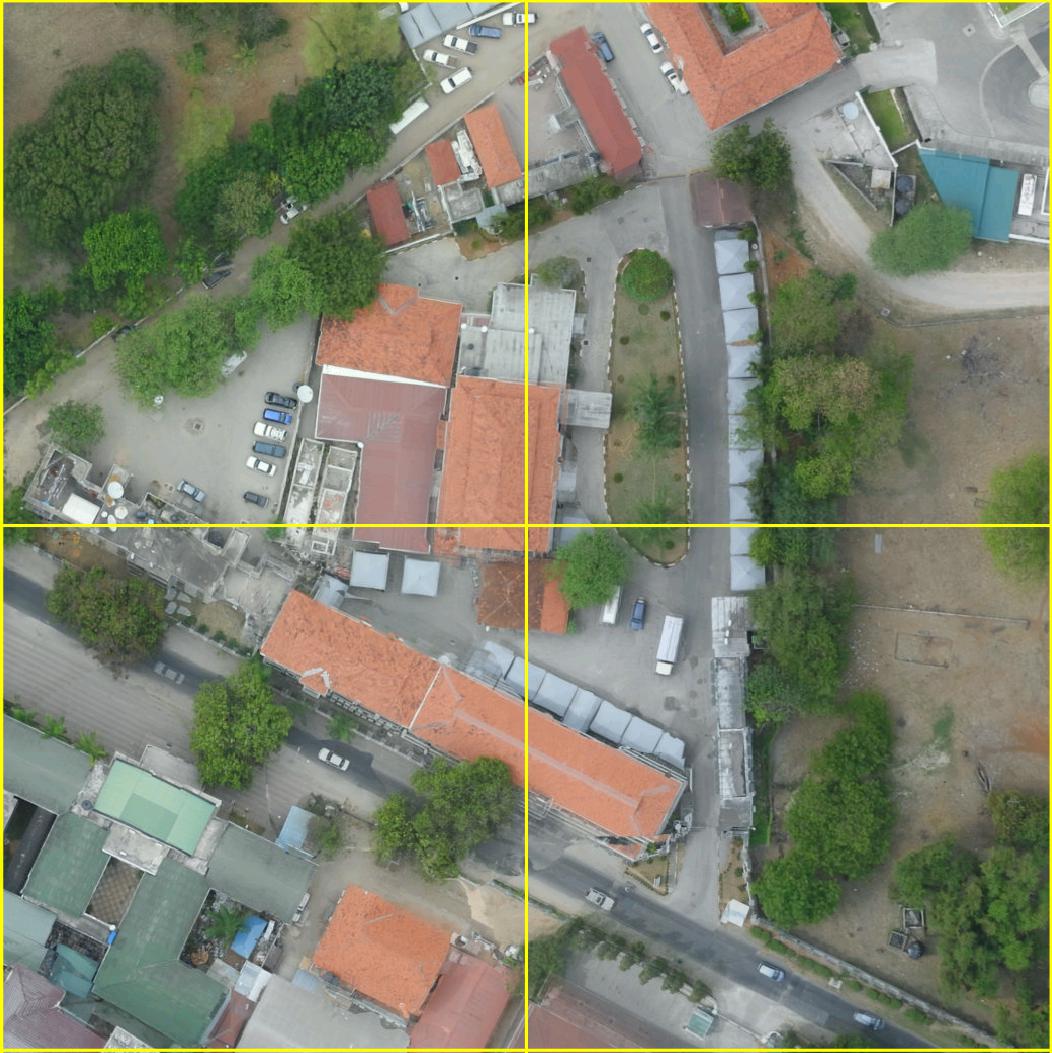} \\
		\fusionvspace
		\includegraphics[width=\tilingtypefigurewidth]{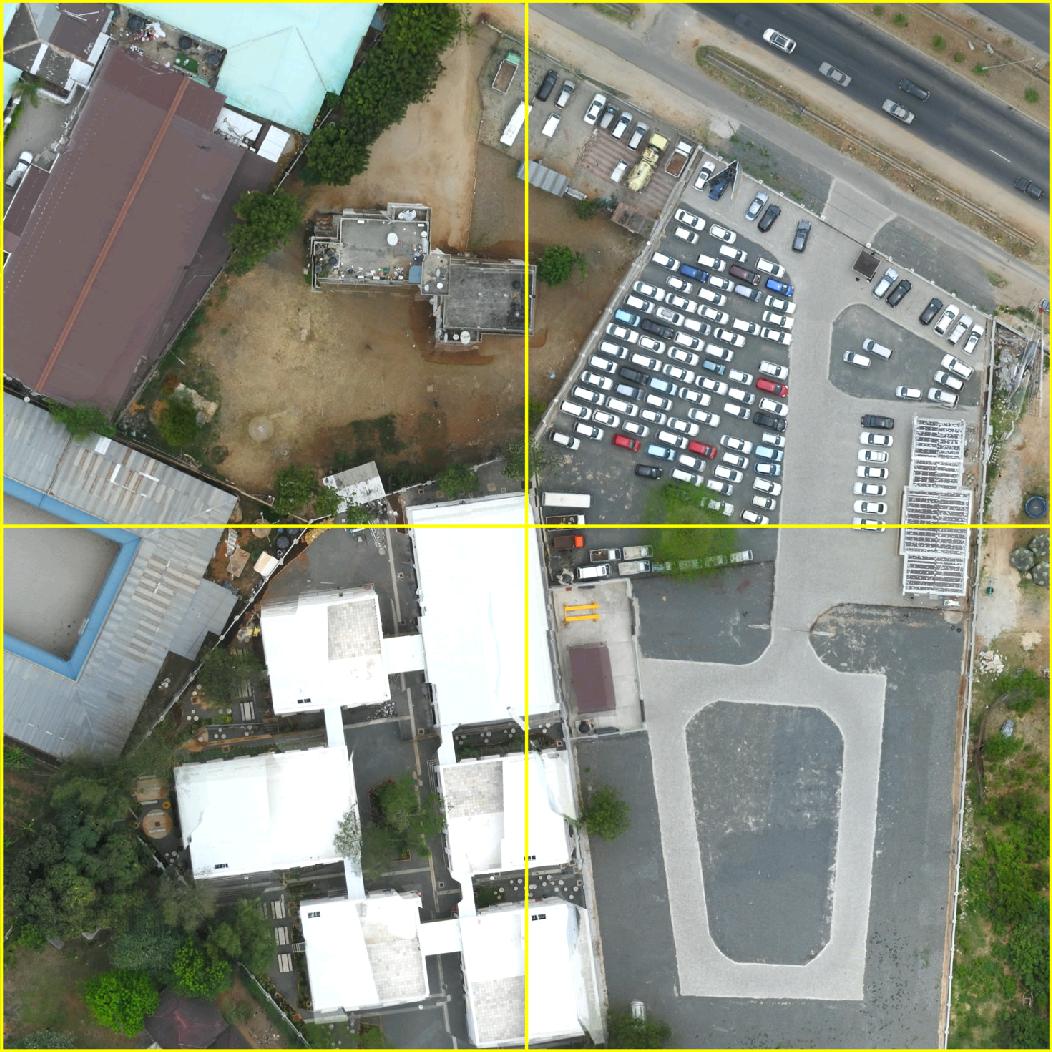}
		\caption{Original images.}
		\label{figure_experiments_inference_fusion_open_cities_potsdam_a}
	\end{subfigure}
	\hfil
	\begin{subfigure}[t]{\tilingtypefiguretext}
		\centering
		\includegraphics[width=\tilingtypefigurewidth]{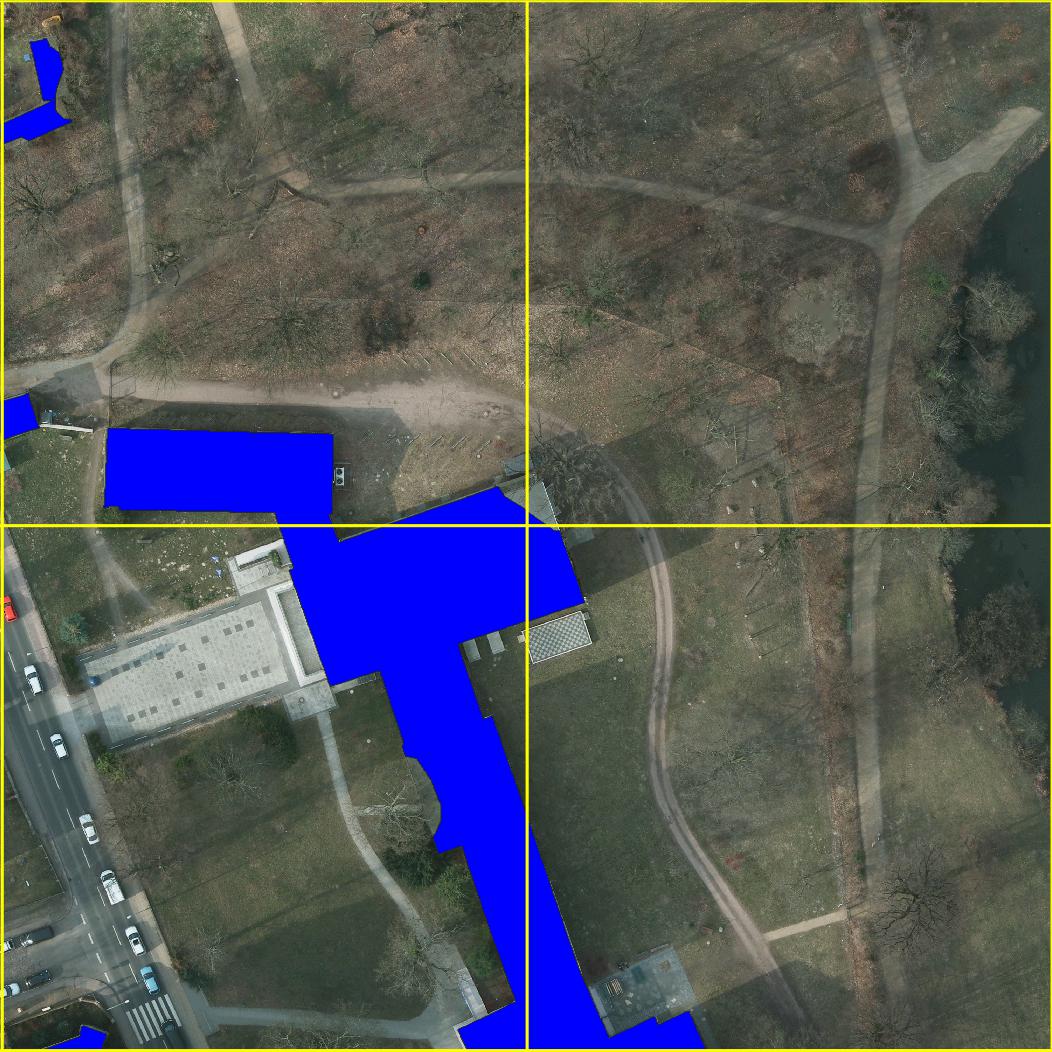} \\
		\fusionvspace
		\includegraphics[width=\tilingtypefigurewidth]{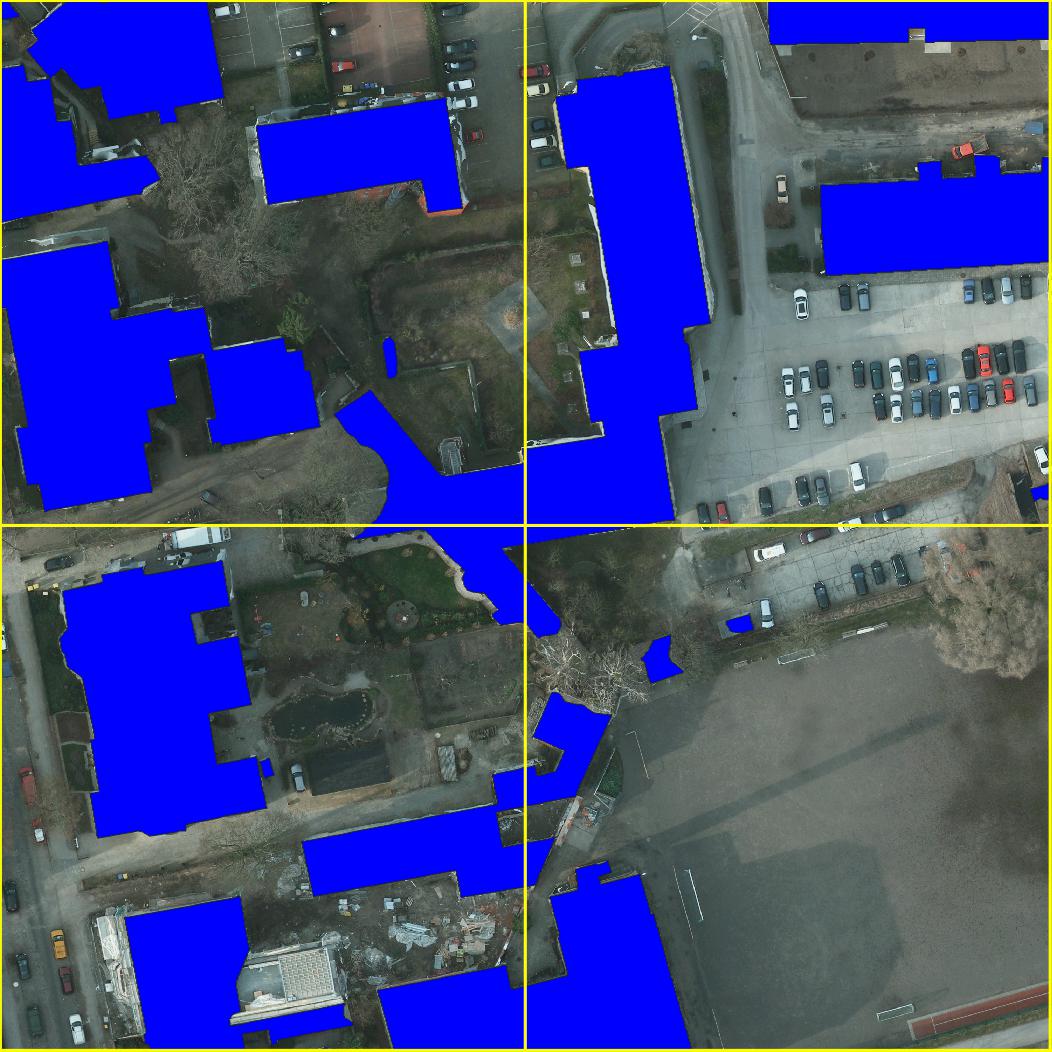} \\
		\fusionvspace
		\includegraphics[width=\tilingtypefigurewidth]{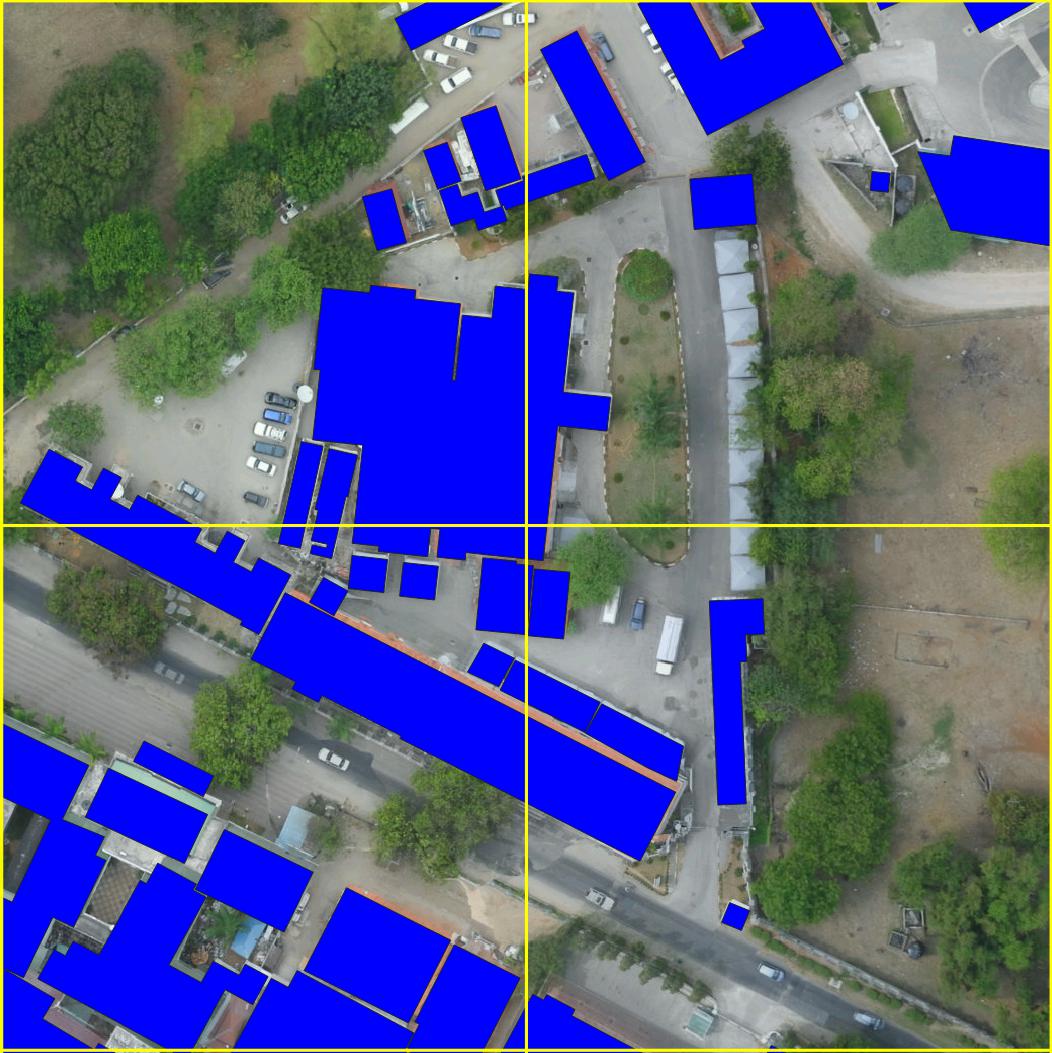} \\
		\fusionvspace
		\includegraphics[width=\tilingtypefigurewidth]{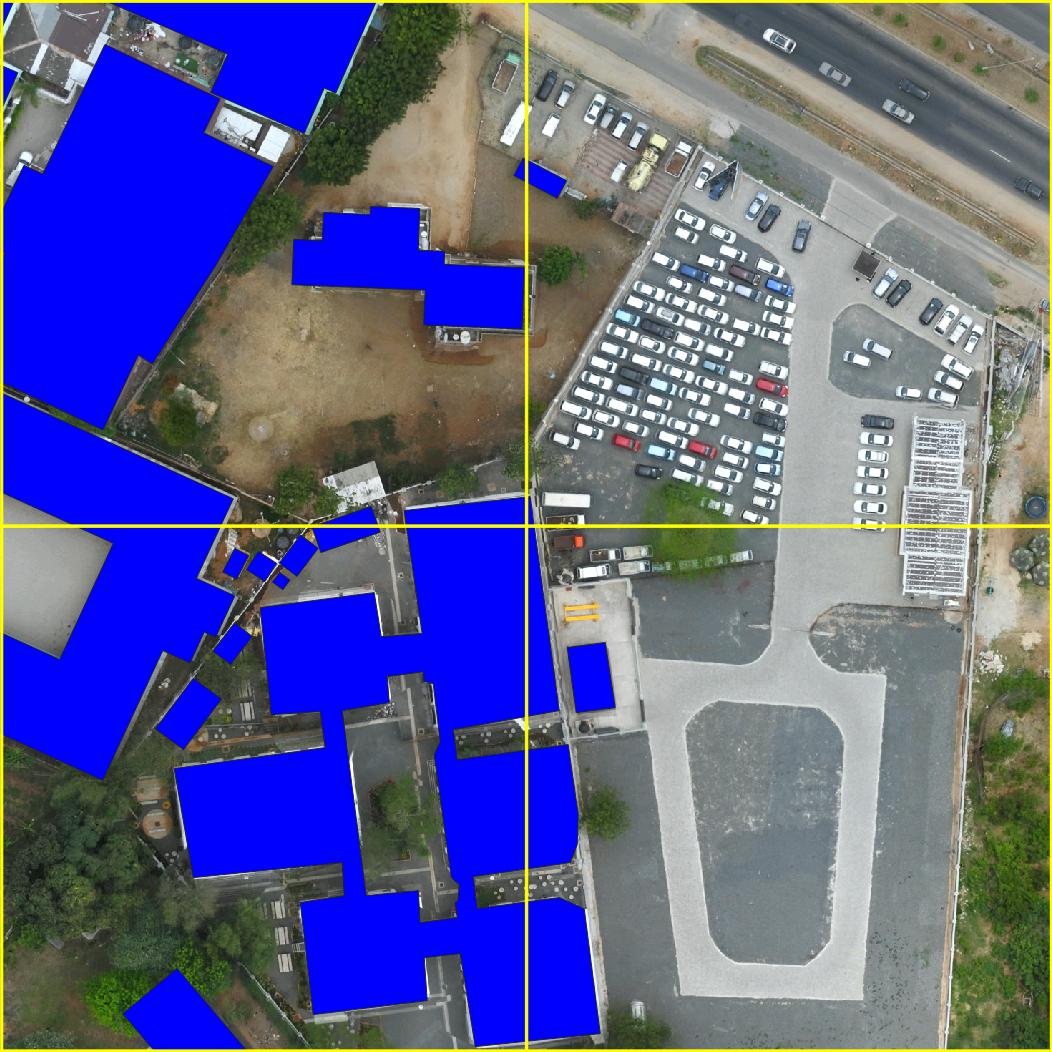}
		\caption{Ground truth.}
		\label{figure_experiments_inference_fusion_open_cities_potsdam_b}
	\end{subfigure}
	\hfil
	\begin{subfigure}[t]{\tilingtypefiguretext}
		\centering
		\includegraphics[width=\tilingtypefigurewidth]{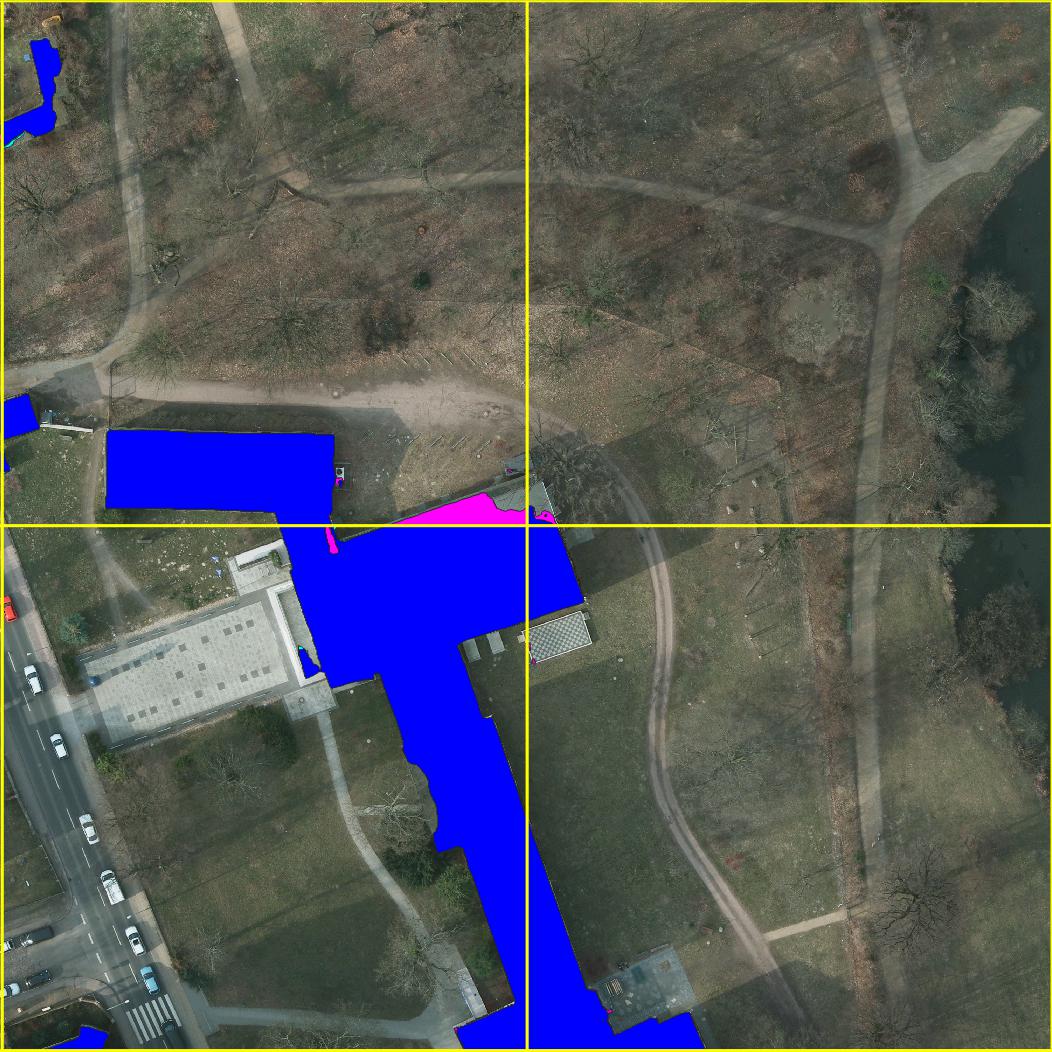} \\
		\fusionvspace
		\includegraphics[width=\tilingtypefigurewidth]{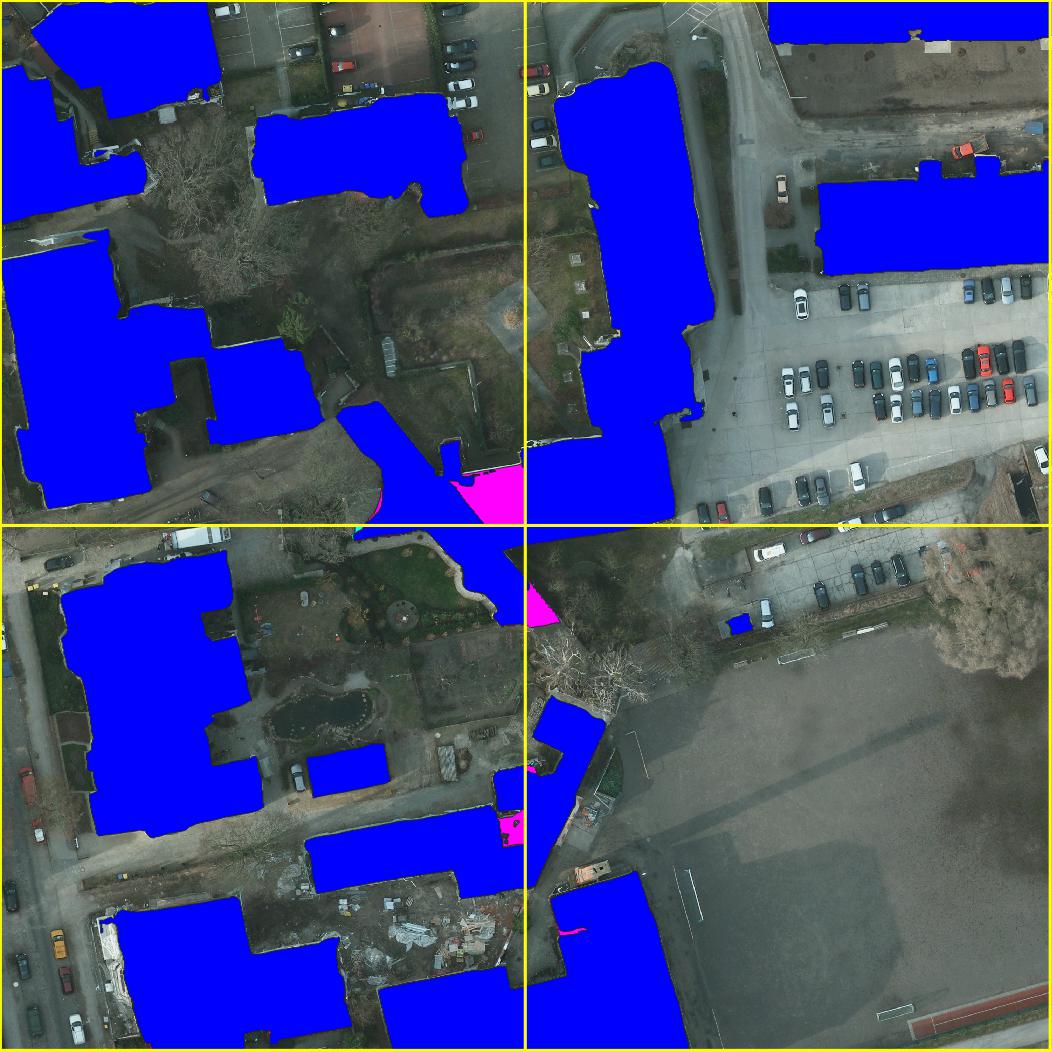} \\
		\fusionvspace
		\includegraphics[width=\tilingtypefigurewidth]{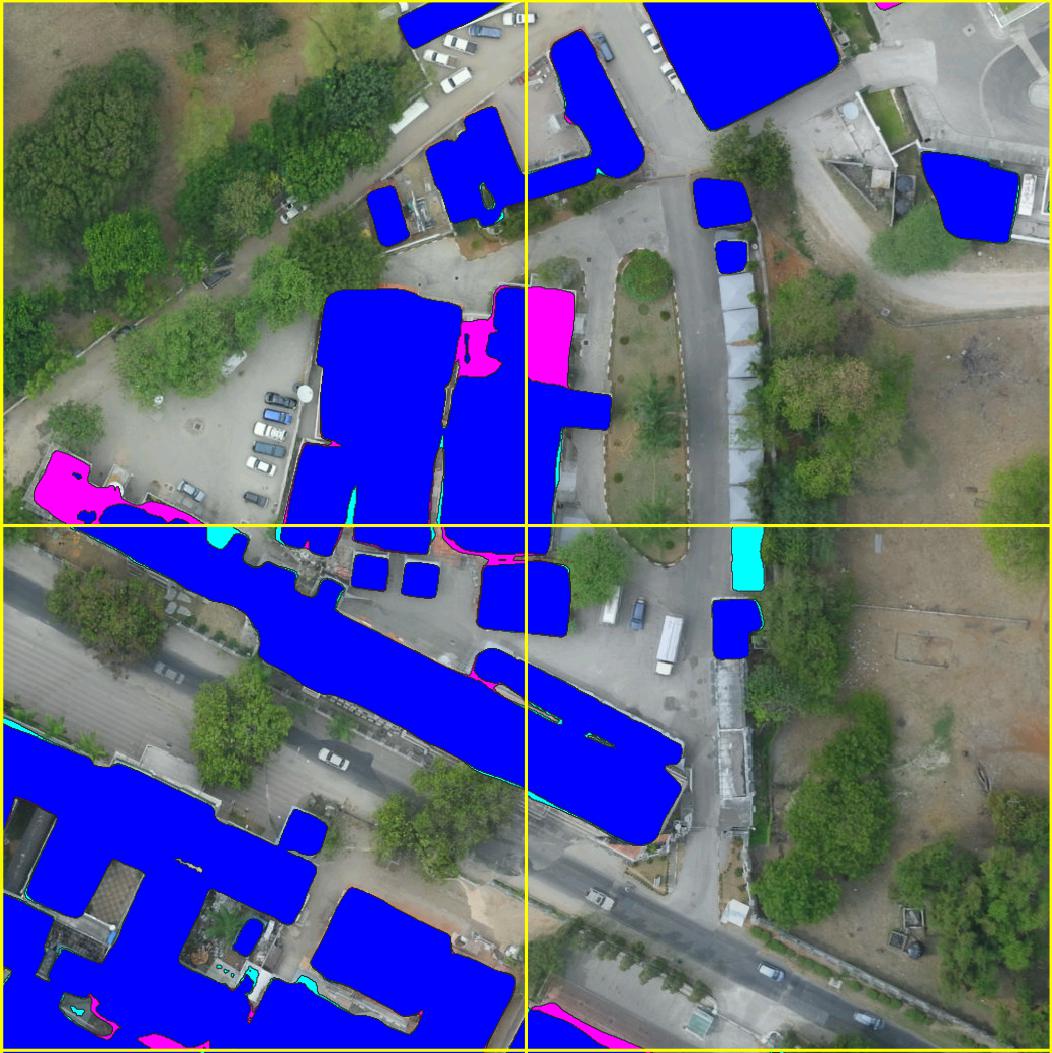} \\
		\fusionvspace
		\includegraphics[width=\tilingtypefigurewidth]{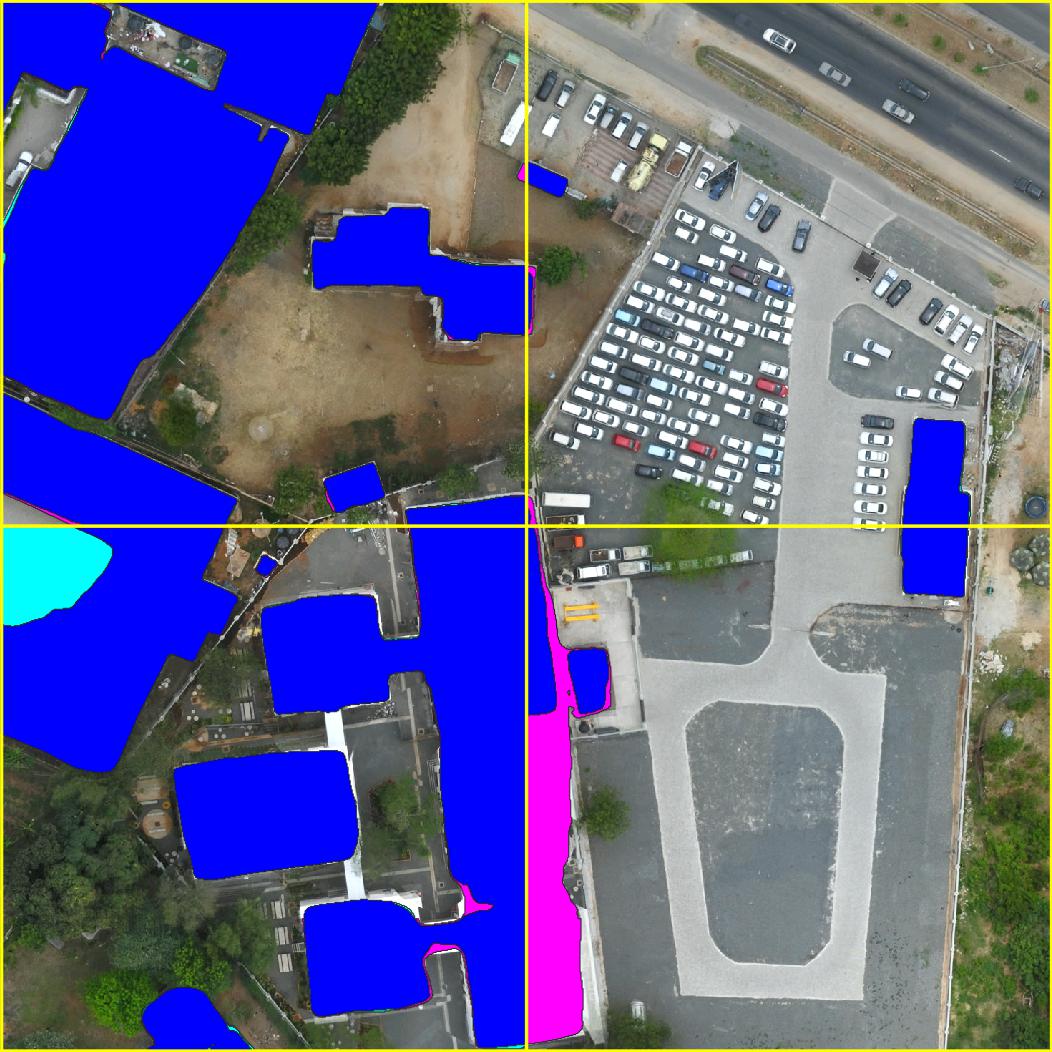}
		\caption{Fusion result.}
		\label{figure_experiments_inference_fusion_open_cities_potsdam_c}
	\end{subfigure}
	\caption{Comparison of semantic segmentations of base tiles with results obtained by the tile fusion approach proposed in section \secref{section_improve_inference_reliability} using the Potsdam data (first and second row) and the Open Cities AI dataset (third and fourth row). The colors represent the following meaning: blue building pixels are contained in both results, pink building pixels are only part of the fused segmentation and teal building pixels are exclusively present in the segmentation of the base tiles.}
	\label{figure_experiments_inference_fusion_open_cities_potsdam}
\end{figure}

%% file: content/conclusion.tex
This paper presents a novel pipeline for semantic segmentation of earth observation imagery using a tiling scheme and tile inference approach that exhibits several advantages over commonly used pixel-based or web map tiling systems. For instance, its flexible tiling layout properties (\ie different degrees of overlap) allow performing a tiling specific training data augmentation as well as a tile fusion approach that maximizes the context information of pixel predictions during inference time. The generated tiles preserve the spatial extent \wrt heterogeneous sensors, varying recording distances and different latitudes. Since the tile size affects the available context data of each pixel and influences the information loss while potentially down-sampling the tile data to the input size of the segmentation model, it is difficult to determine a universal (optimal) tile size. To perform our experiments with reasonable tile sizes we conduct a comprehensive evaluation to empirically determine a suitable tile size for the task of building segmentation. Our quantitative evaluation demonstrates that the proposed tiling and tile fusion approach allows us to further improve the results of current state-of-the-art semantic segmentation models. To foster future research we make the source code of the proposed processing pipeline publicly available.

%% file: content/supplementary.tex
\subsection{State-of-the-Art Reference Model}

To determine a suitable model for earth observation imagery we perform a quantitative evaluation of several state-of-the-art segmentation methods and powerful backbones including UPerNet \citep{Xiao2018ECCV} with ConvNext \citep{Liu2022CVPR}, KNet \citep{Zhang2021NeurIPS} with Swin \citep{Liu2021ICCV}, Segmenter \citep{Strudel2021ICCV} and BEiT \citep{Bao2022ICLR}. We complement this evaluation using the well known baseline method PSPNet \citep{Zhao2017CVPR} with ResNet \citep{Wightman2021resnet}. An overview of the results is presented in \tabref{table_experiments_architecture_evaluation}. Our evaluation shows that UPerNet \citep{Xiao2018ECCV} with ConvNext \citep{Liu2022CVPR} achieves the best results while utilizing a reasonable number of parameters. The low memory requirements of UPerNet with ConvNext compared to the vision transformers mentioned above allows us to increase the batch size to four samples. At the same time a smaller number of parameters reduces the computational costs resulting in an acceleration of training and testing.

\subsection{Orientation Specific Data Augmentation}

The previously presented results employed a standard data augmentation scheme following the \cite{mmseg2020}, and include a \emph{Resize}, a \emph{RandomCrop}, a \emph{RandomFlip}, and a \emph{PhotoMetricDistortion} step. In the context of earth observation imagery several adjustments are reasonable such as the integration of a rotation and a vertical flip, since there is no canonical (up-down / left-right) direction. In addition, neglecting the tile resizing during augmentation might be beneficial for GSD-based tiling schemes, since the extracted tiles capture the scene at fixed real world scale, \ie specific categories like vehicles show a similar extent across different images. The corresponding evaluation is contained in \tabref{table_experiments_scale_prior_evaluation}. Since the modifications of the data augmentation do not lead to an improvement we use the default data augmentation proposed by the \cite{mmseg2020} for our experiments.

\input{content/tables/table_experiments_architecture_evaluation}
\input{content/tables/table_experiments_scale_prior_evaluation}

\subsubsection{Qualitative Results}
\input{content/figures/figure_experiments_tile_aggregation_open_cities}
\input{content/figures/figure_experiments_tile_aggregation_potsdam}
\figref{figure_tile_aggregation_open_cities} and \figref{figure_tile_aggregation_potsdam} present some qualitative results for the Open Cities AI and the Potsdam dataset.

%% file: content/tables/table_experiments_architecture_evaluation.tex
\begin{table*}
	\centering
	\resizebox{\textwidth}{!}{%
	\begin{tabular}{
			ccccccc
		} 
		Architecture & \makecell[c]{Number \\ Parameters}  & \makecell[c]{Batch \\ Size} & \makecell[c]{Background \\ IoU} & \makecell[c]{Building \\ IoU} & mIoU \\
		\hline  
		PSPNet \citep{Zhao2017CVPR} + ResNet50 \citep{Wightman2021resnet} & 49 M & 4 & 95.28 & 71.06 &   83.17 \\
		\makecell[c]{K-Net \citep{Zhang2021NeurIPS} + PSPNet \cite{Zhao2017CVPR} + \\ ResNet50 \citep{Wightman2021resnet}} & 62 M & 2 & 95.25& 70.08 & 82.66 \\
		KNet \citep{Zhang2021NeurIPS} + SwinL \citep{Liu2021ICCV} & 247 M & 2 & 95.65 & 73.21 & 84.43 \\	%
		UPerNet \citep{Xiao2018ECCV} + ConvNext \citep{Liu2022CVPR} & 235 M & 4 & \textbf{95.73} & \textbf{73.36} & \textbf{84.55}  \\
		Segmenter \citep{Strudel2021ICCV} & 333 M & 1 & 95.26 & 70.89 & 83.07 \\
		BEiT \citep{Bao2022ICLR} & 440 M & 2 & 92.65 & 51.04 & 71.84 \\
	\end{tabular}
	} \\
	\caption{Comparison of prominent architectures on the Open Cities AI dataset. The dataset has been split into training and test data using a ratio of 80/20. The image and label data has been tiled using a size of 25m and a stride of 25m, which results in 183 k tiles. All models have been trained with 320k iterations. The data augmentation consists of \emph{Resize}, \emph{RandomCrop}, \emph{RandomFlip}, and \emph{PhotoMetricDistortion}.}
	\label{table_experiments_architecture_evaluation}
\end{table*}

%% file: content/tables/table_experiments_scale_prior_evaluation.tex
\begin{table*}[tbh]
	\tablefontsizestiny
	\newcommand{\tileinfocol}{2cm}
	\newcommand{\tileextent}{3.8cm}
	\newcommand{\scalepriorcol}{2cm}
	\newcommand{\numtilecol}{1.5cm}
	\newcommand{\mioucol}{1cm}
	\newcommand{\buildcol}{1.5cm}
	\newcommand{\backcol}{2cm}
	\centering
	\begin{tabular}{
			ccccccc
		} 
		\makecell[c]{Tile Info \\ Size / Train Stride} & \makecell[c]{Adjustment \\ Prior} & \makecell[c]{Best\\Checkpoint}  & \makecell[c]{Background \\ IoU}  & \makecell[c]{Building \\ IoU} & mIoU & \makecell[c]{Augmentation \\ Improvement} \\
		\hline
		75m / 37.5m &  Baseline & 288 k & \textbf{95.7} & \textbf{73.52} & \textbf{84.61} & - \\
		75m / 37.5m &  Baseline /wo Resizing  & 320 k & 95.67 & 73.28 & 84.48 & -0.13 \\
		75m / 37.5m &  Baseline /w Vertical Flip & 224 k & 95.6 & 73.09 & 84.34 & -0.27 \\
		75m / 37.5m &  Baseline /w Rotation & 160 k & 95.58 & 73.13 & 84.36 & -0.25 \\
	\end{tabular}
	\caption{Scale prior on the Open Cities AI dataset using UPerNet+ConvNext as architecture (batch size: 4, Iterations 320k). The dataset has been split into training and test data using a ratio of 80/20. }
	\label{table_experiments_scale_prior_evaluation}
\end{table*}

%% file: content/figures/figure_experiments_tile_aggregation_open_cities.tex
\newcommand\tileaggregationopencitiessubfigurewidth{0.33\textwidth}
\begin{figure*}[t]
	\begin{subfigure}{\textwidth}
		\centering
		\includegraphics[width=\tileaggregationopencitiessubfigurewidth]{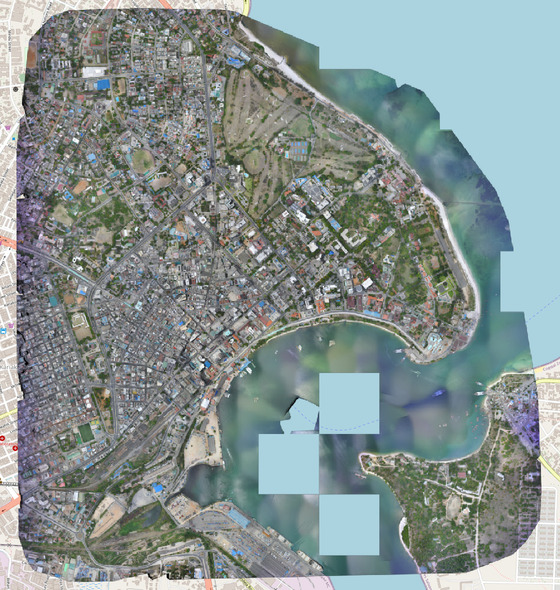}
		\hfil
		\includegraphics[width=\tileaggregationopencitiessubfigurewidth]{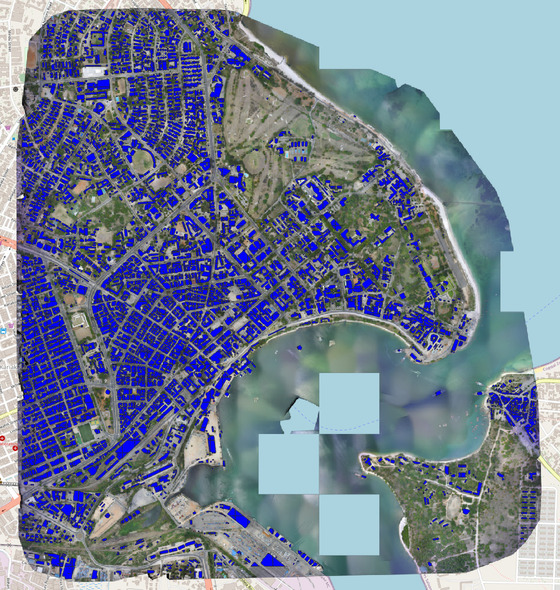}
		\caption{Result for an (entire) earth observation image.}
		\label{figure_tile_aggregation_open_cities_a}
	\end{subfigure}
	\begin{subfigure}{\textwidth}
		\centering
		\includegraphics[width=\tileaggregationopencitiessubfigurewidth]{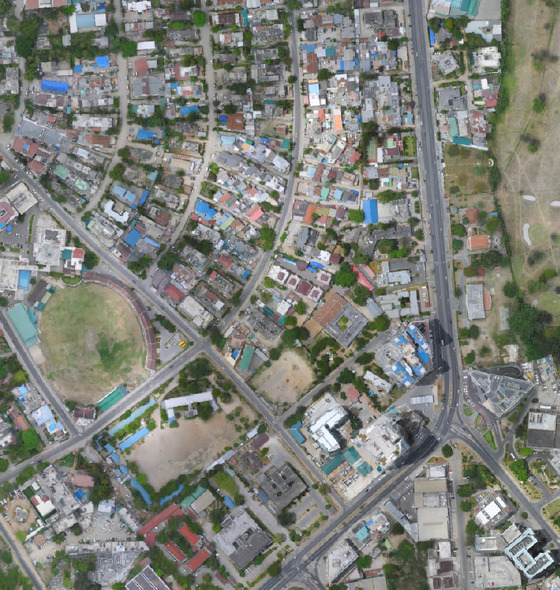}
		\hfil
		\includegraphics[width=\tileaggregationopencitiessubfigurewidth]{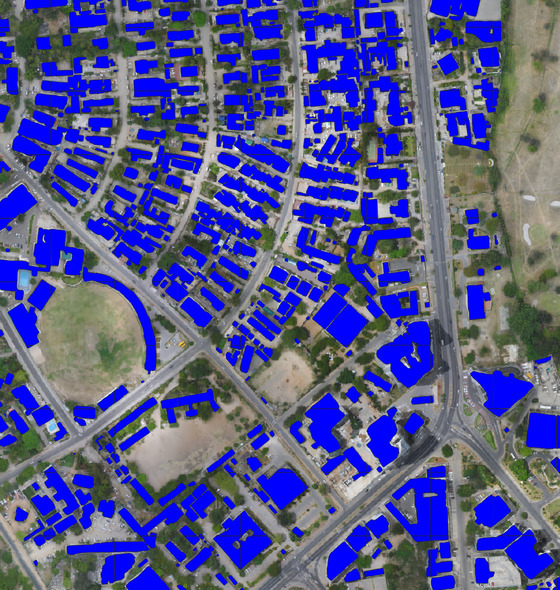}
		\caption{Result for a subarea of \figref{figure_tile_aggregation_open_cities_a}.}
		\label{figure_tile_aggregation_open_cities_b}
	\end{subfigure}
	\begin{subfigure}{\textwidth}
		\centering
		\includegraphics[width=\tileaggregationopencitiessubfigurewidth]{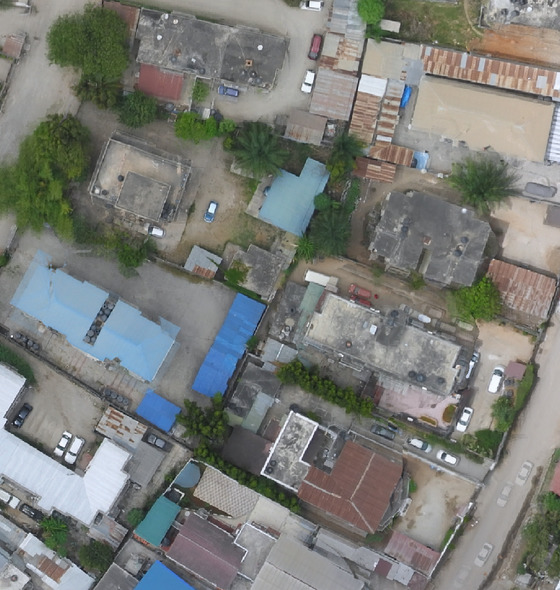}
		\hfil
		\includegraphics[width=\tileaggregationopencitiessubfigurewidth]{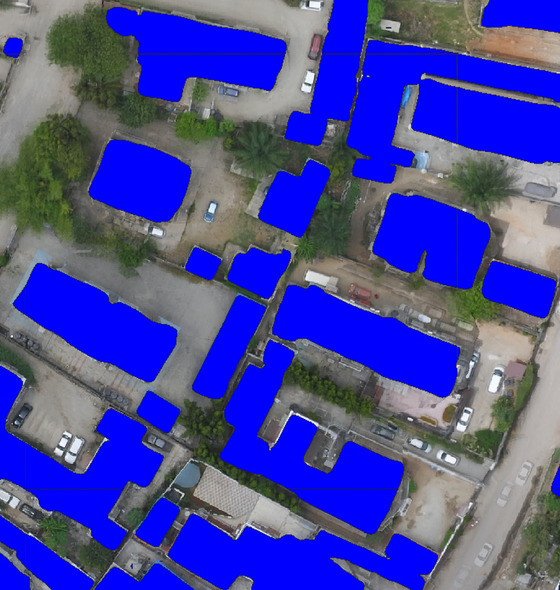}
		\caption{Result for a subarea of \figref{figure_tile_aggregation_open_cities_b}.}
		\label{figure_tile_aggregation_open_cities_c}
	\end{subfigure}
	\caption{Qualitative segmentation results on the Open Cities AI dataset using a tile size of 75m.}
	\label{figure_tile_aggregation_open_cities}
\end{figure*}

%% file: content/figures/figure_experiments_tile_aggregation_potsdam.tex
\newcommand\tileaggregationsubfigurewidth{0.4\textwidth}
\begin{figure*}[t]
		\centering
		\begin{subfigure}{\tileaggregationsubfigurewidth}
			\includegraphics[width=\textwidth]{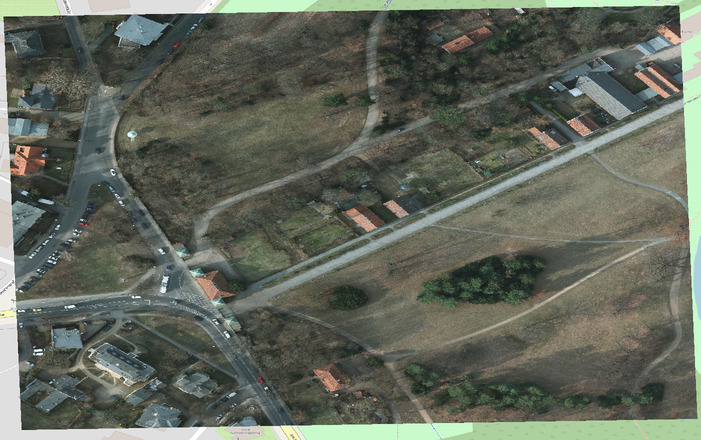}
			\caption{Earth observation image.}
		\end{subfigure}
		\hfil
		\begin{subfigure}{\tileaggregationsubfigurewidth}
			\includegraphics[width=\textwidth]{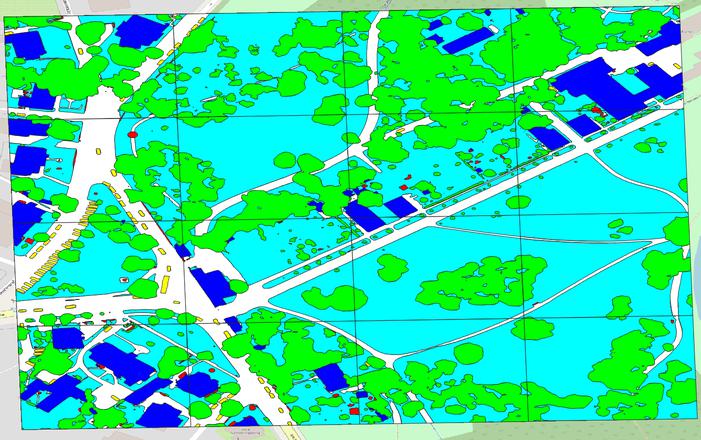}
			\caption{Result of all categories.}
		\end{subfigure}
		\begin{subfigure}{\tileaggregationsubfigurewidth}
			\includegraphics[width=\textwidth]{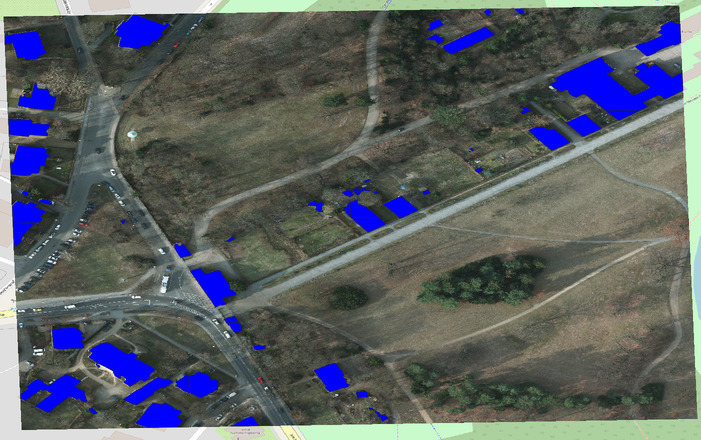}
			\caption{Result for category \emph{building}.}
		\end{subfigure}
		\hfil
		\begin{subfigure}{\tileaggregationsubfigurewidth}
			\includegraphics[width=\textwidth]{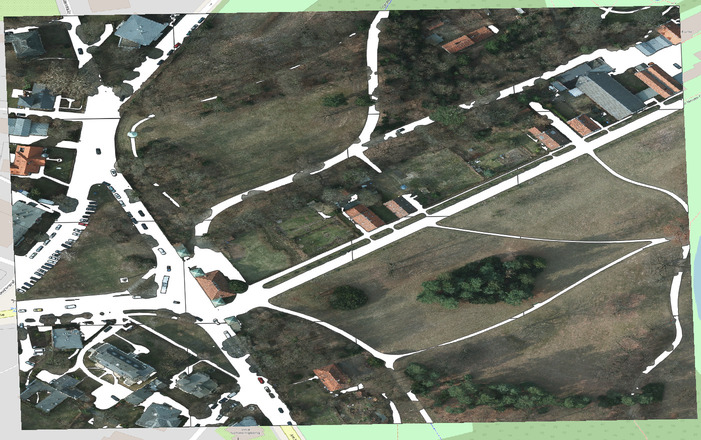}
			\caption{Result for category \emph{impervious surface}.}
		\end{subfigure}
		\begin{subfigure}{\tileaggregationsubfigurewidth}
			\includegraphics[width=\textwidth]{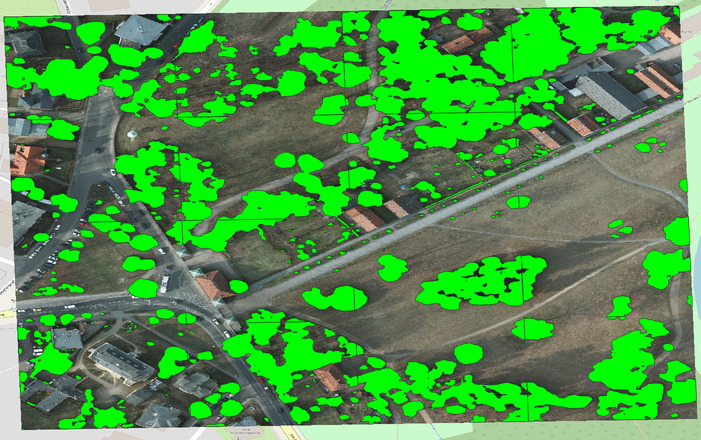}
			\caption{Result for category \emph{tree}.}
		\end{subfigure}
		\hfil
		\begin{subfigure}{\tileaggregationsubfigurewidth}
			\includegraphics[width=\textwidth]{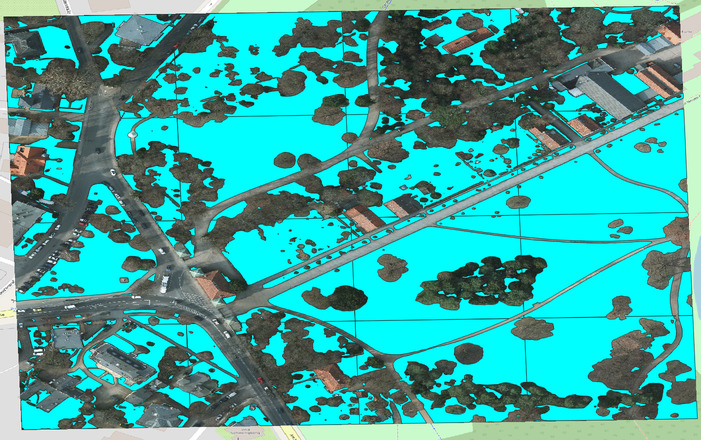}
			\caption{Result for category \emph{low vegetation}.}
		\end{subfigure}
		\begin{subfigure}{\tileaggregationsubfigurewidth}
			\includegraphics[width=\textwidth]{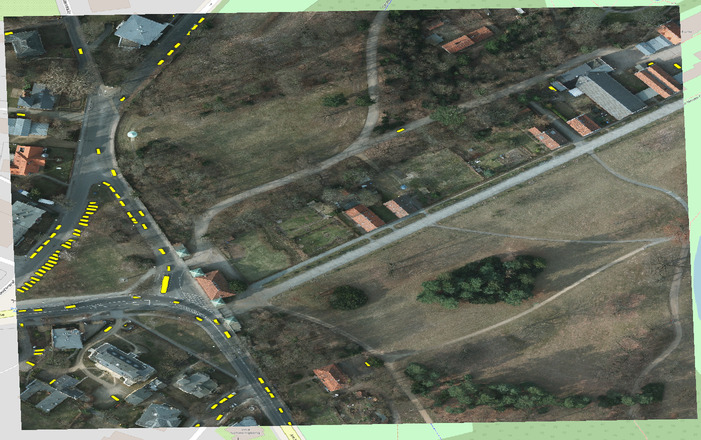}
			\caption{Result for category \emph{car}.}
		\end{subfigure}
		\hfil
		\begin{subfigure}{\tileaggregationsubfigurewidth}
			\includegraphics[width=\textwidth]{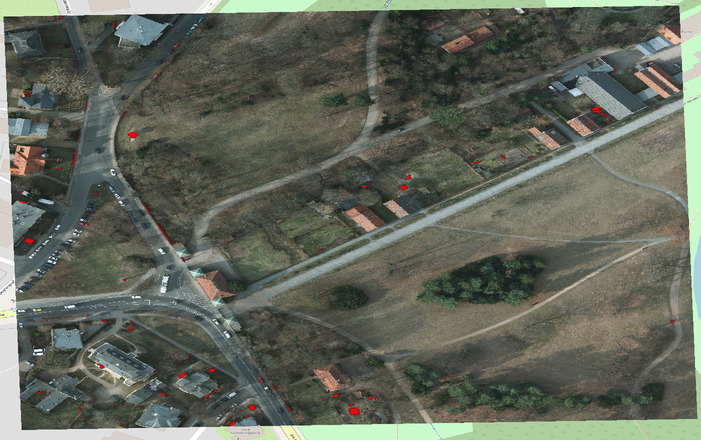}
			\caption{Result for category \emph{background}.}
		\end{subfigure}
	\caption{Qualitative segmentation results on the potsdam dataset for tiles with size 75m.}
	\label{figure_tile_aggregation_potsdam}
\end{figure*}